\documentclass[preprint,12pt,3p]{elsarticle}

\usepackage{booktabs}
\usepackage{makecell}
\usepackage{url}
\usepackage{graphicx}
\usepackage{balance}
\usepackage{amsmath}
\usepackage{subfigure}
\usepackage{graphicx}
\usepackage{balance}
\usepackage{color}
\usepackage{comment}
\usepackage{array}
\usepackage{lipsum}
\usepackage{caption}
\usepackage{blindtext}
\usepackage{enumitem}
\usepackage{lipsum}
\usepackage{capt-of}
\usepackage{tikz}
\usetikzlibrary{arrows,positioning,automata}
\usepackage{balance}
\usepackage{amsmath}
\usepackage[T1]{fontenc}
\usepackage{algpseudocode}
\usepackage{algorithm}
\usepackage{subfigure}
\usepackage{graphicx}
\usepackage[utf8]{inputenc}
\usepackage{balance}
\usepackage{multirow}
\usepackage{color}
\usepackage{comment}
\usepackage{hyperref}
\usepackage{array}
\usepackage{soul}
\usepackage[justification=centering]{caption}
\usepackage[lighttt]{lmodern}
\usepackage{libertine}
\usepackage{amssymb}
\usepackage{float}
\usepackage{algorithm}
\usepackage{algpseudocode}
\usepackage{multirow}
\usepackage{longtable}
\usepackage{booktabs}
\usepackage{subfigure} 
\usepackage{subcaption}
\usepackage{graphicx}
\usepackage{enumitem}
\usepackage{bbding}
\usepackage{pgf-pie}

\usepackage{graphicx}
\usepackage{tikz}
\usepackage{orcidlink}
\usepackage{forest}
\usetikzlibrary{trees,positioning,shapes,shadows,arrows.meta}

\AtBeginDocument{
  }

\journal{Computer Methods and Programs in Biomedicine}

\begin{document}

\begin{frontmatter}


\title{A Comprehensive Survey of Mamba Architectures for Medical Image Analysis: Classification, Segmentation, Restoration and Beyond}

\author[label1]{Shubhi Bansal\orcidlink{0000-0002-8034-8220}\corref{cor1}}
\address[label1]{Indian Institute of Technology Indore, India}

\cortext[cor1]{Authors contributed equally to this research.}
\cortext[cor2]{Corresponding author}

\ead{phd2001201007@iiti.ac.in}

\author[label2]{Sreeharish A\orcidlink{0009-0007-0944-5451}\corref{cor1}}
\address[label2]{R.M.D. Engineering College, Kavaraipettai, India}
\ead{sreeharisharikrishnan@gmail.com}

\author[label2]{Madhava Prasath J\orcidlink{0009-0009-8673-3507}\corref{cor1}}
\ead{Madhavprasath088@gmail.com}

\author[label2]{Manikandan S\orcidlink{0009-0002-8975-8100}\corref{cor1}}
\ead{manikandan.s.jeeva@gmail.com}

\author[label3]{Sreekanth Madisetty\orcidlink{}\corref{cor1}}
\address[label3]{Jio Platforms Limited, India}
\ead{sreekanth2.m@ril.com}

\author[label1]{Mohammad Zia Ur Rehman\orcidlink{0000-0001-6374-8102}\corref{cor1}}
\ead{phd2101201005@iiti.ac.in}

\author[label1]{Chandravardhan Singh Raghaw\orcidlink{0009-0003-2268-9507}\corref{cor1}}
\ead{phd2201101016@iiti.ac.in}

\author[label4]{Gaurav Duggal\orcidlink{}}
\address[label4]{Birla Institute of Technology \& Science Pilani, India}
\ead{p20230302@pilani.bits-pilani.ac.in}

\author[label1]{Nagendra Kumar\orcidlink{}\corref{cor2}}
\ead{cnagendra@iiti.ac.in}






\begin{abstract}
Mamba, a special case of the State Space Model, is gaining popularity as an alternative to template-based deep learning approaches in medical image analysis. While transformers are powerful architectures, they have drawbacks, including quadratic computational complexity and the inability to efficiently address long-range dependencies. This limitation affects the analysis of large and complex datasets in medical imaging, where there are many spatial and temporal relationships. In contrast, Mamba offers benefits that make it well suited for medical image analysis. It has linear time complexity, which is a significant improvement over transformers. In sequence modeling tasks, computational complexity grows linearly with the length of the input sequence. Mamba processes longer sequences without attention mechanisms, enabling faster inference and requiring less memory. Mamba also demonstrates strong performance in merging multimodal data, improving diagnosis accuracy and patient outcomes. The paper’s organization allows readers to appreciate Mamba's capabilities in medical imaging step by step. We begin with clear definitions of relevant concepts regarding SSMs and concept models, including S4, S5, and S6. We then explore Mamba architectures, including pure Mamba, U-Net variants, and hybrid models that combine Mamba with convolutional networks, transformers, and Graph Neural Networks. Subsequent sections cover Mamba optimizations, techniques such as weakly supervised and self-supervised learning, scanning mechanisms, and a detailed analysis of applications across various tasks. We provide an overview of available datasets and several experimental results regarding Mamba’s efficacy in different domains. Furthermore, we detail the challenges and limitations of Mamba, along with other interesting aspects and possible future directions. The final subsection explains the importance of Mamba in medical imaging and provides an analysis and conclusions regarding its usage and enhancement measures. This review aims to demonstrate the transformative potential of Mamba in overcoming existing barriers within medical imaging while paving the way for innovative advancements in the field. A comprehensive list of Mamba architectures applied in the medical field, reviewed in this work, is available on Github\footnotemark[4].
\end{abstract}


\end{frontmatter}





\footnotetext[4]{\href{https://github.com/Madhavaprasath23/Awesome-Mamba-Papers-On-Medical-Domain}{https://github.com/Madhavaprasath23/Awesome-Mamba-Papers-On-Medical-Domain}}

\section{Introduction}

In recent few decades, there has been a remarkable improvement in the field of medicine through the application of machine learning \cite{shehab2022machine} as well as deep learning \cite{suzuki2017overview}. The initial architectures of neural networks like Convolutional Neural Networks (CNNs) \cite{li2021survey} played a pivotal role in better image segmentation \cite{kayalibay2017cnn}, classification \cite{li2014medical,raghaw2024explainable}, and object detection \cite{li2019clu}. Medical images are complex, but CNNs were able to analyze 3D structures in a 2D plane and so proved useful in biomedical image computing especially in image segmentation \cite{ronneberger2015u}, tumor detection \cite{choudhury2020brain}, organ segmentation \cite{zhao2019knowledge}, and disease diagnosis imaging \cite{chen2019use}. CNNs have been applied extensively to medical imaging tasks, namely segmentation, classification, and reconstruction. One weakness, however, is that they can lack when sequencing data, or multitasking, which requires long-range dependencies. For example, in the area of medical image segmentation, the use of CNNs may not perform well as one would expect since they may not be able to model super-resolution inter-dependence of an image and its parts. 

Some of the drawbacks of CNNs have been addressed by transformers \cite{DBLP:journals/corr/VaswaniSPUJGKP17, raghaw2024cotconet} as advancements of technologies that have better sequential data processing and long-range dependencies. Still, there are some disadvantages as well. The main problem is the scaling of computed attention which grows quadratically with the sequence length, thus makes the use of such attention costly and hard on sequences that are very long. Moreover, many additional resources and data are usually needed, which is quite a problem if one has to work in a resource-restricted environment such as in the medical domain. In regard to the shortcomings of classical CNNs and transformers, there has been noticeable progress in research on different types of models which could efficiently represent the long sequences and their intricate dependencies. Of late, State Space Models (SSMs) \cite{gu2021efficiently} have gained much attention as one of such alternatives as the Mamba \cite{gu2024mambalineartimesequencemodeling} model. 

\begin{figure}[!ht]
\centering
\includegraphics[width=\linewidth]{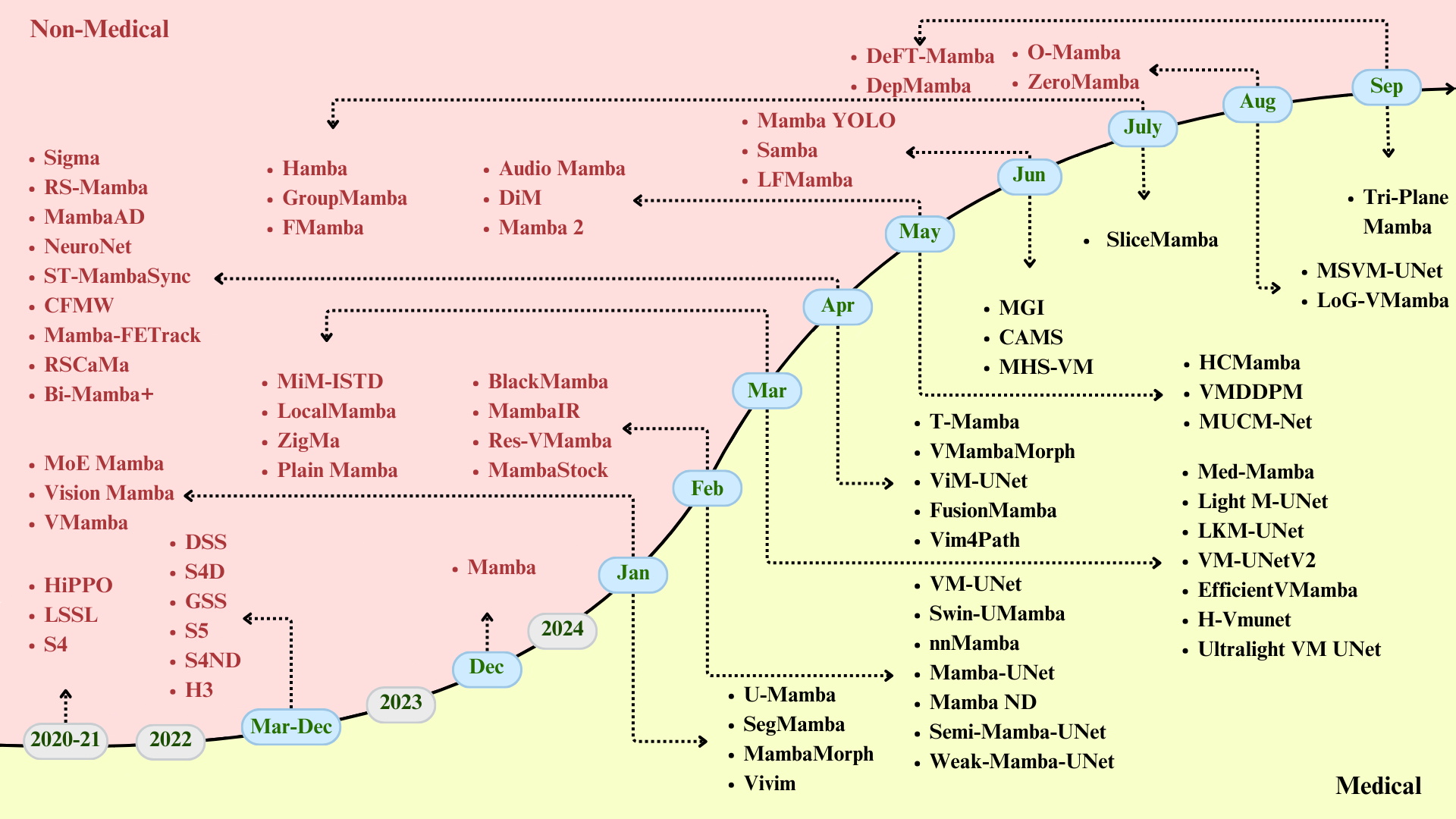}
\caption{Evolution of Mamba from State Space Models (SSMs)}
\label{fig:timeline}
\end{figure} 

Mamba, aims to address the problems related to modern deep learning techniques. Selective state spaces are employed to quickly assimilate vast lengths of sequences, combine various modes, and command extensive yet practical resolutions. The architecture of Mamba incorporates selective scan mechanism and hardware aware algorithm which support high efficiency in storage and computation of the intermediate results. This helps Mamba perform very well in some tasks such as medical image segmentation \cite{xing2024segmambalongrangesequentialmodeling, wu2024hvmunethighordervisionmamba, wang2024lkmunetlargekernelvision, raghaw2025t}, classification \cite{gong2024nnmamba, yue2024medmambavisionmambamedical, nasiri2024vim4path}, registration and reconstruction \cite{huang2024mambamirarbitrarymaskedmambajoint, zheng2402fdvisionmamba, li2024mambadfuse} where long range dependency and high complexity is involved. Mamba has performed promisingly well in the biomedical field, especially in the fields of biomedical imaging, genomics and processing clinical notes. Thus, the model comes in handy in capturing long range and multi-modal data oriented tasks which involve subtle relationships and dependencies between units of information. \autoref{fig:timeline} shows the timeline of Mamba's evolution over time, starting from HiPPO \cite{NEURIPS2020_102f0bb6} and SSMs such as Linear State Space Layer (LSSL) \cite{gu2021combiningrecurrentconvolutionalcontinuoustime}, 
S4 \cite{gu2021efficiently}, Diagonal State Space (DSS) \cite{gupta2022diagonalstatespaceseffective} , S4D \cite{gu2022parameterizationinitializationdiagonalstate}, S5 \cite{smith2023simplifiedstatespacelayers}, S4ND \cite{nguyen2022s4ndmodelingimagesvideos}, Hungry Hungry Hippos (H3) \cite{fu2023hungryhungryhipposlanguage} to Mamba \cite{gu2024mambalineartimesequencemodeling}. It also includes variants of Mamba that were created as the model evolved.

\begin{figure}[!ht]
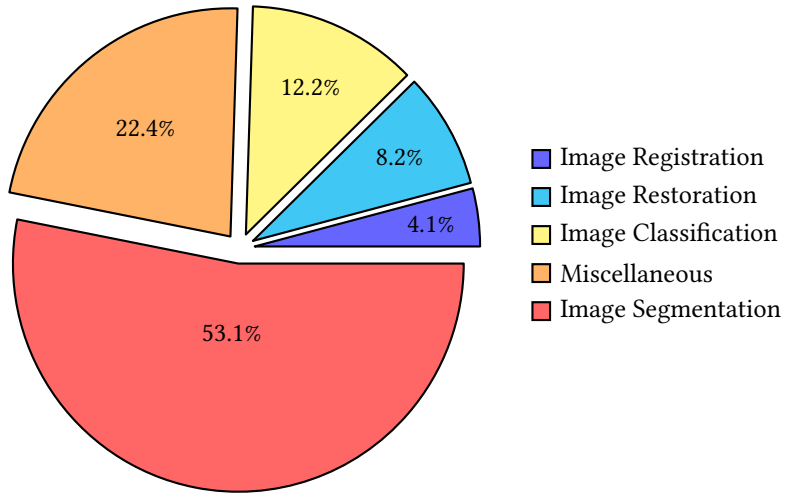

\centering
\sbox0{\tikz{\pie[sum=auto]{100/}}}
\sbox2{\tikz{\pie[sum=auto,text=legend, after number = {\%}, ,explode=0.2]{4.1/Image Registration,8.2/Image Restoration,12.2/Image Classification ,22.4/Miscellaneous, 53.1/Image Segmentation}}}
\makebox[\wd0][l]{\box2}
 \caption{Distribution of Research Papers Utilizing Mamba in Medical Domain}
 \vspace*{1em}
\label{fig:pie}
\end{figure}

\begin{figure}
    \centering
    \includegraphics[width=\linewidth]{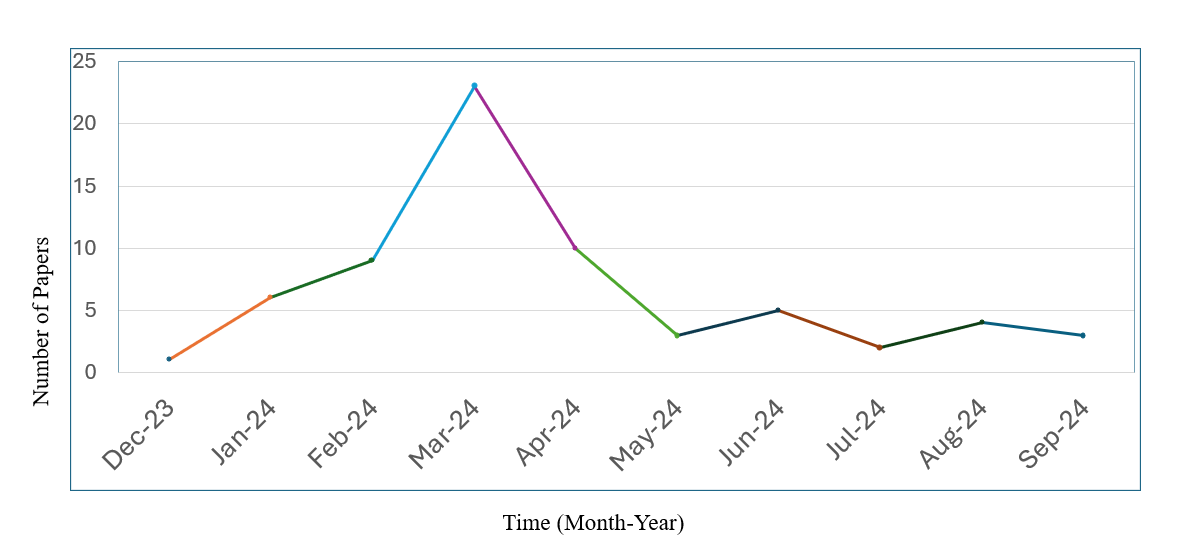}
    \caption{Publication Trend of Mamba Research in the Medical Domain}
    \label{fig:line_chart}
\end{figure}

The pie chart, as depicted in \autoref{fig:pie} illustrates the distribution of research papers utilizing the Mamba framework across various tasks in the medical domain. The chart is divided into five segments, each representing a specific task and its corresponding percentage contribution to the total number of papers. Moreover, \autoref{fig:line_chart} illustrates the fluctuation in the number of publications related to Mamba in the medical domain during the period from December 2023 to September 2024. A notable surge in research activity is evident in March and April 2024.

Existing surveys in this domain either cover the framework broadly \cite{patro2024mamba, qu2024survey} or are restricted to its use only in the vision domain \cite{xu2024survey, liu2024vision, zhang2024survey}. 
It is worth noting that only \cite{heidari2024computation} provided a review of Mamba which was focused on its uses in the medical domain. However, our survey paper is more extensive and detailed than that of \cite{heidari2024computation}. In particular, the present work emphasizes the analysis of public resources, such as medical datasets and presents some empirical data on the applicability of Mamba within medical practice, including various resources and interventions to be utilized within Mamba in a healthcare context. Beyond this, we incorporate recent advances in self-supervised and multimodal learning within Mamba architectures, demonstrating their potential to improve medical image analysis beyond traditional supervised approaches. We also address practical challenges related to deploying Mamba models in real-world clinical settings, including hardware limitations and generalization across diverse medical datasets.
In addition to this, we include the latest research and developments of Mamba architectures for medical image analysis. 
Furthermore, we explore future directions aimed at improving computational efficiency, adapting scanning mechanisms for non-causal visual data, and expanding Mamba’s usability in real-time and edge computing scenarios.
Moreover, we have structured our work in such a way that readers will appreciate the organizational spans of Mamba including its strengths, weaknesses, and prospects within the medical arena. \autoref{fig:taxonomy} displays the taxonomy of this survey, providing a comprehensive overview of the topics covered.

\begin{figure}
    \centering
    \fontsize{6.8}{1}\selectfont
\tikzset{
    basic/.style  = {draw, text width=2.5cm, align=center, font=\sffamily, rectangle},
    root/.style   = {basic, rounded corners=2pt, thin, align=center, fill=green!30},
    onode/.style = {basic, thin, rounded corners=2pt, align=center, fill=green!60, text width=2.5cm},
    tnode/.style = {basic, thin, align=left, fill=pink!60, text width=36.5em},
    xnode/.style = {basic, thin, rounded corners=2pt, align=center, fill=blue!20, text width=2cm},
    wnode/.style = {basic, thin, align=left, fill=pink!10!blue!80!red!10, text width=6em},
    edge from parent/.style={draw=black, edge from parent fork right}
}

\begin{forest} for tree={
    grow=east,
    growth parent anchor=west,
    parent anchor=east,
    child anchor=west,
    edge path={\noexpand\path[\forestoption{edge},->, >={latex}] 
         (!u.parent anchor) -- +(10pt,0pt) |- (.child anchor)
         \forestoption{edge label};}
}
[A Comprehensive Survey of Mamba Architectures for
Medical Image Analysis: Classification{,} Segmentation{,} Restoration and Beyond, basic, l sep=8mm,
    [{\hyperref[sec:conclusion]{Conclusion}}, xnode, l sep=8mm,
        [{\hyperref[subsec:confut]{Future directions}}, tnode]
        [{\hyperref[subsec:consum]{Key findings}}, tnode]
        [{\hyperref[subsec:consig]{Significance for the field}}, tnode] ]
    [{\hyperref[sec:discussion]{Discussion}}, xnode, l sep=8mm,
        [{\hyperref[subsec:emerge]{Emerging Areas}}:\\Mamba 2\cite{dao2024transformersssmsgeneralizedmodels}{,} xLSTM\cite{beck2024xlstmextendedlongshortterm}, tnode]
        [{\hyperref[subsec:limit]{Limitations}}, tnode] ]
    [{\hyperref[sec: datasets]{Datasets}}, xnode, l sep=8mm, 
        [BraTS2023\cite{4b589b6824a64a2a91e8e3b26cc0bf9e, kazerooni2024braintumorsegmentationpediatrics, Bakas2017AdvancingTC}{,} AIIB2023\cite{nan2022fuzzyattentionneuralnetwork, Li2022HumanTT}{,} CRC-500\cite{xing2024segmambalongrangesequentialmodeling}{,} ISIC2017\cite{8363547}{,} 
        ISIC2018\cite{codella2019skinlesionanalysismelanoma, Tschandl_2018}{,} CVC-ClinicDB\cite{PMID:25863519}{,} 3D Abdomen CT\cite{FLARE22}{,} 2D Abdomen MRI\cite{ji2022amos}{,} Endoscopy\cite{allan20192017roboticinstrumentsegmentation}{,} Microscopy\cite{Ma_2024}{,} ACDC MRI Cardiac\cite{8360453}{,} Synapse Multi-Organ Abdominal CT{\footnotemark[5]}{,} LiTS\cite{BILIC2023102680}{,} Montgomery and Shenzhen\cite{article}{,} Brain MRI Multiple Sclerosis\cite{article4}{,} ADNI\cite{e3cc3721ea544fa6b325eab29573bdff, Lian2020HierarchicalFC}{,} Otoscopy\cite{zeng2021efficient}{,} PathMNIST\cite{medmnistv2, medmnistv1}{,} CAMELYON16\cite{camelyon, 15b880f0e9424a5eb5cf74f6fc22f28a}{,} Colorectal Cancer Histopathology\cite{Loughrey2020Colorectal}{,} SynthRAD\cite{Guo2024MambaMorphAM}{,} FastMRI\cite{zbontar2018fastmri}{,} Low-Dose CT Image and Projection\cite{moen2021low}, tnode]]
    [{\hyperref[sec:applications_various_domains]{Applications in Various Medical Domains}}, xnode, l sep=8mm, 
        [{\hyperref[subsec:mis]{Miscellaneous}}:\\
        Prompt-Mamba\cite{xie2024promamba}{,} ClinicalMamba\cite{yang2024clinicalmamba}{,} P-Mamba\cite{ye2024pmambamarryingperonamalik}{,}  MD-Dose\cite{fu2024mddose}{,} Vivim\cite{yang2024vivimvideovisionmamba}{,} VM-DDPM\cite{ju2024vmddpmvisionmambadiffusion}, tnode]
        [{\hyperref[subsec:reg]{Medical Image Registration}}:\\
        MambaMorph\cite{Guo2024MambaMorphAM}{,} VMambaMorph\cite{Wang2024VMambaMorphAM}, tnode]
        [{\hyperref[subsec:res]{Medical Image Restoration/Reconstruction}}:\\
        FDVM-Net\cite{zheng2402fdvisionmamba}{,} MambaMIR\cite{huang2024mambamirarbitrarymaskedmambajoint}{,} MambaDFuse\cite{li2024mambadfuse}{,} FusionMamba\cite{xie2024fusionmambadynamicfeatureenhancement}, tnode]
        [{\hyperref[subsec:clas]{Medical Image Classification}}:\\
        nnMamba\cite{gong2024nnmamba}{,} MedMamba\cite{yue2024medmambavisionmambamedical}{,} Vim4path\cite{nasiri2024vim4path}{,} Microscopic-Mamba\cite{zou2024microscopicmambarevealingsecretsmicroscopic}, tnode]
        [{\hyperref[subsec:seg]{Medical Image Segmentation}}:\\
        SegMamba\cite{xing2024segmambalongrangesequentialmodeling}{,} H-Vmunet\cite{wu2024hvmunethighordervisionmamba}{,} LKM-UNet\cite{wang2024lkmunetlargekernelvision}{,} 
        Mamba-UNet\cite{wang2024mambaunet}{,} Weak Mamba UNet\cite{wang2024weakmambaunetvisualmambamakes}{,} LightM-UNet\cite{liao2024lightmunetmambaassistslightweight}{,} UltraLight-VM UNet\cite{wu2024ultralightvmunetparallelvision}{,} T-Mamba\cite{hao2024tmamba}{,} HC-Mamba\cite{xu2024hcmambavisionmambahybrid}{,} Semi-Mamba-UNet\cite{ma2024semimambaunetpixellevelcontrastivepixellevel}{,} 
        ViM-UNet\cite{archit2024vimunet}{,} U-Mamba\cite{Ma2024UMambaEL}{,} Mamba-HUNet\cite{sanjid2024integratingmambasequencemodel}{,} UU-Mamba\cite{tsai2024uumamba}{,} CAMS-Net\cite{khan2024camsconvolutionattentionfreemambabased}{,} MUCM-Net\cite{yuan2024mucmnetmambapowereducmnet}, tnode]]
    [{\hyperref[sec:techniques_adaptations]{Techniques and Adaptations}}, xnode, l sep=8mm, 
        [{\hyperref[subsec:multi]{Multimodal Learning}}{:} Fusion Mamba\cite{xie2024fusionmambadynamicfeatureenhancement}{,} 
         MGI\cite{zhou2024mgimultimodalcontrastivepretraining}{,} GFE-Mamba\cite{fang2024gfemambamambabasedadmultimodal}, tnode]
        [{\hyperref[subsec:self]{Self Supervised Learning}}{:} Vim4Path\cite{nasiri2024vim4path}{,} MGI\cite{zhou2024mgimultimodalcontrastivepretraining}{,} MambaMIM\cite{tang2024mambamimpretrainingmambastate}{,} CMViM\cite{yang2024cmvim}, tnode]
        [{\hyperref[subsec:semi]{Semi-Supervised Learning}}{:} Semi-Mamba-UNet\cite{ma2024semimambaunetpixellevelcontrastivepixellevel}, tnode]
        [{\hyperref[subsec:weakly]{Weakly Supervised Learning}}{:} Weak-Mamba-UNet\cite{wang2024weakmambaunetvisualmambamakes}
            , tnode]]
    [{\hyperref[sec:optimizations]{Mamba Optimizations}}, xnode, l sep=8mm,
        [{\hyperref[subsec:light]{Lightweight and Efficient}}{:}\\
        LightM UNet\cite{liao2024lightmunetmambaassistslightweight}{,} UltraLight VM-UNet\cite{wu2024ultralightvmunetparallelvision}{,} MUCM-Net\cite{yuan2024mucmnetmambapowereducmnet}{,}
        LightCF-Net\cite{bioengineering11060545}{,}  MiM-ISTD\cite{Chen2024MiMISTDMF}, tnode] ]
    [{\hyperref[sec:scanning]{Scanning}}, xnode, l sep=8mm,
        [BiDirectional Scan \cite{zhu2024vision}{,} Selective Scan 2D \cite{liu2024vmamba}{,} Spatiotemporal Selective Scan \cite{yang2024vivimvideovisionmamba}{,} Zigzag Scan \cite{hu2024zigmaditstylezigzagmamba}{,} Local Scan \cite{huang2024localmambavisualstatespace}{,} Efficient 2D Scan \cite{Pei2024EfficientVMambaAS}{,} Continuous 2D Scan \cite{yang2024plainmambaimprovingnonhierarchicalmamba}{,} Three Directional Scan \cite{xing2024segmambalongrangesequentialmodeling}{,} Pixelwise and Patchwise Scan \cite{wang2024lkmunetlargekernelvision}{,} Omnidirectional Selective Scan \cite{shi2024vmambairvisualstatespace}{,} Hierarchical Scan \cite{zhang2024motionmambaefficientlong}{,} Multi-Head Scan \cite{ji2024mhsvmmultiheadscanningparallel}{,} Multi-Path Scan \cite{chen2024rsmambaremotesensingimage}{,} 3D BiDirectional Scan \cite{li2024videomambastatespacemodel}{,} Bidirectional Slice Scan \cite{fan2024slicemambaneuralarchitecturesearch}, tnode] ]
    [{\hyperref[sec:mamba_architectures]{Mamba architectures}}, xnode, l sep=8mm,
        [{\hyperref[subsec:hybrid]{Hybrid Architectures}}:\\
        Mamba with convolution: U-Mamba\cite{Ma2024UMambaEL}{,} Mamba-U-Net\cite{wang2024mambaunet}{,} HC-Mamba\cite{xu2024hcmambavisionmambahybrid}{,} TS-Mamba\cite{he2015deep}{,} nnMamba\cite{gong2024nnmamba}{,} Swin-UMamba\cite{liu2024swinumambamambabasedunetimagenetbased}\\Mamba with Attention and Transformers: Weak-Mamba-UNet\cite{wang2024weakmambaunetvisualmambamakes}{,} TP-Mamba\cite{wang2024triplanemambaefficientlyadapting}{,} HMT-UNet\cite{zhang2024hmtunethybirdmambatransformervision}\\ Mamba with Recurrence: VMRNN\cite{Tang2024VMRNNIV} \\Miscellaneous: Vim4Path\cite{nasiri2024vim4path}{,} MambaMIL\cite{yang2024mambamil}, tnode]
        [{\hyperref[subsec:variants]{Variants of U-Net}}:\\
        VM-UNet\cite{ruan2024vmunetvisionmambaunet}{,} ViM-UNet\cite{archit2024vimunet}{,} VM-UNet-V2\cite{zhang2024vmunetv2rethinkingvisionmamba}{,} LKM-UNet\cite{wang2024lkmunetlargekernelvision}{,} H-vmunet\cite{wu2024hvmunethighordervisionmamba}, tnode] 
        [{\hyperref[subsec:puremamba]{Pure Mamba}}:\\
        ViM\cite{zhu2024vision}{,} VMamba\cite{liu2024vmamba}{,} Plain Mamba\cite{yang2024plainmambaimprovingnonhierarchicalmamba}{,} EVSS\cite{Pei2024EfficientVMambaAS}{,} MambaND\cite{Li2024MambaNDSS}{,} MHS-VM\cite{ji2024mhsvmmultiheadscanningparallel}{,} Local Mamba\cite{huang2024localmambavisualstatespace}{,} Weak-Mamba-UNet\cite{wang2024weakmambaunetvisualmambamakes}{,} TMamba\cite{hao2024tmamba}{,} H-vmunet\cite{wu2024hvmunethighordervisionmamba}{,} UltraLight VM-UNet\cite{wu2024ultralightvmunetparallelvision}{,} Light M-UNet\cite{liao2024lightmunetmambaassistslightweight}{,} SegMamba\cite{xing2024segmambalongrangesequentialmodeling}{,} LKM-UNet\cite{wang2024lkmunetlargekernelvision}
            , tnode] ]
    [{\hyperref[sec:key_terms]{Core Concepts of SSM}}, xnode, l sep=8mm,
        [{\hyperref[subsec:s6]{Selective Structured State Space Models (S6)}} \cite{gu2024mambalineartimesequencemodeling}, tnode]
        [{\hyperref[subsec:s5]{Simplified State Space Layers for Sequence Modeling (S5)}} \cite{smith2023simplifiedstatespacelayers}, tnode]
        [{\hyperref[subsec:s4]{Structured State Space Sequence Models (S4)}} \cite{gu2021efficiently}, tnode]
        [{\hyperref[subsec:ssm]{State Space Models}}, tnode] ]
]
\end{forest}

\caption{Structural Taxonomy of Survey Content}
\label{fig:taxonomy}
\end{figure}

In this survey, we focus on use, methods and problems of Mamba state space models within the medical domain. We provide a complete overview of the current state of development of this direction, focusing on determining the advantages and disadvantages of the Mamba models, as well as their future prospects. The rest of the paper is organized as follows. Section \ref{sec:key_terms} discusses the key terms related to SSM, different Mamba architectures are explained in Section \ref{sec:mamba_architectures}. Several Mamba optimizations are discussed in Section \ref{sec:optimizations}. Several techniques such as weakly supervised, semi-supervised, self supervised, contrastive learning, and multimodal learning are explained in Section \ref{sec:techniques_adaptations}. Different scanning mechanisms in Mamba are discussed in Section \ref{sec:scanning}, several applications in different domains are explained in Section \ref{sec:applications_various_domains}. Datasets are summarized in Section \ref{sec: datasets}. Experimental results showing Mamba performance across different tasks are discussed in Section \ref{sec:experiments}. Limitations and emerging areas are explained in Section \ref{sec:discussion}, finally we conclude the work by giving future directions in Section \ref{sec:conclusion}.

\addtocontents{toc}{\setcounter{tocdepth}{2}}
\section{Core Concepts of SSM} \label{sec:key_terms}
In the realm of deep learning, Transformers have consistently dominated in both Computer Vision (CV) and Natural Language Processing (NLP) tasks. The self-attention \cite{DBLP:journals/corr/VaswaniSPUJGKP17} mechanism within Transformers has greatly improved the understanding of these modalities by generating an attention matrix from the query, key, and value vectors. While the attention matrix is beneficial, it suffers from quadratic time complexity. The recent advancements, such as FlashAttention by Dao \textit{et al.} \cite{dao2022flashattentionfastmemoryefficientexact,dao2023flashattention2fasterattentionbetter} and linear attention \cite{katharopoulos2020transformersrnnsfastautoregressive}, have addressed this issue by reducing the time complexity. For instance, in linear attention, the key is multiplied by the value instead of the query, and the softmax function is replaced with a similarity function. Mamba developed by Gu \textit{et al.} \cite{gu2024mambalineartimesequencemodeling} further mitigates this problem by transforming the quadratic time complexity into linear time complexity in a recurrent manner. Mamba is the first model without attention to match the performance of a very strong Transformer. The core concepts of Mamba and its derivation from SSM  are explained in the following sections.

\subsection{State Space Models}  \label{subsec:ssm}

State Space Models (SSMs) uses an approach similar to Kalman filter \cite{kalmanrudolphemil}. SSMs convert a one-dimensional input sequence $u(t)$ into an N-dimensional continuous latent state $x(t)$, which is then projected into a 1D output signal $y(t)$. The entire process of a state space model can be represented as shown in \autoref{eq:1} and \autoref{eq:2}:
\begin{align} \label{eq:1}
x'(t)=\textbf{\textit{A}}x(t)+\textbf{\textit{B}}u(t) \end{align}
\begin{align} \label{eq:2}
y(t)=\textbf{\textit{C}}x(t)+\textbf{\textit{D}}u(t)\end{align}Parameters $\textbf{\textit{A}}, \textbf{\textit{B}}, \textbf{\textit{C}}, \textbf{\textit{D}}$ are initialized differently in SSM models such as S4, S5 and S6. To apply a discrete input sequence $u(u_{0}, u_{1},...)$, instead of a continuous function $u(t)$ , the sequence should be discretized using a parameter $\Delta$, known as the step size. This process also involves discretizing the parameters $(\textbf{\textit{A}}, \textbf{\textit{B}}, \textbf{\textit{C}}, \textbf{\textit{D}})$. In the following sections, we discuss each model and its specific discretization steps.
\subsection{Structured State Space Sequence Models (S4)} \label{subsec:s4}

S4 proposed by Gu \textit{et al.} \cite{gu2021efficiently} demonstrates how to efficiently compute all forms of the SSM: the recurrent representation, and the convolutional representation. Additionally, S4 employs a bilinear method for discretizing the parameters of the SSM, converting the state space parameter $\textbf{\textit{A}}$ into an approximation $\overline{\textbf{\textit{A}}}$. In S4, the parameter $D$ from the original state space model is either set to 0 or used as a residual connection. The discrete SSM can be expressed in its recurrent form as shown in \autoref{eq:3} and \autoref{eq:4}. The recurrent representation of the equation below can be scanned in parallel using a prefix sum operation, as the current input does not explicitly depend on the previous hidden layer. This makes parallel scan achievable, thereby reducing time complexity from linear to logarithmic.
\begin{align} \label{eq:3}
x_{k}=\overline{\textbf{\textit{A}}}x_{k-1}+\overline{\textbf{\textit{B}}}u_{k}\;\;\;\;\;\overline{\textbf{\textit{A}}}=(\mathbf{I}-\Delta/2\cdot\textbf{\textit{A}})^{-1}(\mathbf{I}+\Delta/2\cdot\textbf{\textit{A}})\end{align}
\begin{align} \label{eq:4}
y_{k}=\overline{\textbf{\textit{C}}}x_{k}\;\;\;\;\;\;\;\overline{\textbf{\textit{B}}}=(\mathbf{I}-\Delta/2\cdot\textbf{\textbf{\textit{A}}})^{-1}(\Delta\textbf{\textit{B}})\;\;\;\;\;\;\;\overline{\textbf{\textit{C}}}=\textbf{\textit{C}}\end{align}

Unrolling the equation above leads us to the convolutional aspect of S4 in \autoref{eq:5} \& \autoref{eq:6}.
\begin{align} \label{eq:5}
x_{0}=\overline{\textbf{\textit{B}}}u_{0}\;\;\;\;\;\;x_{1}=\overline{\textbf{\textit{AB}}}u_{0}+\overline{\textbf{\textit{B}}}u_{1}\;\;\;\;\;\;x_{2}=\overline{\textbf{\textit{A}}}^{2}\overline{\textbf{\textit{B}}}u_{0}+\overline{\textbf{\textit{AB}}}u_{1}+\overline{\textbf{\textit{B}}}u_{2}\end{align}
\begin{align} \label{eq:6}
y_{0}=\overline{\textbf{\textit{CB}}}u_{0}\;\;\;\;\;\;y_{1}=\overline{\textbf{\textit{CAB}}}u_{0}+\overline{\textbf{\textit{CB}}}u_{1}\;\;\;\;\;\;y_{2}=\overline{\textbf{\textit{CA}}}^{2}\overline{\textbf{\textit{B}}}u_{0}+\overline{\textbf{\textit{CAB}}}u_{1}+\overline{\textbf{\textit{CB}}}u_{2}\end{align}
This can be vectorized into a convolution as shown in \autoref{eq:7} \& \autoref{eq:8} with an explicit formula for the convolution kernel as shown in \autoref{eq:9}.
\begin{align} \label{eq:7}
y_{k}=\overline{\textbf{\textit{CA}}}^{k}\overline{\textbf{\textit{B}}}u_{0}+ \overline{\textbf{\textit{CA}}}^{k-1}\overline{\textbf{\textit{B}}}u_{1}+ ......+ \overline{\textbf{\textit{CAB}}}u_{k-1}+\overline{\textbf{\textit{CB}}}u_{k}\end{align}
\begin{align} \label{eq:8}
y = \overline{\textbf{\textit{K}}} * u.\end{align}
\begin{align} \label{eq:9}
\overline{\textbf{\textit{K}}}\;\in\;\mathbb{R}^{L}:=\kappa_{L}(\overline{\textbf{\textit{A}}},\overline{\textbf{\textit{B}}},\overline{\textbf{\textit{C}}}):=(\overline{\textbf{\textit{CA}}}^{i}\overline{\textbf{\textit{B}}})_{i\in[L]}=(\overline{\textbf{\textit{CB}}},\overline{\textbf{\textit{CAB}}},...,\overline{\textbf{\textit{CA}}}^{L-1}\overline{\textbf{\textit{B}}}).\end{align}
\autoref{eq:7} \& \autoref{eq:8} represents a single convolution, and $K$ is referred as the SSM convolution kernels or filters. The parameters of S4 are initialized randomly, except for $A$. The $A$ parameter is initialized as a HiPPO Matrix, which is defined in \autoref{eq:10}:
\begin{align} \label{eq:10}
    \textbf{(HiPPO Matrix)}\;\;\;\;\;\;\;\;\;\;\textbf{\textit{A}}_{nk}=-\left\{\begin{matrix}(2n+1)^{1/2}(2k+1)^{1/2}\;\;if\;n>k\\n+1\;\;\;\;\;\;\;\;\;\;\;\;\;\;\;\;\;\;\;\;\;\;\;\;\;\;if\;n=k\\0\;\;\;\;\;\;\;\;\;\;\;\;\;\;\;\;\;\;\;\;\;\;\;\;\;\;\;\;\;\;\;\;if\;n<k\end{matrix}\right.
\end{align}
S4 addresses the limitations of transformers by implementing these strategies. This empowers SSMs to excel in tasks requiring long-range dependencies such as Path-X \cite{tay2021long} . In contrast, Transformers \cite{choromanski2021rethinking, pmlr-v119-katharopoulos20a} in Path-X exhibit accuracy below 50\% (worse than random guessing), whereas S4 achieves approximately 80\% accuracy.
\subsection{Simplified State Space Layers for Sequence Modeling (S5)} \label{subsec:s5}
S5 proposed by Smith \textit{et al.} \cite{smith2023simplifiedstatespacelayers}, extends from S4 with similar initialization conditions, but enhances the architecture by adopting a Multiple Input Multiple Output (MIMO) approach. Notably, S5 introduces a learnable time scale parameter ($\Delta$), replacing the fixed parameter used in S4. Parameters in S5 are discretized using the Zero Order Hold (ZOH) method as mentioned in \autoref{eq:11}, providing a refined parameter system compared to S4.

\begin{align} \label{eq:11}
\overline{\mathbf{\Lambda}}=e^{\mathbf{\Lambda}\Delta},\;\;\;\;\;\overline{\textbf{\textit{B}}}=\mathbf{\Lambda}^{-1}(\overline{\mathbf{\Lambda}}-\mathbf{I})\tilde{\textbf{\textit{B}}},\;\;\;\;\;\overline{\textbf{\textit{C}}}=\tilde{\textbf{\textit{C}}},\;\;\;\;\;\overline{\textbf{\textit{D}}}=\tilde{\textbf{\textit{D}}}.
\end{align}
In terms of computation, S5 employs a fully recurrent connection with parallel scanning , as detailed in \autoref{eq:3} and \autoref{eq:4}. The authors highlight that a smaller latent space employs SSM to do parallel scanning where an associative operation is used in between during offline settings. This characteristic positions S5 for both online and offline processing, highlighting its utility in recurrent tasks within the time domain.
\subsection{Selective Structured State Space Models (S6)} \label{subsec:s6}
S6 introduced by Gu \textit{et al.} \cite{gu2024mambalineartimesequencemodeling} initializes the parameter A from S4D which employs a diagonalized matrix structure for A. Extended from S5, S6 incorporates SSMs within the Mamba Architecture. S6 builds upon the foundational assumptions of previous SSMs by leveraging a projection of the input function (using a linear layer) for initializing parameters $B$ and $C$. Notably, S6 also applies this projection to the step size parameter, i.e., $\Delta$. To enhance computational efficiency, S6 implements faster recurrence connections using operations on the Static Random Access Memory (SRAM) of the GPU, with storage on the High Bandwidth Memory (HBM) similar to the principles outlined in FlashAttention mechanism by Dao \textit{et al.} \cite{dao2022flashattentionfastmemoryefficientexact} and FlashAttention-2 mechanism by Dao \textit{et al.} \cite{dao2023flashattention2fasterattentionbetter}.

\begin{minipage}{0.48\textwidth}
\begin{algorithm}[H]
    \centering
    \caption{SSM (S4)}\label{algorithm1}
    \begin{algorithmic}
      
        \State $\textbf{Input :}  \;\;x\;:\; (B,L,D)$
        \State $\textbf{Output :}  \;\;y\;:\; (B,L,D)$
    \end{algorithmic}
    \begin{algorithmic}[1]
        \State $\textbf{\textit{A}} :(D,N) \gets \text{Parameter}$ \hfill
        
        \Comment{Represents structured $N \times N$ matrix}
        \State $\textbf{\textit{B}} :(D,N) \gets \text{Parameter}$
        \State $\textbf{\textit{C}} :(D,N) \gets \text{Parameter}$
        \State $\mathbf{\Delta} :(D) \gets \mathbf{\tau}_{\Delta}(\text{Parameter})$
        \State $\overline{\textbf{\textit{A}}},\overline{\textbf{\textit{B}}}: (D,N) \gets $discretize$(\Delta, \textbf{\textit{A}}, \textbf{\textit{B}})$
        \State $\textbf{\textit{y}} \gets SSM(\overline{\textbf{\textit{A}}}, \overline{\textbf{\textit{B}}}, \textbf{\textit{C}})(x)$\hfill
        
        \Comment{Time-invariant: recurrence or convolution}
        \State \textbf{return}  \text{y}
    \end{algorithmic}
\end{algorithm}
\end{minipage}
\hfill
\begin{minipage}{0.48\textwidth}
\begin{algorithm}[H]
    \centering
    \caption{SSM + Selection (S6)}\label{algorithm2}
    \begin{algorithmic}
        \State $\textbf{Input :}  \;\;x\;:\; (B,L,D)$
        \State $\textbf{Output :}  \;\;y\;:\; (B,L,D)$
    \end{algorithmic}
    \begin{algorithmic}[1]
        \State $\textbf{\textit{A}} :(D,N) \gets \text{Parameter}$ \hfill
        
        \Comment{Represents structured $N \times N$ matrix}
        \State $\textbf{\textit{B}} :(B,L,N) \gets s_{\mathbf{B}}(x)$
        \State $\textbf{\textit{C}} :(B,L,N) \gets s_{\mathbf{C}}(x)$
        \State $\mathbf{\Delta} :(B,L,D) \gets \mathbf{\tau}_{\Delta}(\text{Parameter}+s_{\mathbf{\Delta}}(x))$
        \State $\overline{\textbf{\textit{A}}},\overline{\textbf{\textit{B}}}: (B,L,D,N) \gets $discretize$(\Delta, \textbf{\textit{A}}, \textbf{\textit{B}})$
        \State $\textbf{\textit{y}} \gets SSM(\overline{\textbf{\textit{A}}}, \overline{\textbf{\textit{B}}}, \textbf{\textit{C}})(x)$\hfill
        
        \Comment{Time-varying: recurrence (scan) only}
        \State \textbf{return}  \text{y}
    \end{algorithmic}
\end{algorithm}
\end{minipage}\\
\hfill
\hfill

\begin{figure}[!ht]
\centering
\includegraphics[width=\linewidth]{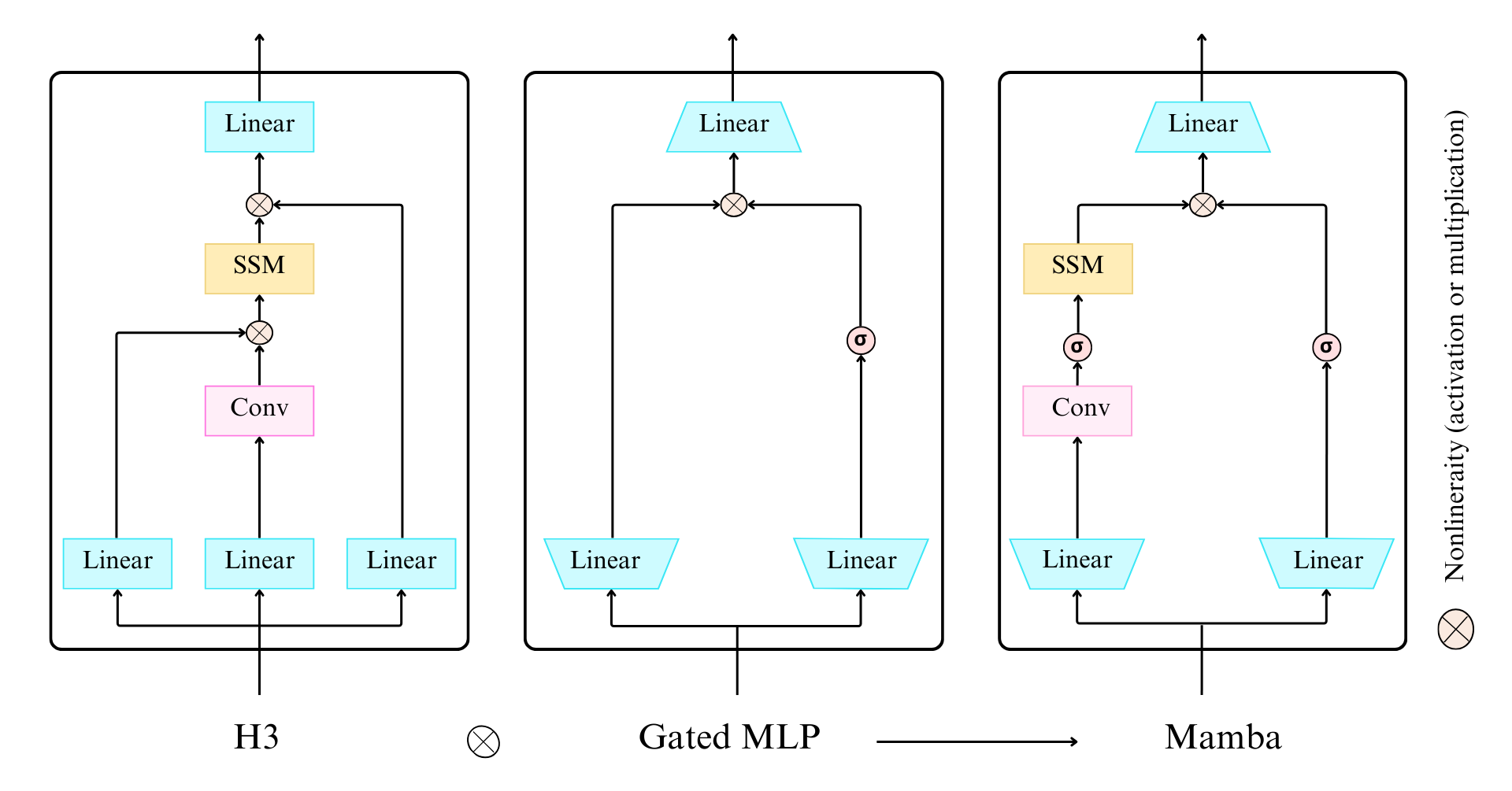}
\caption{Architecture of Mamba Block \cite{gu2024mambalineartimesequencemodeling} combining H3 \cite{fu2023hungryhungryhipposlanguage} and Gated MLP}
\label{fig:mamba}
\end{figure}

 The parameters in S6 are discretized using the Zero Order Hold method, following the approach set by S5. The Mamba architecture integrates components from both H3\cite{fu2023hungryhungryhipposlanguage} and Gated MLP, incorporating an additional SSM layer and connections resembling the green parallelograms found in H3. \autoref{algorithm1} \& \autoref{algorithm2} outlined above shows the differences between S4 and S6.
Mamba integrates a selective mechanism into its state space models (S6) to prioritize important content dynamically during training. \textbf{SSM + Selection (S6)} outlines a computational process for handling input sequences $x$ and generating corresponding output sequences $y$. It begins by initializing a structured matrix \textbf{\textit{A}} and projecting the input sequence $x$ into tensors \textbf{\textit{B}} and \textbf{\textit{C}} using specific functions $s_{\mathbf{B}}$ and $s_{\mathbf{C}}$.  The parameter $\mathbf{\Delta}$, serves as a time step parameter which is used in discretization. It is computed based on a function $\mathbf{\tau}$ incorporating additional parameters and $s_{\mathbf{\Delta}}$. Subsequently, \textbf{\textit{A}} and \textbf{\textit{B}} undergo discretization alongside $\mathbf{\Delta}$, resulting in transformed tensors $\overline{\textbf{\textit{A}}}$ and $\overline{\textbf{\textit{B}}}$ as mentioned below in \autoref{eq:12}.
\begin{align} \label{eq:12}
    \overline{\mathbf{A}} = exp(\Delta \mathbf{A})\;\;\;\;\;\;\;\;\;\; \overline{\mathbf{B}}=(\Delta \mathbf{A})^{-1}(exp(\Delta \mathbf{A})-\mathbf{I})\cdot \Delta \mathbf{B}
\end{align}
 The core of the algorithm involves applying a state space model (SSM) function to $x$ using ($\overline{\textbf{\textit{A}}}$, $\overline{\textbf{\textit{B}}}$, \textbf{\textit{C}}) facilitating a time-varying recurrence process (``scan''). This approach ensures that the output sequence $y$ reflects the transformations and interactions specified by SSM and selection mechanisms integrated within the architecture of S6. \autoref{fig:mamba} illustrates how selective components from H3 and Gated MLP are combined to construct the Mamba.

\addtocontents{toc}{\setcounter{tocdepth}{2}}
\section{Medical Image Analysis using Mamba}
In this section, we cover the categorization of literature related to Mamba architectures, explore optimizations that enhance their performance, and discuss various techniques and adaptations that expand their capabilities. Furthermore, we examine scanning techniques pertinent to Mamba and conclude by showcasing its diverse and impactful applications in the medical field.
\addtocontents{toc}{\setcounter{tocdepth}{3}}
\subsection{Mamba Architectures}\label{sec:mamba_architectures}
In this section, we explore and discuss the architectural landscape of Mamba, beginning with an exploration of the foundational pure Mamba design and its evolution through variants of U-Net. We then transition into the realm of hybrid architectures, where Mamba is ingeniously combined with other powerful techniques to achieve enhanced performance and tackle complex tasks.
\subsubsection{Pure Mamba}\label{subsec:puremamba}

\begin{figure}[!ht]
\centering
\begin{minipage}[b]{0.49\textwidth}
  \subfigure[Vision Mamba
    \label{fig:viM}
    ]{
        \includegraphics[width=7.5cm]{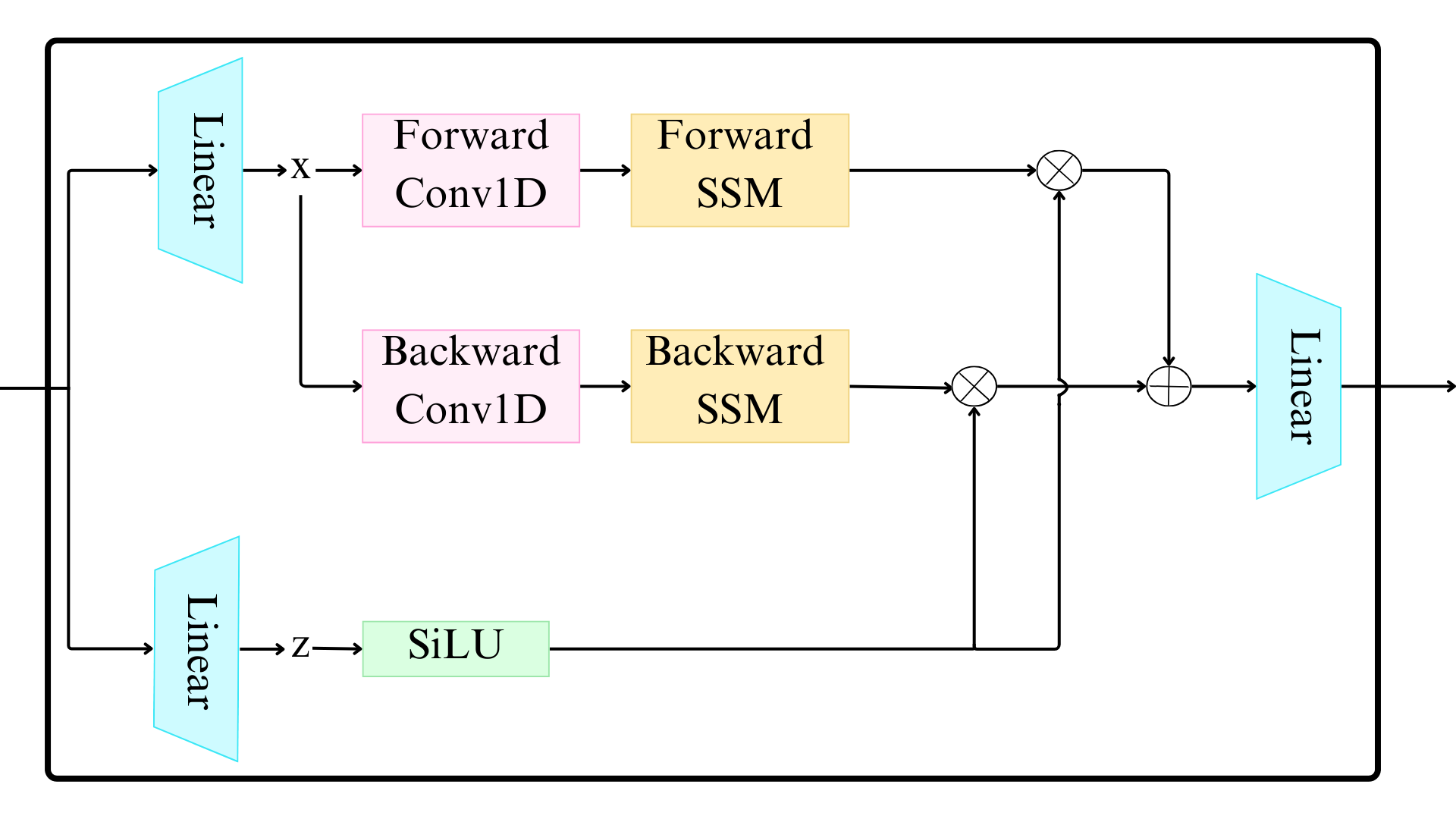}}
    \end{minipage}
\begin{minipage}[b]{0.49\textwidth}
    \subfigure[Visual State Space
    \label{fig:vss}
    ]{        \includegraphics[width=7.5cm]{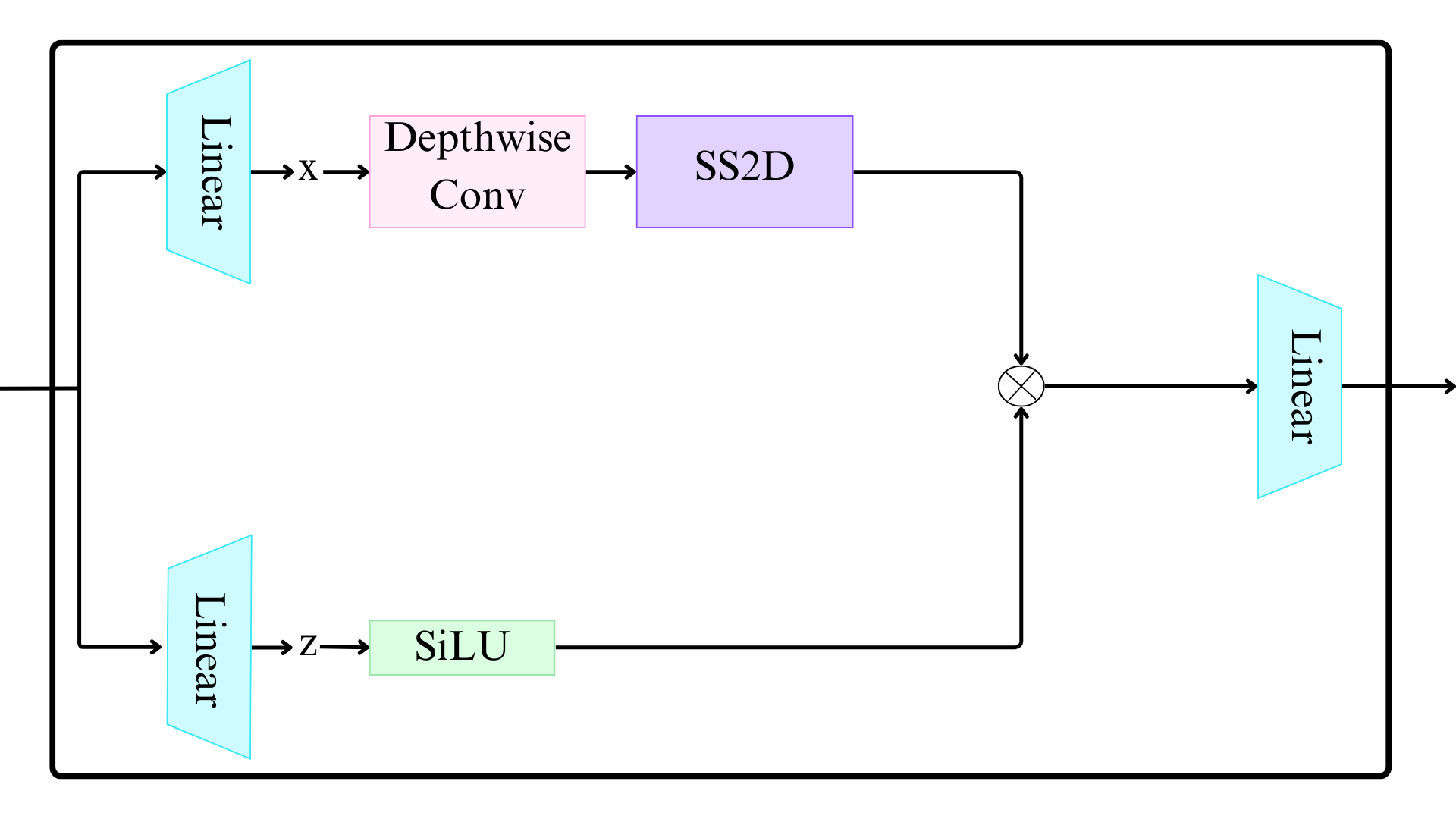}}
\end{minipage}
\caption{\centering Architectures of \autoref{fig:viM} Vision Mamba (ViM) \cite{zhu2024vision} and \autoref{fig:vss} Visual State Space (VSS) Block \cite{liu2024vmamba}}
\label{fig:visionmamba,vss}
\end{figure}

\textbf{Vision Mamba (ViM)} proposed by Zhu \textit{et al.} \cite{zhu2024vision} incorporates bidirectional SSM by combining convolution with S6 in both forward and backward directions. Moreover, softplus \cite{10.1007/s10489-017-1028-7} function is applied to the selective scan parameter $\Delta_{0}$ to make sure the parameter stays positive. The parameters of the S6 are discretized with the $\Delta_{0}$ parameter mentioned above. ViM employs a training strategy similar to Vision Transformers (ViT) \cite{dosovitskiy2021imageworth16x16words} where patches of inputs are tokenized by separating them into non-overlapping patches and applying a convolution layer on each patch with dimension $d$. The tokenized patches are then concatenated with class labels, and learnable position encoding are added to the class label and tokenzied patches. 
Overall, ViM shows comparable differences in memory and performance to parametric heavy models such as DeiT-Ti (Data-efficient image Transformers-Tiny), DeiT-S (Data-efficient image Transformers-Small) proposed by Touvron \textit{et al.} \cite{touvron2021trainingdataefficientimagetransformers} on tasks such as object detection, classification, and segmentation. Particularly in higher-dimensional images such as $1248 \times 1248$, ViM consumes 73.2\% less memory than DeiT and $2.8\times$ faster than DeiT. Figure \autoref{fig:viM} depicts the architecture of the ViM block, illustrating its key elements and their role in improving the model's performance in vision tasks.

\textbf{VMamba }proposed by Liu \textit{et al.} \cite{liu2024vmamba} employs an architecture similar to transformers, replacing the traditional multi-head attention block \cite{DBLP:journals/corr/VaswaniSPUJGKP17} with a novel approach. It utilizes a Visual State Space (VSS) block, which differentiates it from the standard Mamba architecture by incorporating depthwise convolution and SS2D (2D selective scan). This new implementation features a non-multiplicative branching method and replaces S6 used in Mamba with SS2D. While S6 performs well for NLP tasks, extending it to 2D vision data presents challenges in making S6 modules scan-independent. SS2D addresses this by implementing a gating mechanism, eliminating the need for branched multiplication as proposed in Mamba and ViM. In VMamba, patches are initially partitioned within the stem module, resulting in a feature map of size $H/4 \times W/4$. As the data progresses through the layers, the feature map dimensions change sequentially to $H/8 \times W/8$, $H/16 \times W/16$, and finally $H/32 \times W/32$, with \textbf{\textit{C}} representing the network's arbitrary dimensionality. Each stage, except the first, includes a downsampling block alongside VSS. Ultimately, a prediction head is employed to generate outputs for the designated task. SS2D consists of two stages: 1) cross scan module 2) cross merge module. The cross scan module in VMamba scans patches in four different directions: top to bottom, bottom to top, left to right and right to left. For each direction, an independent SSM is utilized, and the representations from these SSMs are combined using cross merging. On the performance perspective, VMamba achieves superior metrics with a minimal number of parameters and memory usage. Its performance is comparable to ViM-S, DeiT-S, and DeiT-B across various tasks, including semantic segmentation, classification, and object detection. Figure \autoref{fig:vss} depicts the architecture of the Visual State Space (VSS) block, showcasing its key components such as SS2D scanning and depthwise convolution.

\begin{figure}[!ht]
\centering
\begin{minipage}[b]{0.49\textwidth}
  \subfigure[Plain Mamba 
    \label{fig:plain}
    ]{
        \includegraphics[width=7.5cm]{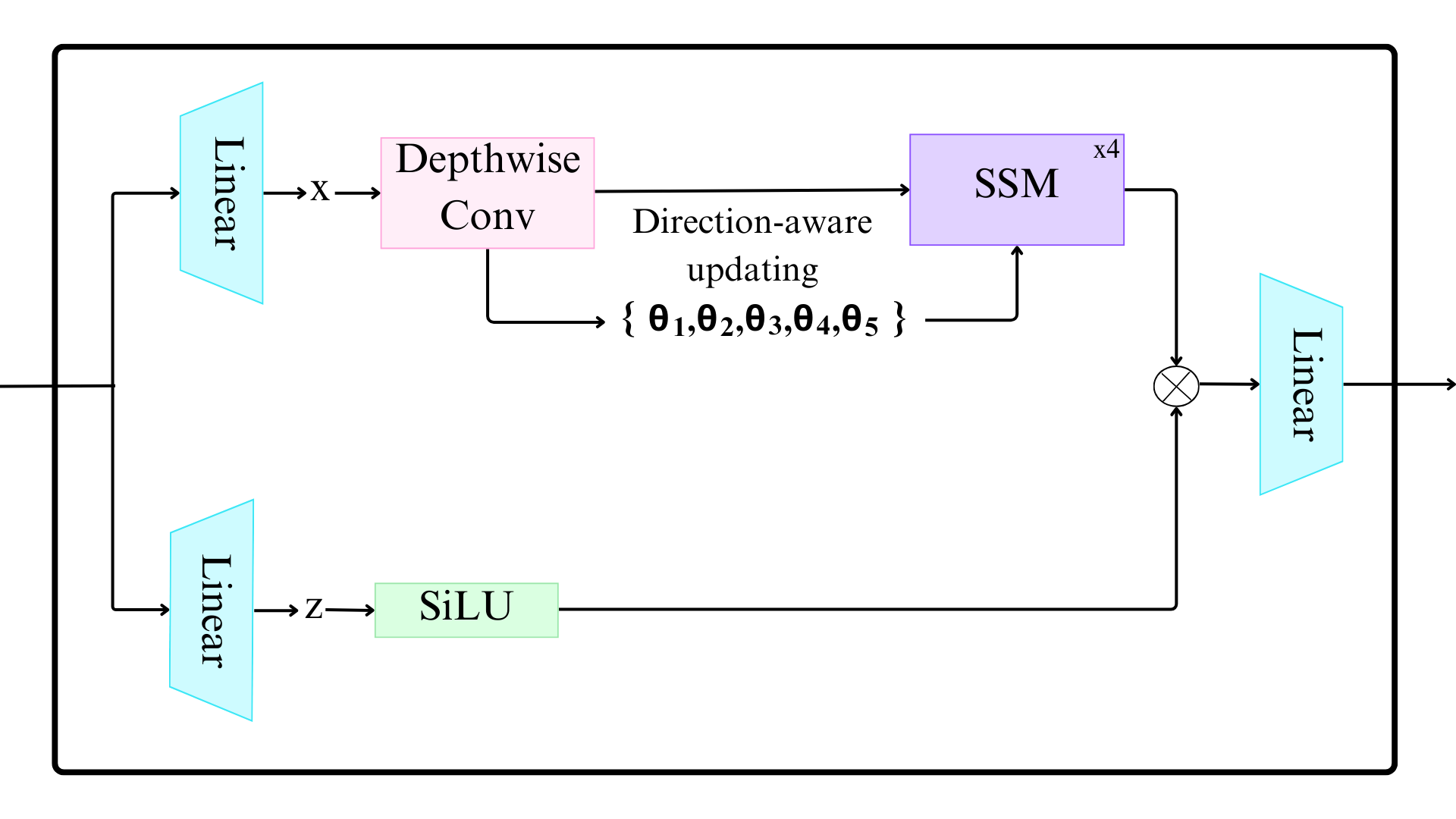}}
    \end{minipage}
\begin{minipage}[b]{0.49\textwidth}
    \subfigure[Efficient Visual State Space
    \label{fig:evss}
    ]{        \includegraphics[width=7.5cm]{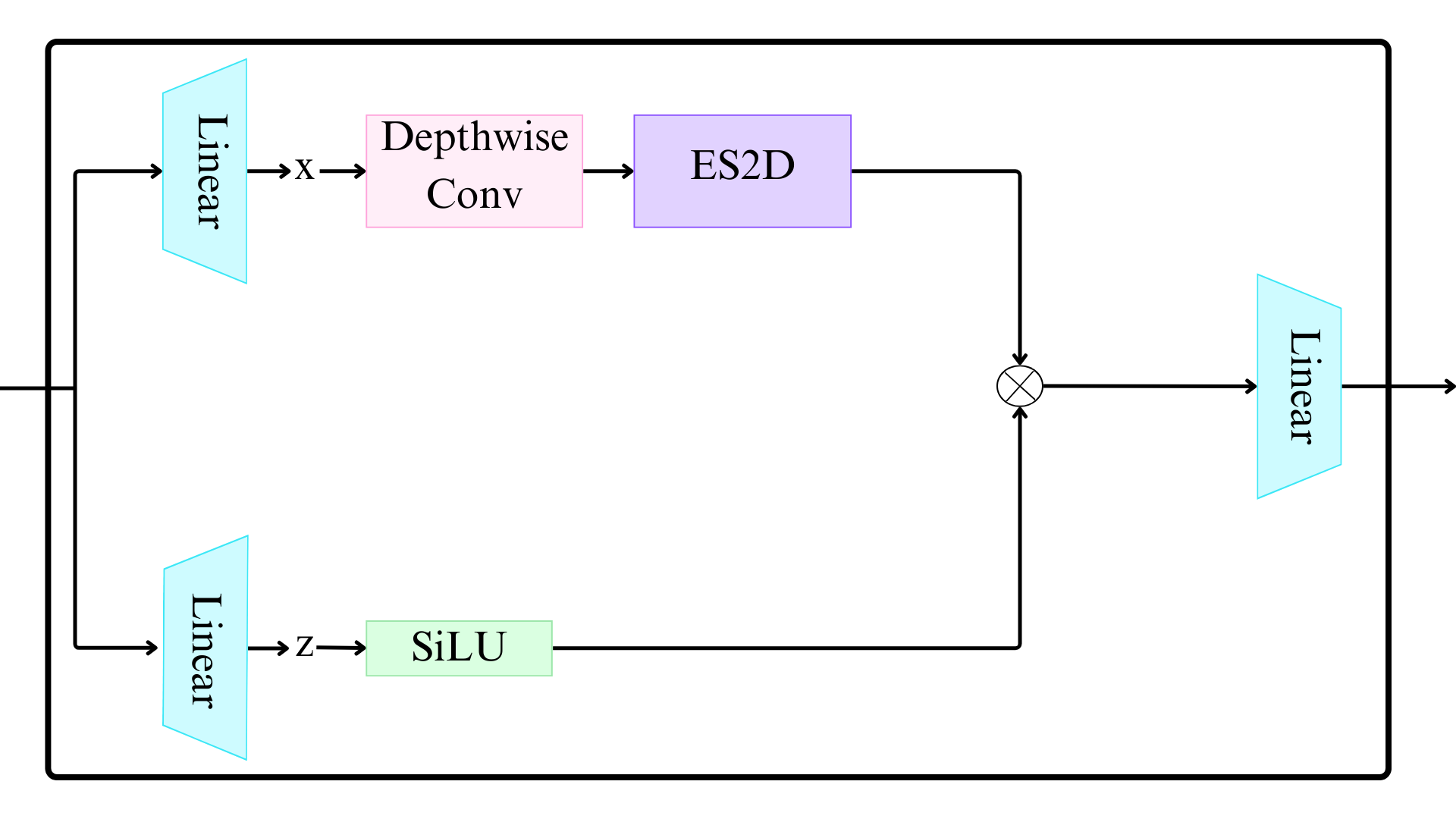}}
\end{minipage}
\caption{\centering Architectures of \autoref{fig:plain} Plain Mamba \cite{yang2024plainmambaimprovingnonhierarchicalmamba} and \autoref{fig:evss} Efficient Visual State Space (EVSS) Block \cite{Pei2024EfficientVMambaAS}}
\label{fig:planmamba,evss}
\end{figure}

\textbf{Plain Mamba} proposed by Yang \textit{et al.} \cite{yang2024plainmambaimprovingnonhierarchicalmamba} is a non-hierarchical SSM, similar to Vision Transformers (ViT). Drawing inspiration from ViT, Plain Mamba begins with patch embedding combined with position embedding, followed by a plain Mamba layer. In contrast to VSS, the plain Mamba layer utilizes gated multiplication of features, similar to ViM, but instead of a single SSM block, it employs four SSMs with continuous 2D scanning. The scan orders are the same as in VMamba, ensuring no positional bias and promoting uniform image understanding. Plain Mamba as shown in Figure \autoref{fig:plain}, also incorporates direction-aware updating by embedding 2D relative position information into a flattened 1D layer within the SSM. Unlike ViM, it uses global pooling, followed by a classification head on top.

\textbf{Extension of VMamba:} Pei \textit{et al.} \cite{Pei2024EfficientVMambaAS} proposed the \textbf{EVSS block}, which stands for \textbf{Efficient Visual State Space}, which combines local and global features effectively. EVSS block shown in Figure \autoref{fig:evss}, employs the ES2D module to extract the global feature map and a depth-wise convolutional branch to obtain the local feature map. It also incorporates a squeeze excitation block similar to SqueezeNet which is proposed by Iandola \textit{et al.} \cite{Iandola2016SqueezeNetAA}. The global and local features are then combined. ES2D introduces a novel method of patch-wise scanning. Initially, the input image is divided into patches belonging to different groups. These patches undergo forward and backward 2D scanning, similar to the VSS method, and are then passed to S6 and merged. In the overall architecture, EVSS blocks are utilized in the initial stages, while inverted residual blocks are employed in the later stages. In conclusion, EfficientVMamba achieves better performance with less number of parameters compared to transformer-based and convolution-based models.

\begin{figure}[!ht]
\includegraphics[width=\linewidth]{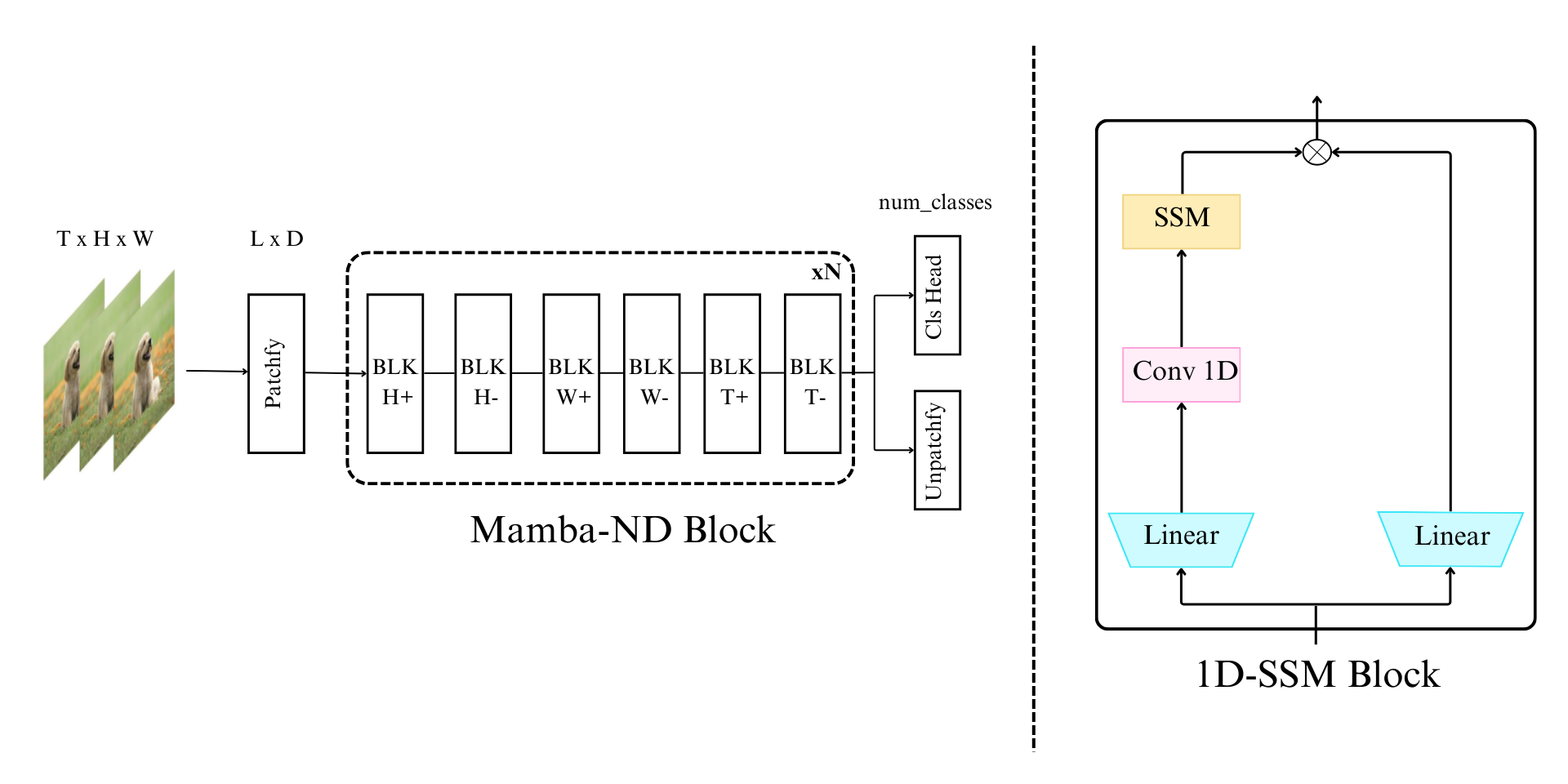}
\caption{\centering Architecture of Mamba ND Block \cite{Li2024MambaNDSS}}
\label{fig:mamband}
\centering
\end{figure}

Li \textit{et al.} \cite{Li2024MambaNDSS} introduced the \textbf{Mamba ND Block}, which evaluates different scanning methods for 2D and 3D image representations. They found that the most effective approach involves scanning forward and then backward across height, width, and volume/time (if applicable). Separate Mamba blocks are designated for each scanning direction. The Mamba ND models outperformed transformer-based models while utilizing fewer parameters, resulting in enhanced performance in tasks such as video classification. The architecture of the Mamba ND block is detailed in \autoref{fig:mamband}, showcasing its design and structural components. 

\begin{figure}[!ht]
\centering
\begin{minipage}[b]{0.49\textwidth}
  \subfigure[Local Mamba 
    \label{fig:local}
    ]{
        \includegraphics[width=7.5cm]{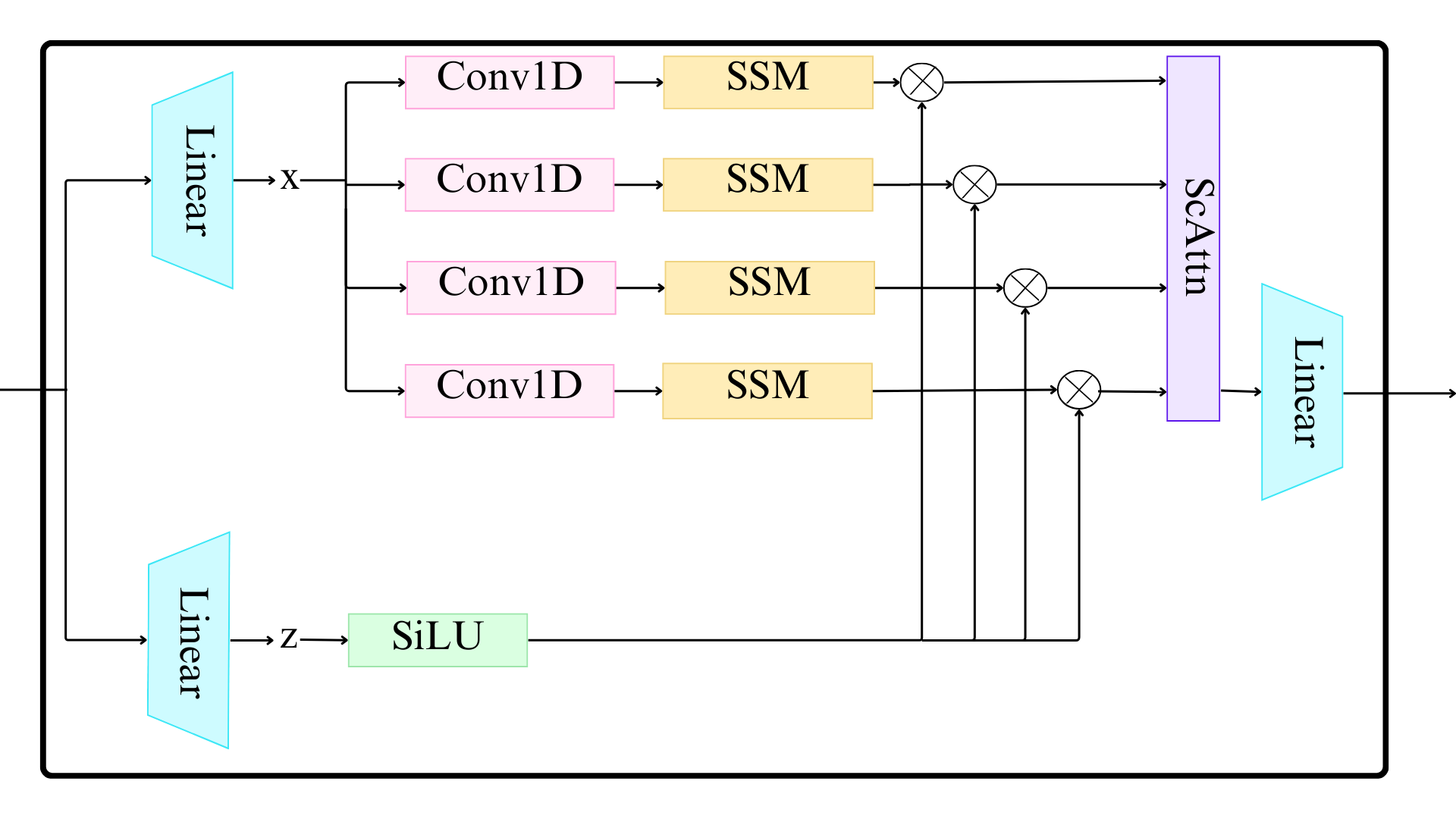}}
    \end{minipage}
\begin{minipage}[b]{0.49\textwidth}
    \subfigure[MHS-VM
    \label{fig:mhsvm}
    ]{        \includegraphics[width=7.5cm]{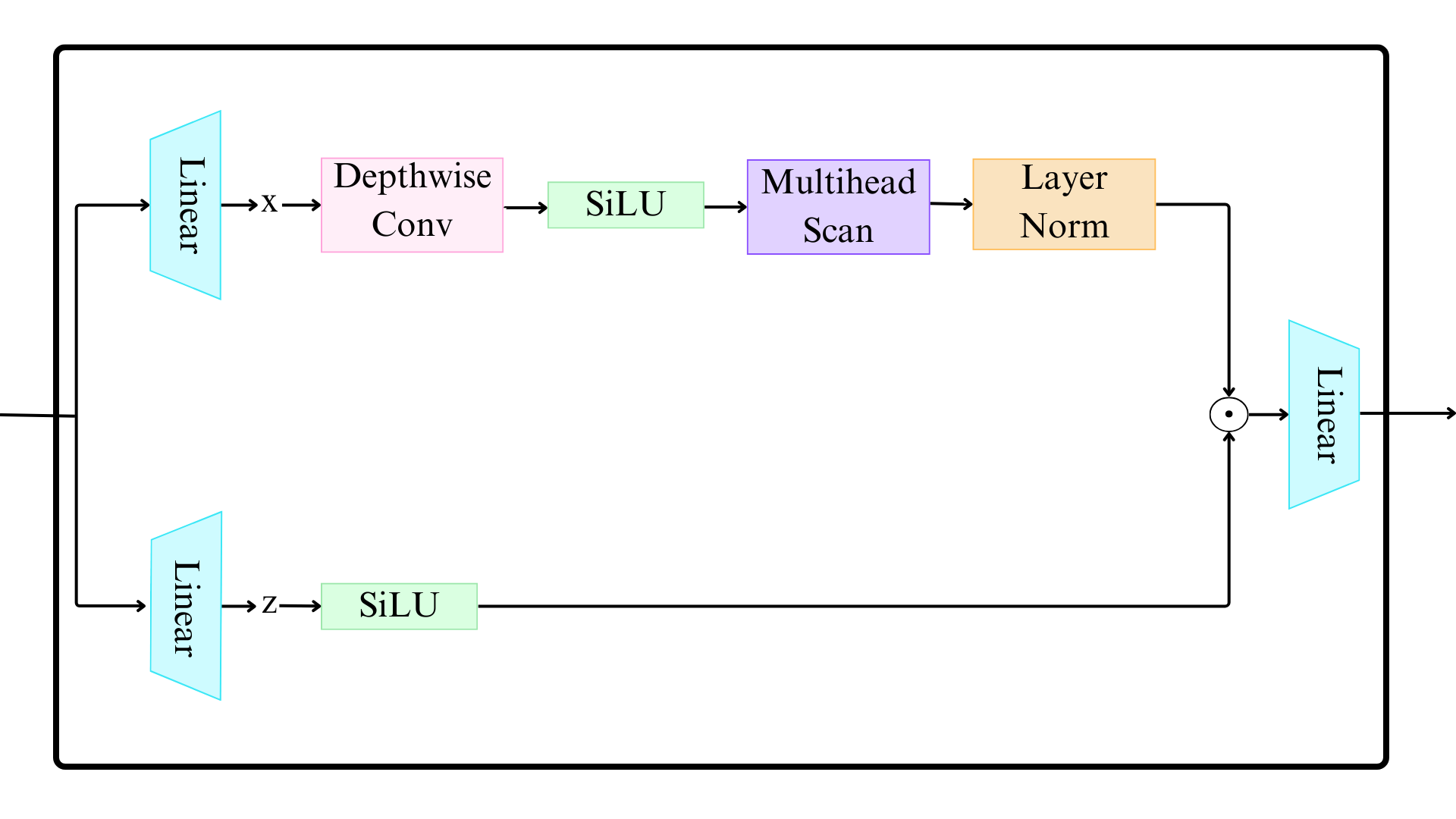}}
\end{minipage}
\caption{\centering Architectures of \autoref{fig:local} Local Mamba Block \cite{huang2024localmambavisualstatespace} and \autoref{fig:mhsvm} MHS Block \cite{ji2024mhsvmmultiheadscanningparallel}}
\label{fig:mhs,local}
\end{figure}

\textbf{MHS-VM} proposed by Ji \textit{et al.} \cite{ji2024mhsvmmultiheadscanningparallel}, introduced an organized approach to construct visual features within 2D image spaces using Multi-Head Scan (MHS) module. This module projects embeddings from the previous layer into multiple lower-dimensional subspaces, where selective scan is performed along distinct scan routes. The resulting sub-embeddings undergo integration and projection back into the high-dimensional space. Additionally, the module incorporates a Scan Route Attention (SRA) mechanism to improve its ability to discern complex structures. To validate its efficiency, the SS2D block replaces the original block in VM-UNet, resulting in significant performance improvements while reducing parameters. Figure \autoref{fig:local} illustrates the architecture of MHS-VM, detailing its use of components such as depthwise convolution, layer normalization, and multihead scanning.

\textbf{Extension of ViM:} Huang \textit{et al.} \cite{huang2024localmambavisualstatespace} proposed \textbf{Local Mamba}, which implements a novel scanning mechanism where tokens are partitioned into distinct windows, and scanning is performed continuously within each patch of these windows. Local Mamba consists of four SSM blocks with different scanning patterns: a $7\times 7$ local scan, a $2\times2$ local scan, a vertical scan (both forward and backward), and a horizontal scan (both forward and backward). Figure \autoref{fig:mhsvm} depicts the architecture of Local Mamba, which incorporates four SSM blocks.

A mechanism that plays a crucial role in Mamba architectures is scanning, which is detailed in the upcoming Section \ref{sec:scanning}.

\subsubsection{Variants of U-Net}\label{subsec:variants}

\begin{figure}[!ht]
    \centering
    \includegraphics[width=\linewidth]{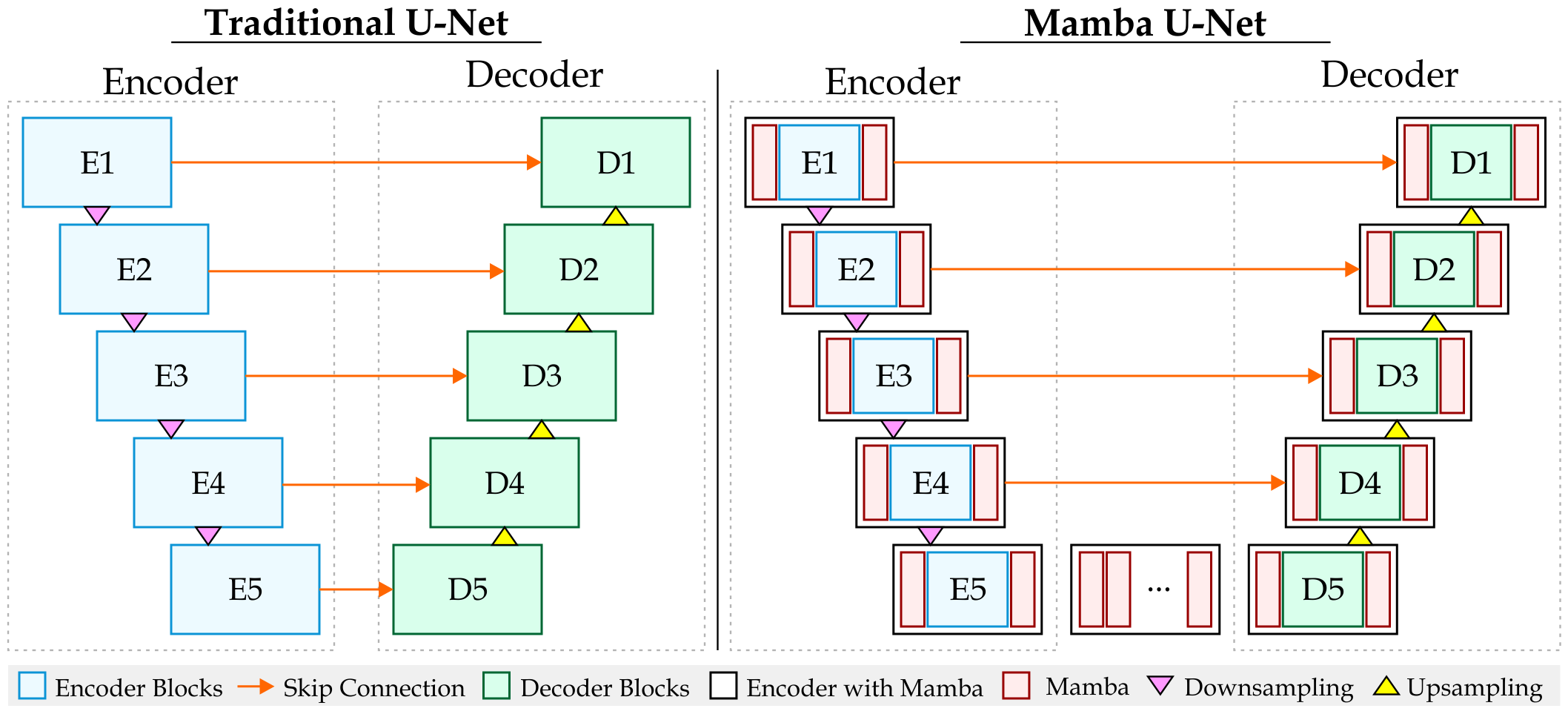}
    \caption{Architectural Comparison of Traditional U-Net and Mamba U-Net}
    \label{fig:uetvsmambaunet}
\end{figure}

\begin{figure}[!ht]
\includegraphics[width=\linewidth]{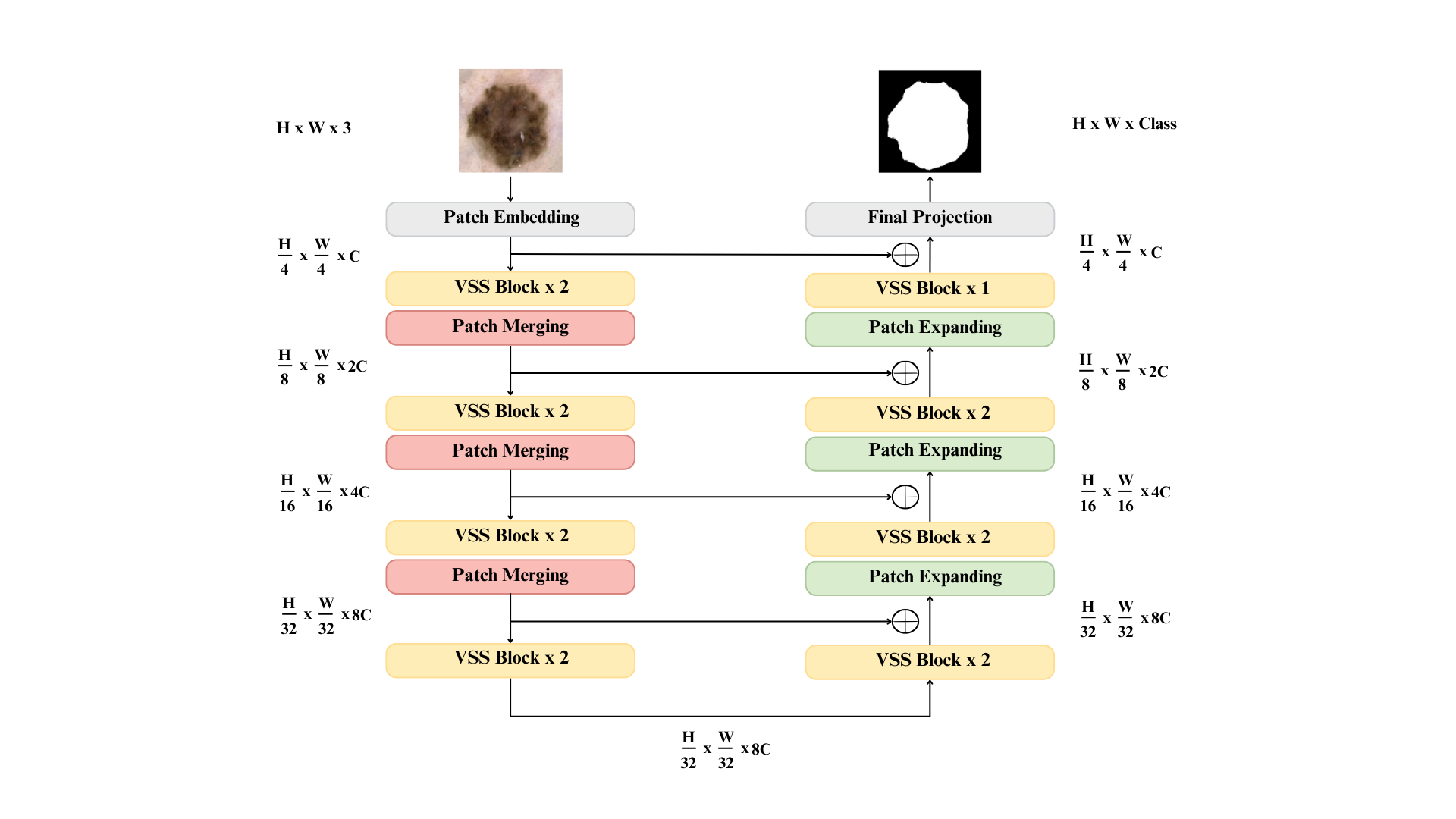}
\caption{\centering Architecture of VM-UNet \cite{ruan2024vmunetvisionmambaunet}}
\label{fig:VM-Unet}
\centering
\end{figure}

The integration of Mamba blocks into the U-Net \cite{ronneberger2015u} architecture
has resulted in several variants which are designed to improve its performance. For instance, they can be added before the first encoder layer, after an encoder layer, within the skip connections or can even replace the whole encoder block in UNet architecture. \autoref{fig:uetvsmambaunet} presents a comparision of traditional U-Net and Mamba U-Net \cite{wang2024mambaunet} architecture. Both networks share a standard UNet architecture having encoder and decoder blocks connected by skip connections but Mamba U-Net incorporates mamba blocks. Ruan \textit{et al.} \cite{ruan2024vmunetvisionmambaunet} proposed VM-UNet where images are converted into tokens using a patch embedding block. The encoder of the network contains two VSS blocks with patch merging block and skip connection in each layer of the encoder. The decoder contains a patch expanding block followed by a VSS block with skip connections added from the encoder. Finally, the representations from the decoder are then passed on to the final projection layer, which reconstructs the image back into its original size, i.e., the number of classes. \autoref{fig:VM-Unet} illustrates the UNet-based architecture of VM-UNet. 

Archit \textit{et al.} \cite{archit2024vimunet} 
introduced ViM-UNet which uses ViM as its base Mamba. ViM-UNet features the ViM encoder with bidirectional SSM layers, while its decoder copies the design of UNEt 
TRansformers (UNETR) \cite{hatamizadeh2021unetrtransformers3dmedical} using convolution and transposed convolution layers. To enhance efficiency, ViM-UNet introduces flexibility in its encoder sizes, such as tiny and small. 
ViM-UNet takes a different approach by excluding these skip connections and using ViM as the encoder. This key architectural change allows ViM-UNet to achieve efficient global feature extraction and improved segmentation performance, without the need for the skip connections that are important in the traditional UNet. Inspired by VM-UNet \cite{ruan2024vmunetvisionmambaunet},  Zhang \textit{et al.} \cite{zhang2024vmunetv2rethinkingvisionmamba} proposed the second version called VM-UNet-V2. The model differs from VM-UNet by not using skip connections between subsequent encoder to decoder layers 
. Instead, features from each block of encoder are fused using the Semantics and Details Infusion block (SDI block). VSS captures contextual information within images, while the SDI module improves the integration of low-level and high-level features, leading to a comprehensive understanding of image features. By combining these elements, VM-UNetV2 maximizes the potential of SSMs within the UNet, thus offering a more efficient and powerful solution for segmentation tasks. 

Wang \textit{et al.} \cite{wang2024lkmunetlargekernelvision} proposed LKM-UNet, a novel architecture designed for efficient 2D and 3D medical image segmentation. LKM-UNet uses the strengths of Mamba to achieve superior performance in both local and global modeling with linear time complexity. The use of large windows within SSM improves the receptive field compared to CNNs and Transformers. The network architecture includes a hierarchical and bidirectional Mamba block, enhancing spatial modeling capabilities by integrating Pixel-level (PiM) and Patch-level (PaM) SSMs. LKM-UNet excels in capturing both fine-grained details and long-range dependencies from the data. The model achieves efficient feature extraction and segmentation through a U-shaped network with encoder-decoder blocks, layers for downsampling and upsampling along with skip connections. 
Wu \textit{et al.} \cite{wu2024hvmunethighordervisionmamba} proposed High-order Vision Mamba UNet (H-vmunet) which enhances the 2D-selective-scan (SS2D) mechanism and also introduced higher-order interactions to reduce redundant information and improve extraction of local features. Traditional architecture like CNN and ViT face limitations in processing long sequences and capturing local features, respectively. The proposed architecture features a U-shaped structure with six layers, which incorporates High-order Visual State Space (H-VSS) module and it applies higher-order SS2D from layers three to six. Inclusion of Spatial Attention Bridge (SAB) and Channel Attention Bridge (CAB) modules in the architecture 
further enhances multilevel and multiscale information fusion, which is crucial for capturing detailed medical image features.

In Section~\ref{subsec:variants}, both Mamba-based and Transformer-based models aim to improve the U-Net architecture~\cite{ronneberger2015u} by replacing standard convolutional layers. Mamba-based U-Nets incorporate blocks such as Vision Mamba (ViM)\cite{zhu2024vision} and Visual State Space (VSS)\cite{liu2024vmamba}, which improve the model's ability to capture long-range dependencies efficiently. For example, the architecture of VM-UNet~\cite{ruan2024vmunetvisionmambaunet}, shown in Figure~\ref{fig:VM-Unet}, uses VSS blocks as core components within a U-Net-based structure. In contrast, Transformer-based U-Nets employ attention mechanisms, such as the Swin Transformer~\cite{liu2021swin} and Vision Transformer (ViT)\cite{dosovitskiy2021imageworth16x16words}, to model global relationships in images. For instance, Swin U-Net\cite{cao2022swin} follows a similar architectural design to VM-UNet but replaces the VSS blocks with Swin Transformer blocks. While both approaches improve upon the original U-Net, Mamba-based models offer a better balance between accuracy and efficiency, achieving competitive segmentation performance with fewer parameters and reduced computational cost.

\subsubsection{Hybrid Architectures}\label{subsec:hybrid}
\addtocontents{toc}{\setcounter{tocdepth}{2}}

In this section, we explore hybrid architectures that combine Mamba with other powerful techniques. We cover Mamba's integration with convolutions for enhanced feature extraction, attention mechanisms and transformers for capturing contextual relationships, recurrence for modeling sequential data and GNNs for graph structured data. 
Additionally, we touch upon other miscellaneous hybrid approaches that demonstrate Mamba's versatility.
\paragraph{\textbf{Mamba with Convolution:}} CNNs often face challenges in capturing long-range dependencies due to their inherent focus on local features and computational complexity. SSMs have the ability to handle long sequences of data. The combination of Mamba with convolution plays a huge role in capturing local spatial information along with capturing long range dependencies from medical images. Some of the papers that demonstrate these hybrid approaches are as follows:
 Ma \textit{et al.} \cite{Ma2024UMambaEL} introduced U-Mamba, a hybrid CNN-SSM architecture integrating Mamba blocks within the encoder of U-Net, demonstrating superior performance over traditional CNN-based and transformer-based segmentation networks across various modalities and segmentation targets. Wang \textit{et al.} \cite{wang2024mambaunet} proposed Mamba-U-Net which integrates pure Vision mamba in the U-Net along with linear embedding and VSS block in the model. Xu \textit{et al.} \cite{xu2024hcmambavisionmambahybrid} developed HC-Mamba, which uses dilated convolution followed by depth-wise separable convolution along with Mamba. It improves the receptive field and reduces the parameters of the model. Another innovative model, SegMamba designed by  Xing \textit{et al.} \cite{xing2024segmambalongrangesequentialmodeling}, mimics a Transformer-based U-Net architecture, utilizing a series of TS-Mamba block and downsample block with residual connections inspired by He \textit{et al.} \cite{he2015deep}. TS-Mamba employs gated spatial convolution to analyze spatial relationships between 3D features, followed by the Tri-oriented Spatial Mamba (TOM) block, which replaces the traditional multi-headed attention layers in transformers.
 Gong \textit{et al.} \cite{gong2024nnmamba} proposed nnMamba, which features the Mamba-In-Convolution with Channel-Spatial Siamese (MICCSS) block. The backbone architecture of nnMamba leverages the MICCSS block to maintain computational efficiency. This helps in improving the model’s representational capacity and improving the performance on tasks that require a deep understanding of visual data.

Liu \textit{et al.} \cite{liu2024swinumambamambabasedunetimagenetbased} proposed Swin-UMamba which is a hybrid architecture that combines VSS with UNet for medical image segmentation tasks. Swin-UMamba 
integrates a Mamba-based encoder pretrained on ImageNet with a series of VSS blocks and upsample blocks. The architecture of Swin-UMamba includes patch embedding, VSS blocks, patch merging, up-sample blocks, residual blocks, and a 1x1 convolution for segmentation output which combines Mamba with traditional segmentation network components. By combining the strengths of Mamba-based models with segmentation network structures, Swin-UMamba achieves better results in medical image segmentation tasks, highlighting its effectiveness in improving pretrained models for improved segmentation accuracy. 
\paragraph{\textbf{Mamba with Attention and Transformers:}}
Wang \textit{et al.} \cite{wang2024weakmambaunetvisualmambamakes} proposed Weak-Mamba-UNet  which uses a CNN-based UNet for extracting local features from data, a Swin Transformer-based SwinUNet for understanding global context and a VMamba based Mamba-UNet for modeling long-range dependencies. The results of Weak-Mamba-UNet shows that it solves inaccurate predictions caused by other approaches.  Kirillov \textit{et al.} proposed the Segment Anything Model (SAM) \cite{kirillov2023segment} which was originally done for 2D images. Inspired by this, Wang \textit{et al.} \cite{wang2024triplanemambaefficientlyadapting} proposed Efficiently Adapting Segment Anything Model (SAM) which shows good performance on 3D medical images. The authors have modified the encoder of SAM by adding Tri-Plane Mamba (TP-Mamba) block on top of each ViT and the decoder of the model uses 3D convolution with instance normalization \cite{ulyanov2017instancenormalizationmissingingredient} and GELU activation function \cite{hendrycks2023gaussianerrorlinearunits}. The authors have also used LoRA \cite{hu2022lora} to adapt weights of Multi-Head self Attention module.
TP-Mamba shows better performance compared to transformer based models, U-Mamba \cite{Ma2024UMambaEL} and other adapter-based algorithms for SAM such as SA-Med \cite{zhang2023customizedsegmentmodelmedical}, MA-SAM \cite{chen2023masammodalityagnosticsamadaptation}. 

Zhang \textit{et al.} \cite{zhang2024hmtunethybirdmambatransformervision} proposed HMT-UNet which combines MambaVision Mixer with the transformer's self attention block for segmentation of medical images. MambaVision mixer is a modification of the Mamba layer, where the first casual convolution is replaced with regular convolution and passed on to SSM. Initially, the encoder of UNet consists of four layers where the initial two layers uses two $3\times3$ convolution with stride of two and downsampling uses batch normalized $3\times3$  convolution with stride of two which scales down the resolution of the spatial dimension by half and doubles the channel. The layer 3 and layer 4 contains MambaVision Mixer and transformer’s self attention block. In the decoder block, the spatial dimensions are doubled and channel size is halved. The final stages of decoder use the same configuration as initial stages of encoder but instead of downsampling the image as in encoder, it linearly upsamples the image.

\paragraph{\textbf{Mamba with Recurrence:}}

VMRNN proposed by Tang \textit{et al.} \cite{Tang2024VMRNNIV} is a recurrent cell that incorporates VSS block within an Long Short Term Memory (LSTM) \cite{lstm1997} network. Mamba's integration into spatial-temporal forecasting based on vision allows robust sequential modeling. The paper introduces two novel architectures: VMRNN-B and VMRNN-D. These architectures excel at extracting spatiotemporal features, establishing a new strong foundation for spatiotemporal forecasting. Spatiotemporal predictive learning differs from traditional image-level vision tasks by predicting future video frames based on past sequences. VMRNN tackles this challenge by processing individual frames, segmenting them into smaller patches. These patches get flattened before feeding them into a patch embedding layer for initial processing. The VMRNN layer then leverages these transformed patches alongside past states to capture the crucial spatiotemporal features necessary for predicting the next frame. The hybrid architecture of VSS and LSTM is attributed to the model’s capability to learn and leverage global spatial dependencies with linear complexity, enabling a more refined understanding of spatiotemporal dynamics.
\paragraph{\textbf{Mamba with GNN:}}
Ding \textit{et al.} \cite{Ding2024CombiningGN} combines Mamba with Graph Neural Network 
 (GNN) to capture global and local tissue spatial relationships respectively in Whole Slide Images (WSI). The model leverages a message-passing GNN, specifically a Graph Attention Network (GAT) proposed by Velickovic \textit{et al.} \cite{Velickovic2017GraphAN}, to process a hierarchical graph constructed from both cell-level and tile-level graphs. Graph convolution operation is performed using GAT which assigns importance scores to neighboring nodes, allowing the model to focus on the most relevant information for the current node during the convolution process. 

\paragraph{\textbf{Miscellaneous:}}
Nasiri \textit{et al.} \cite{nasiri2024vim4path} proposed Vim4Path which used ViM architecture within the DINO framework proposed by Caron \textit{et al.} \cite{caron2021emerging}, for learning representations in computational pathology. ViM is modified to accept arbitrary input image sizes using positional embedding interpolation, making it adaptable within DINO for Self-Supervised Learning (SSL). The study benchmarks the Camelyon16 dataset \cite{camelyon}, extracting image patches from Whole Slide Images (WSIs) without labels for training the ViM encoder using the DINO framework. Yang \textit{et al.} \cite{yang2024mambamil} introduced the MambaMIL framework, which combines Mamba with Multiple Instance Learning (MIL) to enhance long sequence modeling in computational pathology. The core component of the MambaMIL framework is Sequence Reordering Mamba (SR-Mamba), which is designed to be aware of the order and distribution of instances within long sequences. 
The MambaMIL framework partitions tissue regions into a sequence of patches, maps these patches into instance features, reduces feature dimension through linear projection, utilizes stacked SR-Mamba modules for handling long sequences, and finally employs an aggregation module to obtain bag-level representations for downstream tasks.

 This Section \ref{subsec:hybrid}  has explored various architectures that combine Mamba with CNNs, Transformers, GNNs, and other models to tackle complex tasks like medical imaging. Relying on a single architecture often falls short in effectively capturing both fine-grained local features and long-range dependencies, making the combination of complementary models increasingly important. Mamba offers efficient sequence modeling with low computational cost, making it a valuable addition to these architectures. In CNN-based models such as U-Mamba \cite{Ma2024UMambaEL} and HC-Mamba \cite{xu2024hcmambavisionmambahybrid}, it enhances spatial feature learning and long-range dependency capture. In Transformer-based models such as TP-Mamba \cite{wang2024triplanemambaefficientlyadapting} and HMT-UNet \cite{zhang2024hmtunethybirdmambatransformervision}, Mamba reduces the computational burden of self-attention while maintaining global context understanding. In recurrent models like VMRNN \cite{Tang2024VMRNNIV}, it strengthens temporal modeling for video forecasting, while in GNNs, it improves hierarchical spatial reasoning through attention-driven message passing. Common components in these combinations such as attention modules, gating mechanisms, residual connections, patch embeddings, and lightweight convolutions, demonstrate a clear trend toward leveraging Mamba’s strengths to boost performance and efficiency across diverse deep learning tasks.

\subsection{Scanning}\label{sec:scanning}
\begin{figure}[!ht]
    \centering
  \subfigure[Depicts (\romannumeral 1) BiDirectional Scan \cite{zhu2024vision}, (\romannumeral 2) Continuous 2D Scan \cite{yang2024plainmambaimprovingnonhierarchicalmamba}, (\romannumeral 3) Local Scan \cite{huang2024localmambavisualstatespace}, (\romannumeral 4) Cross Scan \cite{liu2024vmamba} and (\romannumeral 5) Multi-Head Scan \cite{ji2024mhsvmmultiheadscanningparallel}
    \label{fig:scanninga}
    ]{
        \includegraphics[width=\linewidth, height=9cm, keepaspectratio]{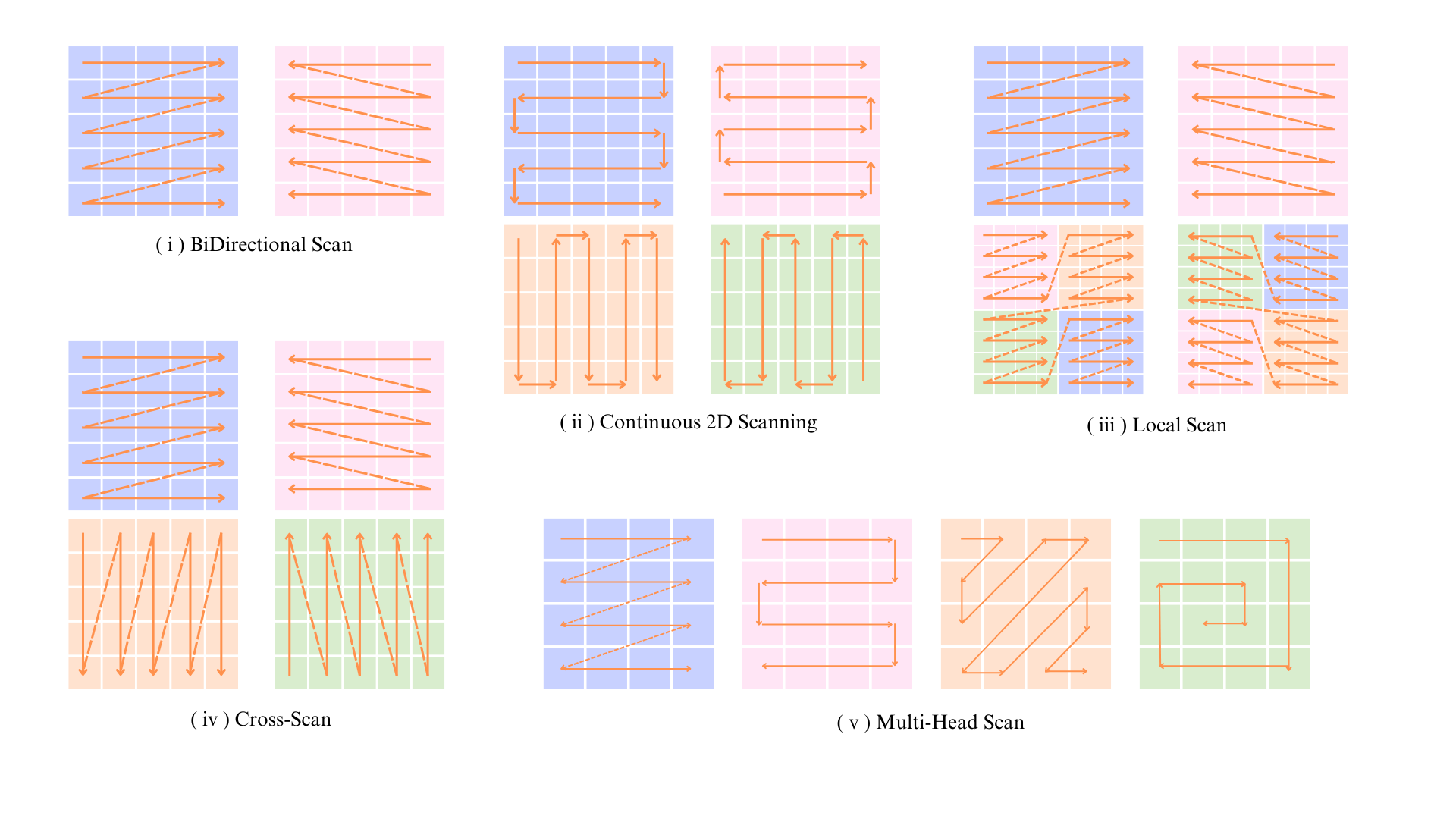}}
    \vspace{1em}
    \subfigure[Demonstrates (\romannumeral 6) Selective 2D Scan \cite{liu2024vmamba}, (\romannumeral 7) Efficient 2D Scan \cite{Pei2024EfficientVMambaAS}, (\romannumeral 8) Omnidirectional Selective Scan \cite{shi2024vmambairvisualstatespace}, (\romannumeral 9) 3D BiDirectional Scan \cite{li2024videomambastatespacemodel}
    \label{fig:scanningb}
    ]{        \includegraphics[width=\linewidth, height=9cm, keepaspectratio]{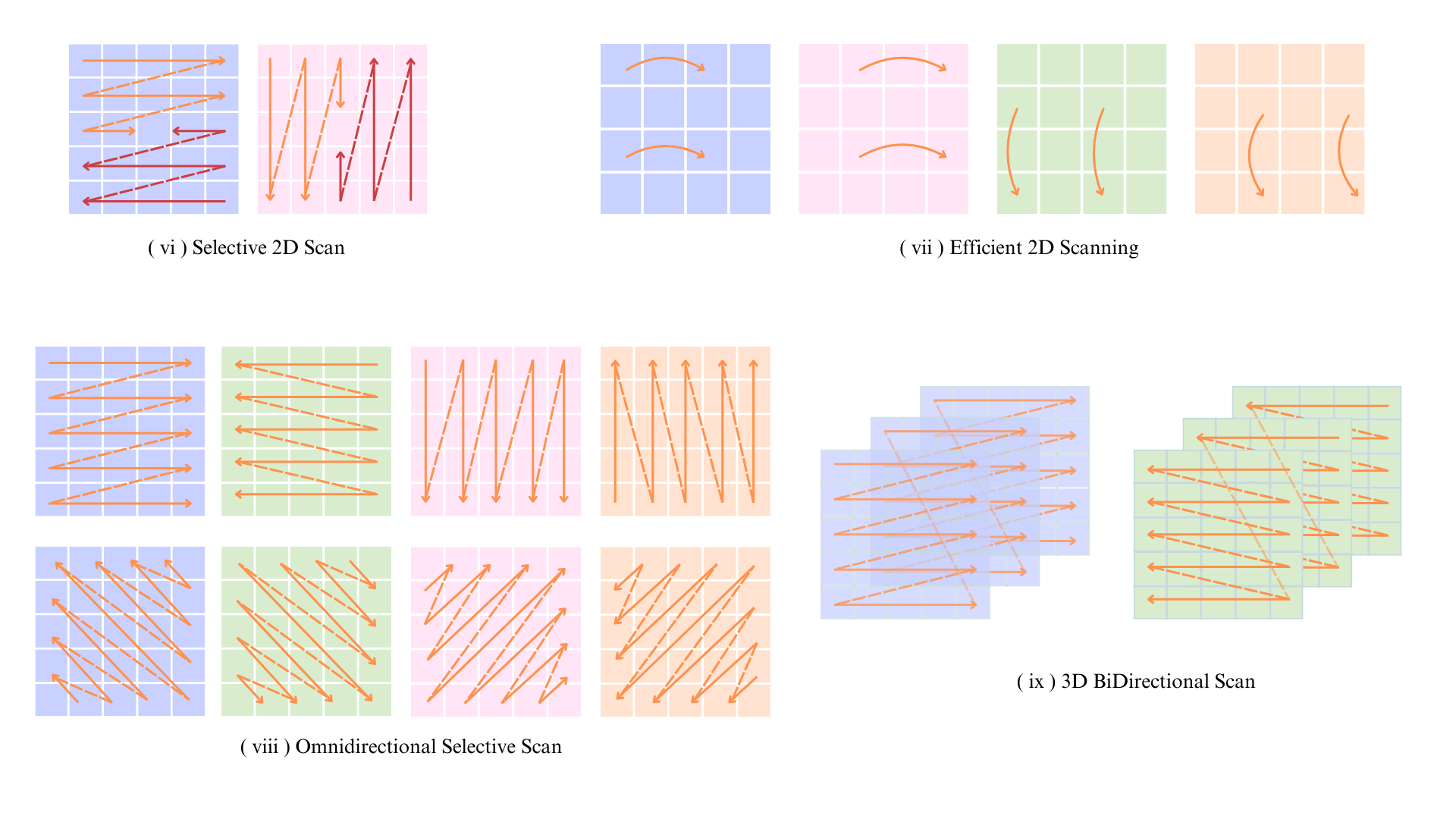}}
    \label{fig:scanning_part1}
\end{figure}

\begin{figure}[!ht]
    \centering
    \subfigure[Shows Visual Representation of (\romannumeral 10) Hierarchical Scan \cite{zhang2024motionmambaefficientlong}, (\romannumeral 11) Zigzag Scan \cite{hu2024zigmaditstylezigzagmamba}, (\romannumeral 12) Spatiotemporal Scan \cite{yang2024vivimvideovisionmamba}, (\romannumeral 13) Multi-Path Scan \cite{chen2024rsmambaremotesensingimage}
    \label{fig:scanningc}
    ]{
        \includegraphics[width=\linewidth, height=8.2cm, keepaspectratio]{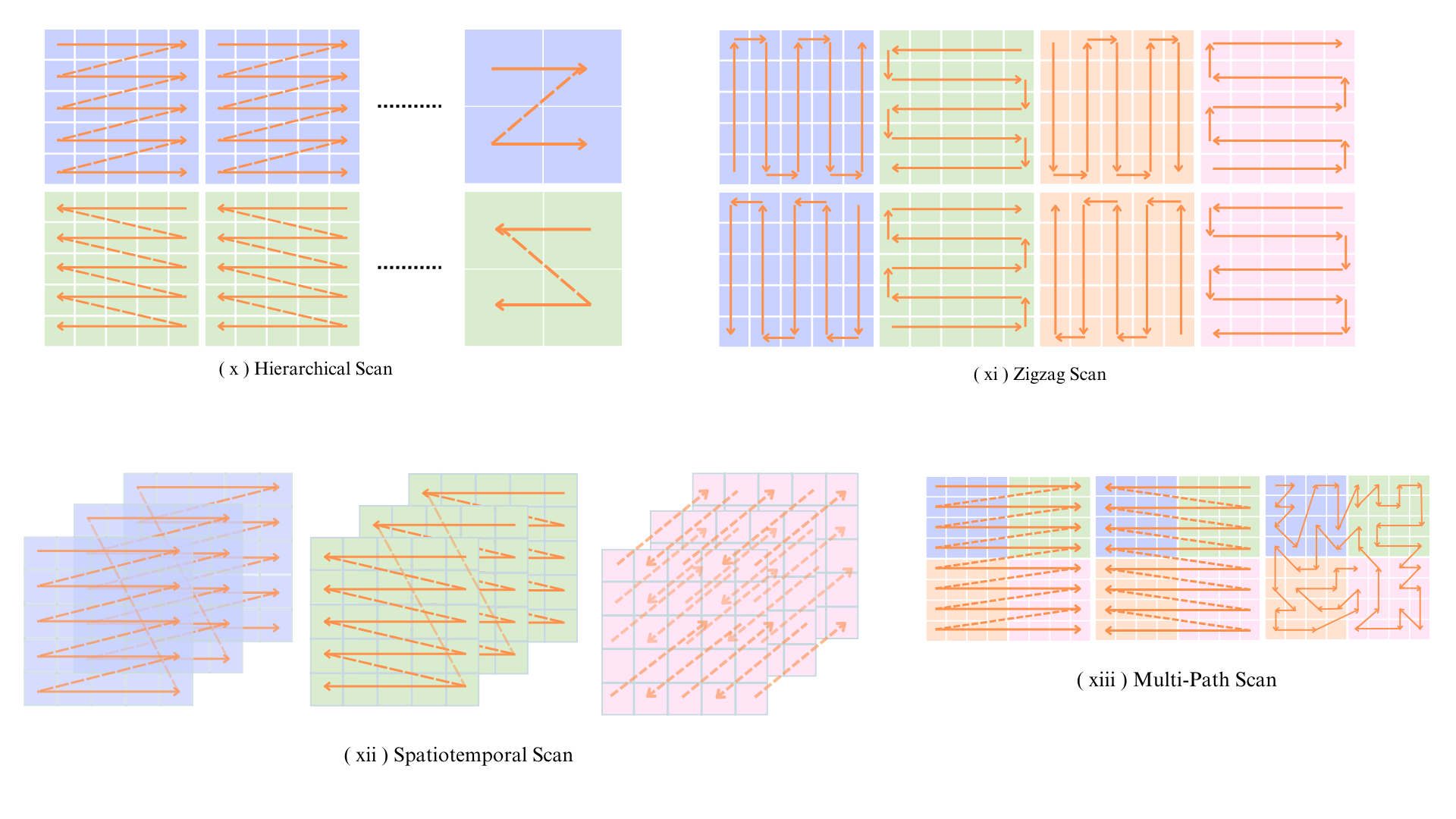}}
        \vspace{1em}
            \subfigure[Depicts (\romannumeral 14) Three Directional Scan \cite{xing2024segmambalongrangesequentialmodeling}, (\romannumeral 15) BSS Scan \cite{fan2024slicemambaneuralarchitecturesearch}  and (\romannumeral 16) Pixelwise and Patchwise Scan \cite{ wang2024lkmunetlargekernelvision}
    \label{fig:scanningd}
    ]{
    \includegraphics[width=\linewidth, height=8.2cm, keepaspectratio]{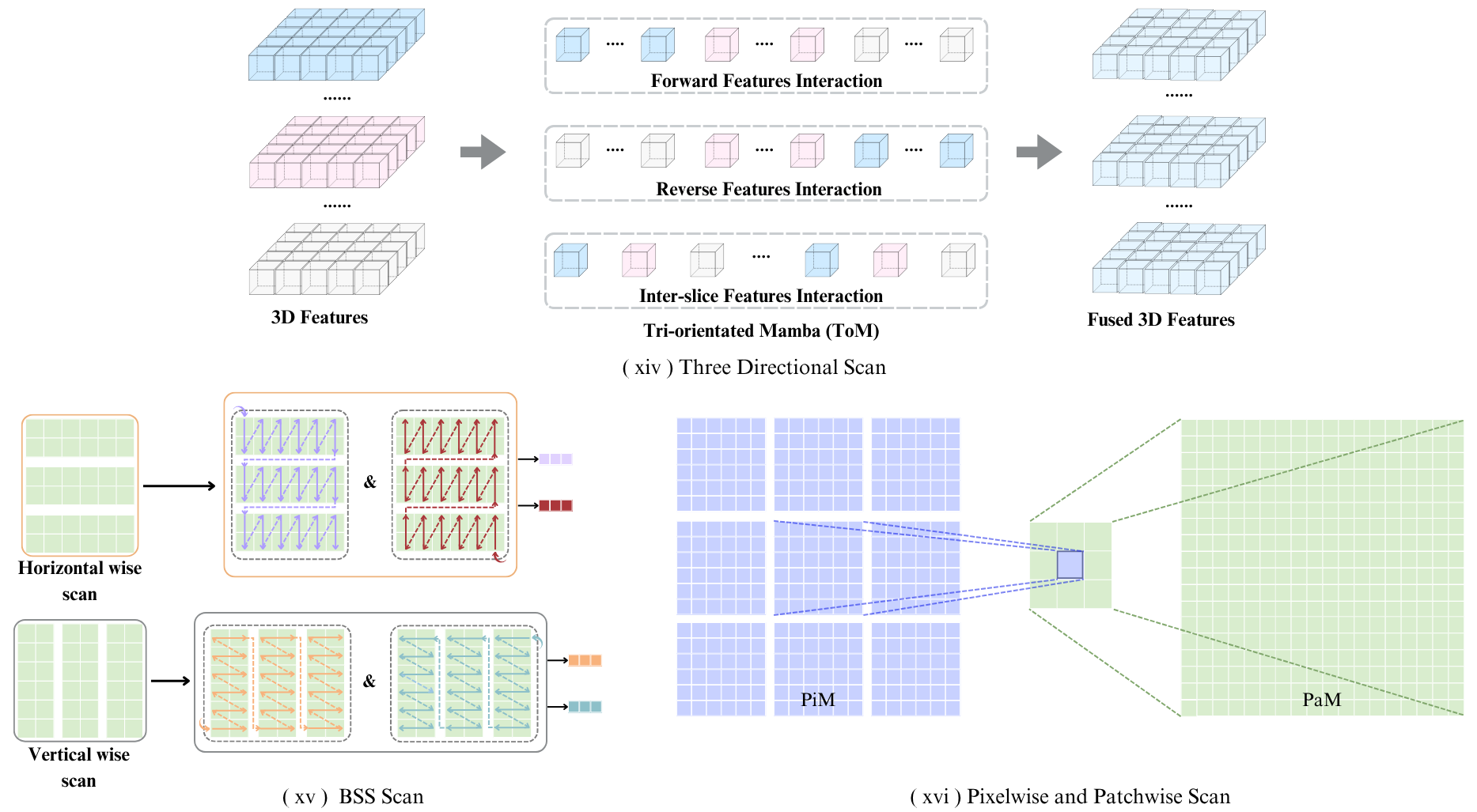}}
    
    \caption{Illustrations in \autoref{fig:scanninga}, \autoref{fig:scanningb}, \autoref{fig:scanningc} \& \autoref{fig:scanningd} present Different Scanning Mechanisms used in Mamba-based Architectures}
    \label{fig:scanning_part2}
\end{figure}

Attention mechanisms, especially self-attention have a quadratic time complexity causing computational costs to grow quadratically with sequence length. In contrast, scanning operations generally have linear time complexity, making them more efficient for long sequences. The scan operation involves calculating an array, like the prefix sum, where each value is determined by using the previously calculated value and the current input. Similarly, the recurrent form of SSM can be viewed as a scan operation. Scanning is a crucial component in mamba, especially when handling multidimensional inputs. The selection of the scanning mechanism in Mamba models is crucial as it enhances efficiency and provides important information. \autoref{fig:scanning_part2} provides visualization of various scanning mechanisms employed in mamba-based architectures. This visualization highlights the diversity of scanning approaches integrated into mamba models. \autoref{tab:scanning} summarizes the various scanning mechanisms and the models associated with each mechanism. The scanning methodologies used in mamba models are detailed as follows:
\begin{enumerate}
\item \textbf{Bidirectional Scan} : In bidirectional scan (forward and backward scan) \cite{zhu2024vision}, after tokenizing image patches, they are processed through the forward SSM. Simultaneously, the same tokenized representations of images are independently processed through the backward SSM. This scanning mechanism primarily used in ViM-based models, enables the model to capture contextual information from both directions, improving its ability to understand and represent the image data effectively.

\item \textbf{Selective Scan 2D} : SS2D \cite{liu2024vmamba} performs scanning operations in three directions: top to bottom, left to right, and in reverse direction. Each mamba block is placed to work independently within these directions. SS2D mirrors the self-attention process seen in transformers. It overcomes the limitations of bidirectional scan in ViM, but it also leads to a loss of patch continuity. To address this, SS2D incorporates a scan merge step, where representations from each scan direction are combined into a unified output.

\item \textbf{Continuous 2D Scan}  : Continuous 2D scan \cite{yang2024plainmambaimprovingnonhierarchicalmamba} resolves the issue which is experienced in SS2D. It involves integrating direction-aware parameters into cross-scan mechanism and organizing patches accordingly. This approach ensures the preservation of patch continuity and maintains the contextual understanding of images. The continuous 2D scanning adds direction-aware parameter into data dependent parameter of SSM (B) which is expressed in  \autoref{eq:13} and \autoref{eq:14}. 
\begin{align} \label{eq:13}
    h'_{k,i}=\overline{\textbf{\textit{A}}}_{i}h_{k,i-1}+(\overline{\textbf{\textit{B}}}_{i}+\overline{\mathbf{\Theta}}_{k,i})x_{i}
\end{align}
\begin{align} \label{eq:14}
    y'_{i}=\mathbf{\sum_{k=1}^{4}}(\textbf{\textit{C}}_{i}h'_{k,i}+\textbf{\textit{D}}x_{i}),\;\;\;\;\;\;y_{i}=y'_{i}\mathbf{\odot}z_{i}
\end{align}

\item \textbf{Zigzag Scan}  : The extension of continuous 2D Scan is zigzag scan \cite{hu2024zigmaditstylezigzagmamba} where the images are scanned with continuous scanning mechanism in both forward and backward direction. Zigzag scan is developed to enhance the continuity of patches in the images which are used for diffusion models such as ZigMa \cite{hu2024zigmaditstylezigzagmamba}. 

\item \textbf{Spatiotemporal Selective Scan} : Spatiotemporal selective scan \cite{yang2024vivimvideovisionmamba} is used to scan on videos where the patches are unfolded on each frame along rows and columns and then concatenated with the frame sequence of $h_{i}^{t}\;\mathbf{\epsilon}\;\mathbb{R}^{C_{i}\times T(HW)}$ . In this setup, scanning is done bidirectionally to know about temporal dependency. Parallelly, scanned patches are stacked around temporal axis to construct the spatial sequence in the form of $h_{i}^{s}\;\mathbf{\epsilon}\;\mathbb{R}^{C_{i}\times(HW)T}$,  to integrate information of each pixel from all frames. In short, one scan focuses on scanning with time dependency along frames and the other focuses on scanning each pixel along the time axis.

\item \textbf{Local Scan} : Local Scanning \cite{huang2024localmambavisualstatespace} overcomes limitations of scanning methods in ViM and VMamba by preserving local dependencies in images through distinct local windows. This technique maintains the global context of the image without compromise. The authors suggest using $7 \times 7$ and $2 \times 2$ local windows to capture the local context while alternating the scan direction. Vertical and horizontal scans with direction flipping are used to grasp the global context of image tokens. 

\begin{table*}[!ht]
\caption{Summary of Scanning Mechanisms and Associated Models}
\label{tab:scanning}
\centering
  \begin{tabular}{ll}
    \toprule
    \textbf{Scanning Mechanism} & \textbf{Models} \\ 
    \midrule
    BiDirectional Scan  & Vision Mamba \cite{zhu2024vision}, MamMIL \cite{fang2024mammilmultipleinstancelearning},\\  
                        &  Motion-Guided Dual-Camera Tracker \cite{zhang2024motionguideddualcameratrackerlowcost}\\
                        &  CAMS-Net \cite{khan2024camsconvolutionattentionfreemambabased}\\
         
    Selective Scan 2D(SS2D)  & VMamba \cite{liu2024vmamba}, MedMamba \cite{yue2024medmambavisionmambamedical},                                                     P-Mamba \cite{ye2024pmambamarryingperonamalik},\\   
                        &  Weak-Mamba-UNet \cite{wang2024weakmambaunetvisualmambamakes},
                        VM-UNET-V2 \cite{zhang2024vmunetv2rethinkingvisionmamba}, \\
                        & LightM-UNet \cite{liao2024lightmunetmambaassistslightweight}, 
                        HC-Mamba \cite{xu2024hcmambavisionmambahybrid},\\
                        & H-vmunet \cite{wu2024hvmunethighordervisionmamba}, Mamba-HUNet \cite{sanjid2024integratingmambasequencemodel}, \\
                        & UltraLight VM-UNet \cite{wu2024ultralightvmunetparallelvision}, VM-UNet \cite{ruan2024vmunetvisionmambaunet}\\
                          &  VMambaMorph \cite{Wang2024VMambaMorphAM}, Mamba-UNet \cite{wang2024mambaunet},\\ 
                & Swin-UMamba \cite{liu2024swinumambamambabasedunetimagenetbased}, Semi-Mamba-UNet \cite{ma2024semimambaunetpixellevelcontrastivepixellevel},\\
                &  MambaMIR \cite{huang2024mambamirarbitrarymaskedmambajoint}, VM-DDPM \cite{ju2024vmddpmvisionmambadiffusion} \\
                        
    Spatiotemporal Selective Scan  & Vivim \cite{yang2024vivimvideovisionmamba} \\  
    Zigzag Scan  & ZigMa \cite{hu2024zigmaditstylezigzagmamba}\\ 
    Local Scan  & LocalMamba \cite{huang2024localmambavisualstatespace} , FreqMamba \cite{zhen2024freqmambaviewingmambafrequency}  \\ 
    Efficient 2D Scan  & EfficientVMamba \cite{Pei2024EfficientVMambaAS}, FusionMamba \cite{peng2024fusionmambaefficientimagefusion} \\ 
    Continuous 2D Scan  & PlainMamba \cite{yang2024plainmambaimprovingnonhierarchicalmamba} \\  
    Tri-orientated Spatial Mamba (TOM)\\ - Three Directional Scan  & SegMamba \cite{xing2024segmambalongrangesequentialmodeling} \\  
    Pixelwise and Patchwise Scan  & LKM-UNet \cite{wang2024lkmunetlargekernelvision} \\ 
    Omnidirectional Selective Scan  & VmambaIR \cite{shi2024vmambairvisualstatespace}  \\  
    Hierarchical Scan  & Motion Mamba\cite{zhang2024motionmambaefficientlong} \\  
    Multi-Head Scan  & MHS-VM \cite{ji2024mhsvmmultiheadscanningparallel} \\  
    Multi-Path Scan & RSMamba\cite{chen2024rsmambaremotesensingimage}  \\  
    3D BiDirectional Scan  & VideoMamba\cite{li2024videomambastatespacemodel}  \\ 
    Bidirectional Slice Scan (BSS) & SliceMamba \cite{fan2024slicemambaneuralarchitecturesearch}\\

    \bottomrule
    \end{tabular}
\end{table*}

\item \textbf{Efficient 2D Scan}  : Efficient Scan 2D (ES2D) \cite{Pei2024EfficientVMambaAS} emphasizes efficient image scanning by skipping scan patches with a step size $p$. It partitions selected spatial dimension features into $m$ and $n$ using sine and cosine functions to determine the patch location. The entire operation is mathematically expressed in \autoref{eq:15}.

    \begin{align} \label{eq:15}
        \textbf{\textit{O}}_{i}\overset{scan}{\leftarrow}\textbf{\textit{X}}[:,m::p,n::p], 
    \end{align}
    \[
        \begin{Bmatrix}{\tilde{\textbf{\textit{O}}}}_{i}\end{Bmatrix}_{i=1}^{4}\leftarrow SS2D(\begin{Bmatrix}\textbf{\textit{O}}_{i}\end{Bmatrix}_{i=1}^{4}), 
    \]
    \[
        \textbf{\textit{Y}}[:,m::p,n::p]\overset{merge}{\leftarrow}\tilde{\textbf{\textit{O}}}_{i}, 
    \]
    \[
        with\;(m,n)=(\left\lfloor\frac{1}{2}+\frac{1}{2}sin(\frac{\pi}{2}(i-2))\right\rfloor,\left\lfloor\frac{1}{2}+\frac{1}{2}cos(\frac{\pi}{2}(i-2))\right\rfloor)
    \]

\item \textbf{Multi-Path Scan} : Multi-path scanning mechanism \cite{chen2024rsmambaremotesensingimage} incorporates reverse, forward and random shuffling paths. A simple approach to combine the information flow from different paths would be  averaging. However, the objective is to selectively activate the information from each path. Consequently, a gating mechanism is designed to manage the information flow from various paths.

\item \textbf{Omnidirectional Selective Scan} : There are two information streams\\
$\textbf{\textit{F}}_{01},\textbf{\textit{F}}_{02}\;\mathbf{\in}\;\mathbf{\mathbb{R}}^{B\times C\times H\times W}$, serving as inputs for the Omnidirectional Selective Scan (OSS) \cite{shi2024vmambairvisualstatespace}. In the first stream, bidirectional scanning is performed both longitudinally and transversely on $\textbf{\textit{F}}_{01}$ to capture planar two-dimensional feature information. The second stream is refined by depthwise convolution and SiLU activation \cite{elfwing2017sigmoidweightedlinearunitsneural}, capturing detailed patterns. These two streams are then fused within OSS, merging refined features with complementary information. After passing through a $1 \times 1$ convolution, the output of OSS, $\textbf{\textit{F}}_{OSS}\;\mathbf{\in}\;\mathbf{\mathbb{R}}^{B\times C\times H\times W}$ provides a detailed input representation, improving feature extraction and modeling capabilities.

\item \textbf{Hierarchical Scan} : Hierarchical Temporal Mamba (HTM) \cite{zhang2024motionmambaefficientlong} block processes compressed latent representations $z$ of dimensions (\textbf{\textit{T}}, \textbf{\textit{B}}, \textbf{\textit{C}}) using a hierarchical scanning methodology. Initially, $z$ undergoes a linear projection to produce representations $x$ and $z$ of dimension $E$. A set of scans $K$ and memory matrices are applied, where each scan involves $1D$ convolution and linear projections to derive transformed outputs. These outputs are combined through a linear projection to produce the final transformed representations $\textbf{\textit{z}}_{\textbf{\textit{HTM}}}$, efficiently capturing diverse motion densities and minimizing computational overhead. 

\item \textbf{Multi-Head Scan} : Multi-headed scanning \cite{ji2024mhsvmmultiheadscanningparallel} processes an input embedding map $\textbf{\textit{X}}_{l-1}$ with shape (\textbf{\textit{B}}, \textbf{\textit{H}}, \textbf{\textit{W}}, \textbf{\textit{C}}) to produce an output embedding map $\textbf{\textit{X}}_{l}$ of the same shape through linear transformations and concatenations. Initially, the input is reshaped to (\textbf{\textit{B}}, \textbf{\textit{C}}, \textbf{\textit{H}}, \textbf{\textit{W}} ) and then processed with $n$ scan heads. Each scan head projects the input onto a subspace and routes it through $K$ scan routes, involving specific transformations, activation, and rearrangement. Scan route outputs are concatenated and fused using the coefficient of variation and summation, followed by ReLU. The combined results from all scan heads are concatenated along the channel axis and optionally projected back to the original number of channels. Finally, the output is reshaped to (\textbf{\textit{B}}, \textbf{\textit{H}}, \textbf{\textit{W}}, \textbf{\textit{C}}) and returned.

\item \textbf{Other Scanning Mechanisms}: Segmamba \cite{xing2024segmambalongrangesequentialmodeling} uses TOM, which stands for Tri-Orientated Spatial Mamba Block. This block employs scanning along three dimensions: height, width and channels and three directions: forward direction, reverse direction, and inter-slice direction. LKM-UNet \cite{wang2024lkmunetlargekernelvision} applies a two-scan strategy where the first scan involves pixel-level scanning. Each pixel is scanned unidirectionally (forward) and max-pooled into a single image. Subsequently, another round of unidirectional scanning (forward) is performed on these pooled images. Bidirectional Slice Scan (BSS) \cite{fan2024slicemambaneuralarchitecturesearch} proposed in Slice Mamba plays a crucial role in its architecture. Firstly the spatial dimension height and width are split into $m$ and $n$ windows separately resulting in the sequence of shape $F^{h}= \{ \mathrm{f}_{i}^{h} \in  \mathbb{R} ^{m \times W \times C}| i=1,2,...\frac{H}{m}\}$ horizontally and $F^{v}= \{ \mathrm{f}_{j}^{v} \in  \mathbb{R} ^{H \times n \times C}| j=1,2,...\frac{W}{n}\}$ sequence of shape vertically. Scanning is applied both in horizontal (both forward and backward) and vertical `(both forward and backward) direction. The optimal size of $m$ and $n$ are chosen through adaptive slice search which uses Neural Architecture Search (NAS) for each layer of mamba blocks using a single path one shot.
\end{enumerate}

\addtocontents{toc}{\setcounter{tocdepth}{3}}
\subsection{Mamba Optimizations}\label{sec:optimizations}
In this section, we discuss research papers that focus on lightweight, efficient, and optimized model architectures.
\subsubsection{Lightweight and Efficient}\label{subsec:light}

Lightweight and efficient models are designed to be smaller, quicker, and use fewer resources, while maintaining good performance. Light Mamba UNet (LightM UNet) proposed by Liao \textit{et al.} \cite{liao2024lightmunetmambaassistslightweight} combines Mamba and UNet architectures in a lightweight framework which aims to tackle computational challenges in real world  medical environment. The Residual Vision Mamba (RVM) layer is proposed to improve SSM for the deep semantic feature extraction from images in a pure Mamba-based manner. 
UltraLight Vision Mamba UNet (UltraLight VM-UNet) introduced by Wu \textit{et al.} \cite{wu2024ultralightvmunetparallelvision} is a lightweight vision Mamba model. An excellent performance is achieved by Parallel vision Mamba (PVM) method that is used for efficiently processing deep features with the lowest computational complexity, while maintaining the overall number of processing channels constant. PVM is primarily composed of Mamba combined with residual connections and adjustment factors. This combination allows traditional Mamba to capture remote spatial relations without introducing additional parameters and computational complexity. 

\begin{table}[!ht]
  \caption{Comparison of Lightweight Models based on Parameters, GFLOPs, and FPS}
  \label{tab:lightweight_models}
  \centering
  \begin{tabular}{cccc}
    \toprule
    Models & Params(M) $\downarrow$ & GFLOPs $\downarrow$ & FPS $\uparrow$\\
    \midrule
    {LightM UNet \cite{liao2024lightmunetmambaassistslightweight}}  & 1.09& 267.19&- \\
    {UltraLight VM-UNet \cite{wu2024ultralightvmunetparallelvision}} & 0.049&0.060 & -\\
    {MUCM-Net \cite{yuan2024mucmnetmambapowereducmnet}} & 0.071 to 0.139& 0.055 to 0.064&-\\
    {LightCF-Net \cite{bioengineering11060545}}  & 1.52& 3.25&33\\
    {MiM-ISTD \cite{Chen2024MiMISTDMF}}  & 4.76&3.95 &-\\
    \bottomrule
    \multicolumn{4}{l}{\small GFLOPS = (Number of Floating-Point Operations) / (Elapsed Time in Seconds) / ($10^{9}$)
    } \\
    \multicolumn{4}{l}{\small $\downarrow$ - denotes lower is better, $\uparrow$ - denotes higher is better
    }
  \end{tabular}
\end{table}
\autoref{tab:lightweight_models} compares the above mentioned lightweight models based on Giga Floating-point Operations Per Second (GFLOPs), number of parameters, and FPS Frames Per Second (FPS). These metrics provide a detailed evaluation of each model's computational efficiency, complexity and speed respectively. MUCM-Net proposed by Yuan \textit{et al.} \cite{ yuan2024mucmnetmambapowereducmnet} is an efficient model which combines Mamba State-Space Models with UCM-Net architecture to improve segmentation and feature learning. In this model, Mamba-UCM is optimized for mobile deployment, providing high accuracy with minimal computational requirements (approximately $0.055 - 0.064$ GFLOPs and $0.071 - 0.139M$ parameters). LightCF-Net proposed by Ji \textit{et al.} \cite{bioengineering11060545} is a novel and efficient lightweight architecture used as a long-range context fusion network for real-time polyp segmentation. A new FAEncoder module has been developed, merging Large Kernel Attention (LKA) with channel attention mechanisms to extract deep representational features of polyps and uncover long-range relationships. Furthermore, a novel Visual Attention Mamba Module (VAM) module has been integrated into skip connections to capture extensive contextual dependencies from encoder-extracted features.
Chen \textit{et al.} \cite{Chen2024MiMISTDMF} proposed MiM-ISTD for Infrared Small Target Detection (ISTD). It utilizes Mamba to effectively capture both local and global information from the given data. This approach ensures higher efficiency with very less computational costs. 

From this Section \ref{subsec:light}, UltraLight VM-UNet emerges as the efficient model in terms of both size and speed, using just 0.049M parameters and 0.060 GFLOPs. It achieves this through its Parallel Vision Mamba (PVM) design, which handles deep feature processing effectively. By combining Mamba with residual connections, the model captures long-range spatial relationships without adding extra complexity. In comparison, the Transformer-based models \cite{aghdam2022attentionswinunetcrosscontextual, ruan2023egeunetefficientgroupenhanced} are much larger and more demanding. This makes UltraLight VM-UNet a more practical choice for resource-constrained medical applications.

\subsection{Techniques and Adaptations}\label{sec:techniques_adaptations}

In this section, we explore techniques and adaptations for Mamba architectures such as weakly supervised, semi-supervised and self-supervised approaches. These approaches are used in scenarios where data annotations are absent, partially present or inconsistent, and used to improve the model's ability to learn from unstructured or incomplete or semi-structured data.
\subsubsection{Weakly Supervised Learning}\label{subsec:weakly}

Weakly Supervised Learning (WSL) uses a small amount of correctly labeled data along with a large amount of data with incomplete labels. Instead of having detailed labels for each data, this approach works with data that have noisy and partial labels. Wang \textit{et al.} \cite{wang2024weakmambaunetvisualmambamakes} proposes Weak-Mamba-UNet, a WSL strategy which incorporates three different architectures but with the same symmetric encoder-decoder networks. The three networks consist of a CNN based U-Net, known for capturing local features; a Swin Transformer-based SwinUNet, which excels in understanding global context and a VMamba-based Mamba-UNet, for efficiently capturing long-range dependency. The proposed WSL framework employs a multi-view cross-supervised learning approach for scribble-based supervised medical image segmentation. The work introduced partial cross-entropy to leverage only the scribble annotations during the training of the network. The overall loss is composed of the scribble-based partial cross-entropy loss and the dense-signal pseudo label dice-coefficient loss. 

\subsubsection{Semi-Supervised Learning}\label{subsec:semi}

Semi-supervised learning uses a fewer amount of labeled data and a larger amount of unlabeled data during training. Ma \textit{et al.} \cite{ma2024semimambaunetpixellevelcontrastivepixellevel} proposes Semi-Mamba-UNet, a semi-supervised learning framework integrated with a mamba-based segmentation network. It supports the complementary strengths of Mamba-UNet and UNet which uses labeled and a large amount of unlabeled data respectively. A Pixel-Level Contrastive learning strategy is proposed to increase feature learning from a pair of projectors. A network via pseudo labeling is used to train other network using a pixel-level cross-supervised learning strategy. The overall loss is the sum of three components: supervision loss, self-supervised contrastive loss, and semi-supervised loss. Semi-Mamba-UNet was tested on ACDC MRI Cardiac Dataset \cite{8360453} using 5\% and 10\% labeled data, along with the remaining unlabeled data. 

\subsubsection{Self Supervised Learning}\label{subsec:self}

 Self-supervised learning is a method in which the model learns from unlabeled data by creating its own labels. Instead of depending on external manually annotated labels, the model generates these labels internally based on the data itself. Nasiri \textit{et  al.} \cite{nasiri2024vim4path} proposes Vim4Path which uses Vision Mamba within DINO by Caron \textit{et al.} \cite{caron2021emerging} for representation learning. The research aims to explore and adapt ViM for use in SSL. DINO, a well known self-supervised learning framework employs self-distillation in a teacher-student setup, where both networks share identical architectures but have different parameters. The study compares two architectures in slide-level and patch-level classification tasks using the Camelyon16 dataset for benchmarking purposes. CLAM framework by Lu \textit{et al.} \cite{lu2020data} is employed to compare various architectures for slide-level classification. It utilizes attention-based multiple-instance learning, enabling the identification of sub-regions within slides that are most indicative of the slide-level label. This approach allows the model to focus on the most relavent features without the need for detailed annotation. 
Zhou \textit{et al.} \cite{zhou2024mgimultimodalcontrastivepretraining} propose MGI, a new multimodal model which uses genetic and image data. A self-supervised contrastive learning strategy is used during pre-training to align visual encoder and gene encoder on paired genetic and image data, allowing visual encoder to learn relevant features from a genetic perspective. 
Tang \textit{et al.} \cite{tang2024mambamimpretrainingmambastate} proposed MambaMIM, a 3D-UNet-based architecture for self-supervised learning. MambaMIM integrates 3D sparse convolution with Mamba blocks in the encoder. It introduces Selective Structured State Space Sequence Token-interpolation (S6T), which generates interpolated vectors between consecutive input vectors, $y_{i}$ and $y_{i+1}$. The interpolated sequence is processed through a linear layer before the decoder block. In the decoder, the unmasked sequence from the encoder is also interpolated using S6T, while sparse features are filled with learnable tokens, transforming them into dense features for upsampling.

Contrastive learning is a technique within self-supervised learning that focuses on learning representations by comparing pairs of data samples. Yang \textit{et al.} \cite{yang2024cmvim} proposes Contrastive Masked Vim autoencoder (CMViM) which is the efficient representation learning method for 3D multi-modal data. To reconstruct 3D masked multi-modal data, it incorporates ViM into mask encoder so that it can effectively capture long range dependencies in 3D medical data. To align multi-modal representations, it introduces intramodal and intermodal contrastive learning mechanisms. Intra-modal contrastive learning module is introduced to capture discriminative features within the same modality. Inter-modal contrastive learning mechanism is introduced to align cross-modality representations from different modalities. CMViM outperforms other state-of-the-art methods in Alzheimer's disease diagnosis.
Ma \textit{et al.} \cite{ma2024semimambaunetpixellevelcontrastivepixellevel} introduces a pixel level contrastive learning strategy for feature learning maximization from representations that are projected in a pair of projectors.

From this Section \ref{subsec:self}, we can see that the need for manual labeling is reduced by learning directly from unlabeled data. These models capture long-range dependencies in complex medical data such as images and gene sequences. Contrastive learning helps align features across different modalities, improving performance on classification and segmentation tasks without requiring detailed annotations. However, these models require careful design of architecture and large amounts of unlabeled data.
\subsubsection{Multimodal Learning}\label{subsec:multi}

 Xie \textit{et al.} \cite{xie2024fusionmambadynamicfeatureenhancement} introduced a U-Net-like architecture called Fusion Mamba which is designed for encoding multimodal images and then decoding them. Fusion Mamba fuses features from two different source images. The encoder part incorporates Dynamic Vision State Space (DVSS) block which contains Efficient State Space Module (ESSM). ESSM uses an Efficient 2D Selective Scan (ES2D), Learnable Descriptive Convolution (LDC) and Efficient Channel Attention (ECA) \cite{wang2020ecanetefficientchannelattention}. 
 Dynamic Feature Fusion Module (DFFM) in the encoder is used to fuse features between different modalities. 
 The decoder contains a patch-expanding block followed by two DVSS blocks. The combined features of DFFM act as skip connections for the decoder. The patch-expanding block can either use transposed convolution or bilinear upsampling of features. Finally, the fused image is obtained from each image modality. Zhou \textit{et al.} \cite{zhou2024mgimultimodalcontrastivepretraining} proposed MGI which is a multimodal approach used for aligning images and gene modalities using a pre-training approach similar to CLIP \cite{radford2021learningtransferablevisualmodels}. Mamba-based encoder is employed for both images and genes. Contrastive loss is applied between modality embeddings to generate a matrix similar to CLIP. For task-specific learning, the model integrates an attention integration module and a mask output module. 
Fang \textit{et al.} introduced GFE-Mamba \cite{fang2024gfemambamambabasedadmultimodal}, a model employing a multi-stage training strategy. First, they train a 3D GAN to convert MRI images to PET images. Afterward, latent representations from MRI and PET are concatenated with tabular data, where continuous features pass through a linear layer, and discrete features pass through an embedding layer, before being fed into a Mamba classifier. The representations from Mamba classifier and MRI and PET latents from 3D GAN undergo a pixel-level bi-cross attention operation, applying attention operation between Mamba classifier’s representation and MRI/PET latents respectively.

From this Section \ref{subsec:multi}, models with Mamba-based designs have shown clear strengths in combining different types of data such as medical images, genetic information and clinical records. Their ability to capture detailed and long-range patterns through components such as DVSS, CMFM and ECA allows for effective fusion of features, making them well suited for tasks such as image segmentation, disease classification and aligning multiple data sources, while keeping computational demands low compared to attention-based models \cite{vs2022imagefusiontransformer, maswinfusion2022multimodal,rehman2025implihatevid,dar2025explainable}. However, these models tend to be complex, often requiring multi-step training processes and significant effort to set up. In some cases, they still need considerable computing power and their reliance on large, well-annotated datasets can be a barrier in real-world medical environments where such data are not always available.

\subsection{Applications in Various Medical Domains}\label{sec:applications_various_domains}
In this section, we explore the application of mamba-based models across a range of medical tasks, including segmentation, classification, registration, and restoration. We also highlight its versatility through miscellaneous applications in medical imaging. Each subsection provides an overview of the task, followed by a discussion of relevant research papers that have utilized mamba-based models in these domains.
\subsubsection{Medical Image Segmentation}\label{subsec:seg}

Medical image segmentation is a technique used to identify and extract specific Regions Of Interest (ROI) from medical images such as tumors, lesions, tissues or organs. The objective is to divide the image into areas that share similar features including color, texture, brightness, and contrast. \autoref{tab:seg_overview} outlines the overview of segmentation models, parameters, descriptions and the code availability. \autoref{fig:segmentation architecture} shows the workflow of Mamba-based models in the medical image segmentation task. Some notable research works on mamba-based models for medical image segmentation include:

\begin{figure}[!ht]
\includegraphics[width=\linewidth]{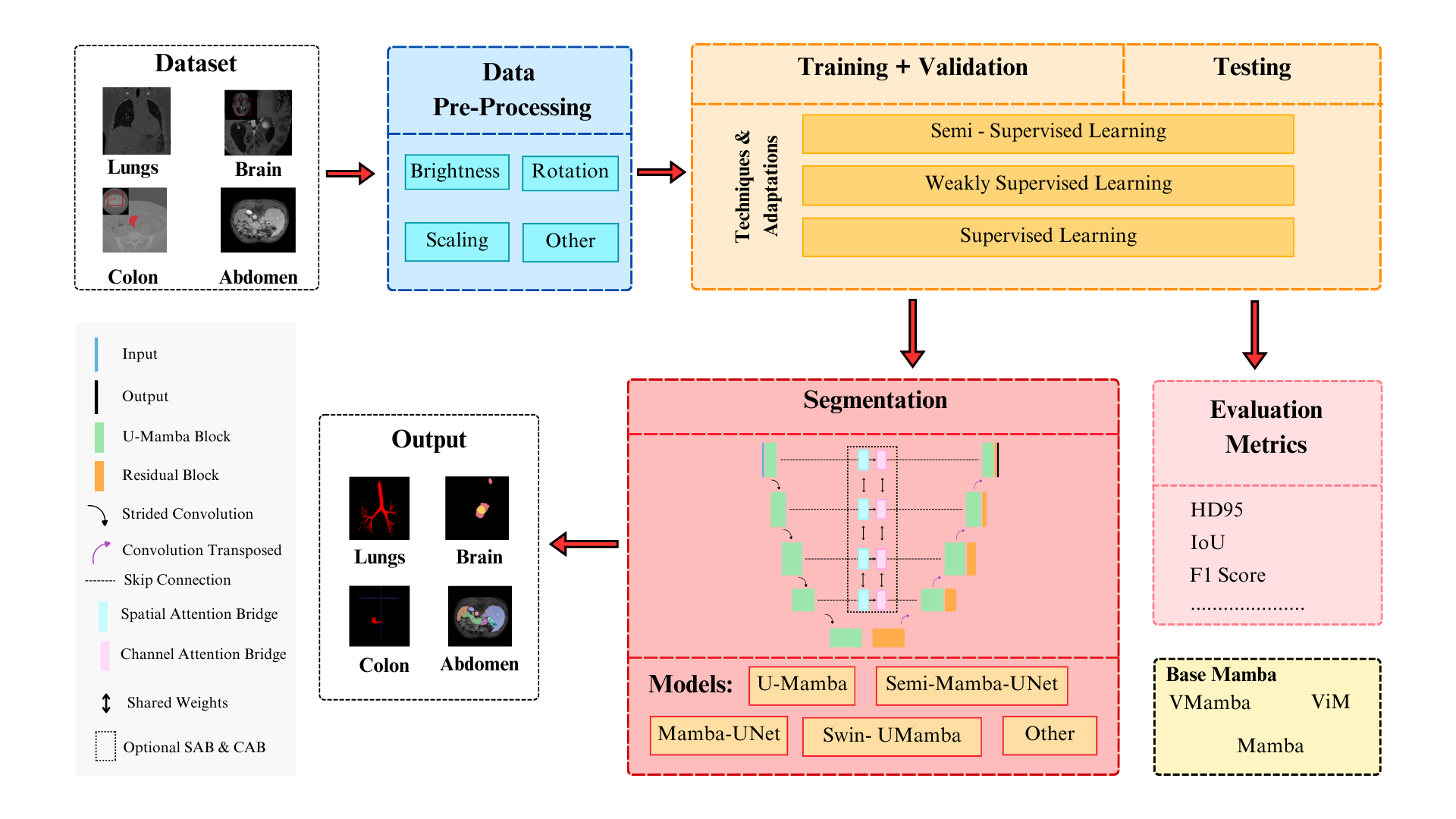}
\caption{\centering Work Flow of  Segmentation Task}
\label{fig:segmentation architecture}
\centering
\end{figure}

Xing \textit{et al.} \cite{xing2024segmambalongrangesequentialmodeling} proposed an architecture called \textbf{SegMamba} along with a new dataset comprising 500 3D computed tomography scans with expert annotations. The proposed model features an architecture similar to the transformer-based U-Net. Initially, it employs depthwise convolution with a kernel size of $7 \times 7\times 7$, padding of $3\times 3\times 3$, and a stride of $2\times 2\times 2$. On the encoder side, the model incorporates a series of TS-Mamba blocks and downsample blocks with residual connections mirrored on the upsample side. 
\begin{table}[!ht]
  \caption{Overview of Segmentation Models}
  \label{tab:seg_overview}
  \centering
  \fontsize{8.5pt}{8.5pt}\selectfont
  \setlength{\tabcolsep}{1.5pt} 
  \begin{tabular}{lcclc}
    \toprule
    Models & Params(M) & Core Mamba   & Description & Code\\
    \midrule
    {SegMamba \cite{xing2024segmambalongrangesequentialmodeling}}   & - & TSMamba & Multiple Tumor analysis & \href{https://github.com/ge-xing/SegMamba}{\Checkmark}\\
    {H-Vmunet \cite{wu2024hvmunethighordervisionmamba}}      & 8.97 & H-VSS & Skin lesion and polyp analysis& \href{https://github.com/wurenkai/H-vmunet}{\Checkmark}\\
    {LKM-UNet \cite{wang2024lkmunetlargekernelvision}}  & - &  Bi-Mamba& 2D \& 3D Multiple organ analysis & \href{https://github.com/wjh892521292/LKM-UNet}{\Checkmark}\\
    {Mamba-UNet \cite{wang2024mambaunet}}  & - & VMamba & Multiple organ analysis & \href{https://github.com/ziyangwang007/Mamba-UNet}{\Checkmark}\\
    {Weak Mamba UNet \cite{wang2024weakmambaunetvisualmambamakes}}  & -&VMamba  & Analysis of heart with WSL framework & \href{https://github.com/ziyangwang007/Weak-Mamba-UNet}{\Checkmark}\\
    {LightM-UNet \cite{liao2024lightmunetmambaassistslightweight}}  & 1.87& RVM & Tumor analysis in chest X-ray images & \href{https://github.com/MrBlankness/LightM-UNet}{\Checkmark}\\
    {UltraLight-VM UNet \cite{wu2024ultralightvmunetparallelvision}}  & 0.049& Mamba& Skin lesion analysis with light weight networks  & \href{https://github.com/wurenkai/UltraLight-VM-UNet}{\Checkmark}\\
    {T-Mamba \cite{hao2024tmamba}}  &  1.04&Tim & 2D \& 3D tooth analysis & \href{https://github.com/isbrycee/T-Mamba}{\Checkmark}\\
    {HC-Mamba \cite{xu2024hcmambavisionmambahybrid}}  & 13.88& Mamba& Skin lesion analysis & \XSolidBrush\\
    {Semi-Mamba-UNet \cite{ma2024semimambaunetpixellevelcontrastivepixellevel}} & -& VMamba & Analysis of heart with SSL framework & \href{https://github.com/ziyangwang007/Mamba-UNet}{\Checkmark}\\
    {ViM-UNet \cite{archit2024vimunet}}  & 18& ViM& Analysis of microscopic organs(cell)& \href{https://github.com/constantinpape/torch-em/blob/main/vimunet.md}{\Checkmark}\\
    {U-Mamba \cite{Ma2024UMambaEL}}  & -& Mamba& 2D \& 3D Multiple organ analysis   & \href{https://github.com/bowang-lab/U-Mamba}{\Checkmark}\\
    {Mamba-HUNet \cite{sanjid2024integratingmambasequencemodel}}  & -&VMamba & Analysis of Multiple Sclerosis Legion in brain& \XSolidBrush\\
    {UU-Mamba \cite{tsai2024uumamba}}  & - & Mamba  & Heart Analysis integrated with SAM optimizer & \XSolidBrush\\
    {CAMS-Net \cite{khan2024camsconvolutionattentionfreemambabased}}  &- & NC-Mamba & Heart region segmentation & \XSolidBrush\\
    {MUCM-Net \cite{yuan2024mucmnetmambapowereducmnet}}  & 0.047 & Mamba& Skin lesion analysis&\href{https://github.com/chunyuyuan/MUCM-Net}{\Checkmark}  \\
    \bottomrule

  \end{tabular}
\end{table}
Wu \textit{et al.} \cite{wu2024hvmunethighordervisionmamba} introduced \textbf{High-order Vision Mamba UNet (H-Vmunet)}, which is also similar to a transformer-based U-Net architecture. This work addresses challenges in skin lesion analysis (ISIC2017), spleen cancer detection, and polyp detection (CVC-ClinicDB dataset). The authors introduced H-VSS module, a modified version of VSS, replacing it with the H-SS2D module. In the H-SS2D module, the input is projected into $2C$ channels. The split of these channels depends on the order of the H-SS2D module. H1-SS2D splits the channels by half, while H3-SS2D splits them into $C/4$ and $7C/4$ for SS2D and Local-SS2D, respectively. 
The overall model architecture is based on U-Net, starting with two convolutional layers, followed by H-VSS modules of order 2, 3, 4, and 5 from layers 3 to 6, each followed by a convolutional layer with a kernel size of 3. The residual layers of this U-Net uses shared weights of the Channel Attention Bridge (CAB) and Spatial Attention Bridge (SAB) for information fusion between the encoder and decoder. On the decoder side, the model starts with H-VSS blocks of order 5, 4, 3, and 2 from layers 1 to 3, with the addition of features from SAB and CAB. The features are upsampled using bilinear interpolation. 

Wang \textit{et al.} \cite{wang2024lkmunetlargekernelvision} devised \textbf{Large Kernel Vision Mamba UNet (LKM-UNet)} for 2D and 3D segmentation tasks. The encoder of LKM-UNet uses depthwise convolution which is followed by LM block and a downsampling block. LM block consists of pixel-wise SSM and patch-level SSM, each followed by a bidirectional mamba block at each SSM level. The bidirectional Mamba block includes forward and backward SSMs which is similar to ViM. 
In the decoder, LKM-UNet uses convolution-based upsampling with residual connections from the corresponding encoder layers. 
Wang \textit{et al.} \cite{wang2024mambaunet} presented \textbf{Mamba-UNet}, a model that adapts a purely ViM-based U-Net architecture with linear embedding and VSS blocks. This network includes encoder, decoder, and bottleneck layers. Initially, medical images, defined by height and width, are converted into a 1D sequence. An embedding layer transforms them into a size denoted by \textbf{\textit{C}}, followed by VSS blocks. 
In decoder, features are reconstructed by upsampling, reducing  feature space by half and doubling the height and width. Skip connections are employed between the corresponding encoder and decoder layers. 
Wang \textit{et al.} \cite{wang2024weakmambaunetvisualmambamakes} proposed \textbf{Weak Mamba UNet}, which combines CNNs, ViTs, and Visual Mamba for WSL. This model uses a weighted mixture of the three architectures, each having a symmetric architecture with identical input and output sizes. For CNN-based model, the authors utilized  classical U-Net with convolution layers for downsampling. ViT-based UNet and Visual mamba-based UNet follow a similar symmetric architecture. Weak Mamba UNet uses scribble annotations from images, and all networks are evaluated using a pseudo-label loss function and a scribble loss. 

Liao \textit{et al.} \cite{liao2024lightmunetmambaassistslightweight} introduced \textbf{LightM-UNet}, which is an extension of the VSS module incorporating a Residual Vision Mamba (RVM) Layer. RVM includes layer normalization, followed by another layer normalization, and then the VSS module, with residual connections and scaling. This layer addresses long-term spatial dependencies. Each encoder block in LightM-UNet contains RVM and downsampling layers. 
The decoder blocks consist of depthwise convolution with residual connections scaled similarly to RVM layer, and bilinear transformation is used to restore the predictions to the original resolution. 
Wu \textit{et al.} \cite{wu2024ultralightvmunetparallelvision} devised \textbf{UltraLight-VM UNet}, which employs Parallel Vision Mamba (PVM). Each PVM consists of $N$ independent Mamba layers, with input representations divided into $N$ parts along the channel dimension $(C/N)$. The authors provided four configurations with 1, 2, and 4 Mamba layers, respectively. Similarly to LightM-UNet, Mamba layers in UltraLight-VM UNet use scaled residual connections. All representations from Mamba layers are concatenated, followed by layer normalization and a projection layer, which allows for customizable downsampling of feature space. 
The encoder layers consist of convolution layers for stages 1-3 and PVM layers for stages 4-6. Residual connections incorporate Spatial Attention Bridge (SAB) and Channel Attention Bridge (CAB) layers, differing from LightM-UNet. The decoder blocks mirror the encoder structure and use bilinear transformation to restore predictions to the original resolution. 
Hao \textit{et al.} \cite{hao2024tmamba} presented an architecture called \textbf{T-Mamba}. It is a modification of the DenseVNet architecture, incorporating sequential layers of convolutional networks, batch normalization, and the Tim layer. In T-Mamba, Tim layer flattens the input, applies shared position embedding (similar to sinusoidal position embedding), normalizes representations using layer norm, and passes them to a modified ViM block, which includes forward and backward SSM similar to Vision Mamba. 
The encoder part, inspired by DenseVNet, connects and stacks each layer repeatedly, followed by downsampling up to three layers. Subsequent encoder layers are upsampled by $2\times$, $2\times$, and $4\times$, respectively, leading to a prediction head for segmentation. 
Xu \textit{et al.} \cite{xu2024hcmambavisionmambahybrid} proposed \textbf{HC-Mamba}, which uses HC-SSM block composed of a series of HC-Conv blocks and modified Mamba block using HC-Conv. The input is split into two parts: the first part undergoes a series of HC-Conv blocks, where HC-Conv involves a two-step convolution process with dilated convolution followed by depth-wise separable convolution. The second part uses HC-Convolution block instead of the traditional convolution block in Mamba. The representations from both parts are concatenated and shuffled randomly. 
The decoder consists of HC-Mamba blocks with patch merging until the final projection. 


Ma \textit{et al.} \cite{ma2024semimambaunetpixellevelcontrastivepixellevel} introduced \textbf{Semi-Mamba-UNet}, which leverages self-supervised learning methodology by implementing two networks: Mamba-UNet and UNet. 
In a semi-supervised learning setting, Semi-Mamba-UNet employs three combinations of loss functions: contrastive loss, semi-supervised loss, and supervised loss. Pixel-level contrastive loss is calculated using a projection layer in the network's final representations. The loss is determined by taking the $L2$ norm of the representations in the labeled and unlabeled data, normalized by the number of input data ($N$). Semi-supervised loss is applied to the final layers using cross-entropy loss and dice loss, with pseudo labels predicted by the network. Supervised loss uses same loss functions but only on the labeled data. 
Archit \textit{et al.} \cite{archit2024vimunet} utilized Vision Mamba for cell structure segmentation, achieved better performance compared to both transformer and Convolution based approaches on U-Net frameworks. However, the \textbf{ViM-UNet} is not a modified version of a existing Vision Mamba architectures but rather implements Vision Mamba along with corresponding downsample and upsample layers to create a U-Net architecture. 
The paper proposed two variations of ViM-UNet: one with a smaller size, termed ``Tiny'' with 18 million parameters, and another with a larger size, termed ``Large'' with 39 million parameters.

Ma \textit{et al.} \cite{Ma2024UMambaEL} proposes a network called \textbf{U-Mamba} for 2D and 3D biomedical image segmentation. 
UMamba encoder is composed of building blocks, consisting of two successive residual blocks, followed by SSM Mamba, while decoder is composed of residual blocks with skip connections. U-Mamba incorporates self-configuring feature from nnU-Net, and number of network blocks is automatically determined based on the datasets. Additionally, the authors propose two variants of UMamba, namely, ``U-Mamba\_Bot,'' which integrates U-Mamba block only in the bottleneck, and ``U-Mamba\_Enc,'' which utilizes Mamba blocks across all encoder layers. 
\textbf{Mamba-HUNet} proposed by Sanjid \textit{et al.} \cite{sanjid2024integratingmambasequencemodel} is a novel architecture that has been designed for reliable and effective medical image segmentation. 
This design greatly improves processing efficiency by mutating Hierarchical Upsampling Network (HUNet) into Mamba-HUNet and optimizing it into a more efficient variety without sacrificing performance. To enable efficient processing, input grayscale pictures are divided into patches and converted into $1D$ sequences, a method influenced by ViT and Mamba. 
Tsai \textit{et al.} \cite{tsai2024uumamba} introduced \textbf{UU-Mamba}, a segmentation model that incorporates an uncertainty-aware loss function. This loss function is a combination of three components: dice coefficient loss (region-based), cross-entropy loss (distribution-based), and focal loss (pixel-based), all combined with a learnable parameter sigma. To address the issue of narrow minima in the uncertainty loss, SAM optimization introduces a hyperparameter epsilon to ensure the minima are flatter, thus preventing the network from overly fitting complicated representations. 
Each U-Mamba block includes two residual connections: one with a Mamba block followed by a convolution block and Instance Normalization (IN). 

Khan \textit{et al.} \cite{khan2024camsconvolutionattentionfreemambabased} proposed a convolution and attention-free segmentation mamba based model named \textbf{CAMS-Net}. The authors proposed NC-Mamba block, which differs from normal Mamba block by not using $1D$ convolution. Instead, so the representations are directly projected into SSM with SiLU activation function. 
When using LIFM block, representations of height and width are combined into a single dimension and later reshaped to their original form. Therefore, LIFM block operates on two levels: Mamba Channel Aggregator (MCA), which works on channel level, and Mamba Spatial Aggregator (MSA), which operates on the combined height and width dimension. 
The initial part of segmentation network includes an MCA block combined with sinusoidal position embedding. Downsampling in encoder is performed using $2 \times 2$ average pooling. CS-IF blocks are employed in bottleneck and last encoder part. The decoder also uses MCA blocks, with upsampling performed by $2 \times 2$ bi-linear transformation. 
Yuan \textit{et al.} \cite{yuan2024mucmnetmambapowereducmnet} presented U-Net-based segmentation network tailored for skin lesion segmentation. This architecture differs from other mamba based U-Nets by omitting SAB and CAB blocks in skip connections. It features six encoder layers followed by downsampling with convolution layers, and six decoder layers with upsampling through convolution. The initial and final layers of both encoder and decoder utilize convolution layers. Aside from these initial and final layers, \textbf{MUCM-Net} integrates UCM-Net blocks and Mamba blocks by adding their representations together.

Section~\ref{subsec:seg} focuses on Mamba-based models such as LightM-UNet~\cite{liao2024lightmunetmambaassistslightweight} and Mamba-UNet~\cite{wang2024mambaunet}. Comparing with Transformer-based models such as Swin UNETR~\cite{hatamizadeh2021swin} and Swin-UNet~\cite{cao2022swin}.  LightM-UNet incorporates a Vision Mamba (RVM) layer with residual connections to reduce model complexity, while Swin UNETR relies on Swin Transformer blocks. On the LiTS~\cite{BILIC2023102680} dataset, LightM-UNet used 1.87M parameters and 457.62 GFLOPs, whereas Swin UNETR required 61.99M parameters and 1570.32 GFLOPs. Similarly, on the Montgomery \& Shenzhen~\cite{article} datasets, LightM-UNet used just 1.09M parameters and 267.19 GFLOPs, compared to Swin UNETR’s 25.12M parameters and 909.26 GFLOPs. This shows that LightM-UNet is more efficient even with the low computation without reducing the performance. For multi-organ segmentation, Mamba-UNet was evaluated on the ACDC~\cite{8360453} and Synapse CT datasets. 
While comparing Mamba-UNet with Swin-UNet,the results shows improved performance: on ACDC, Mamba-UNet achieved a Dice score of 0.9281 and Hausdorff Distance (HD) of 2.464, compared to Swin-UNet's 0.9188 and 3.1817, respectively. On Synapse CT, Mamba-UNet scored 0.6429 Dice and 24.47 HD, outperforming Swin-UNet’s 0.6178 and 30.54. These outcomes highlight the ability of Mamba-based architectures to provide better accuracy with a simple architecture.

\subsubsection{Medical Image Classification}\label{subsec:clas}

\begin{figure}[!ht]
\includegraphics[width=\linewidth]{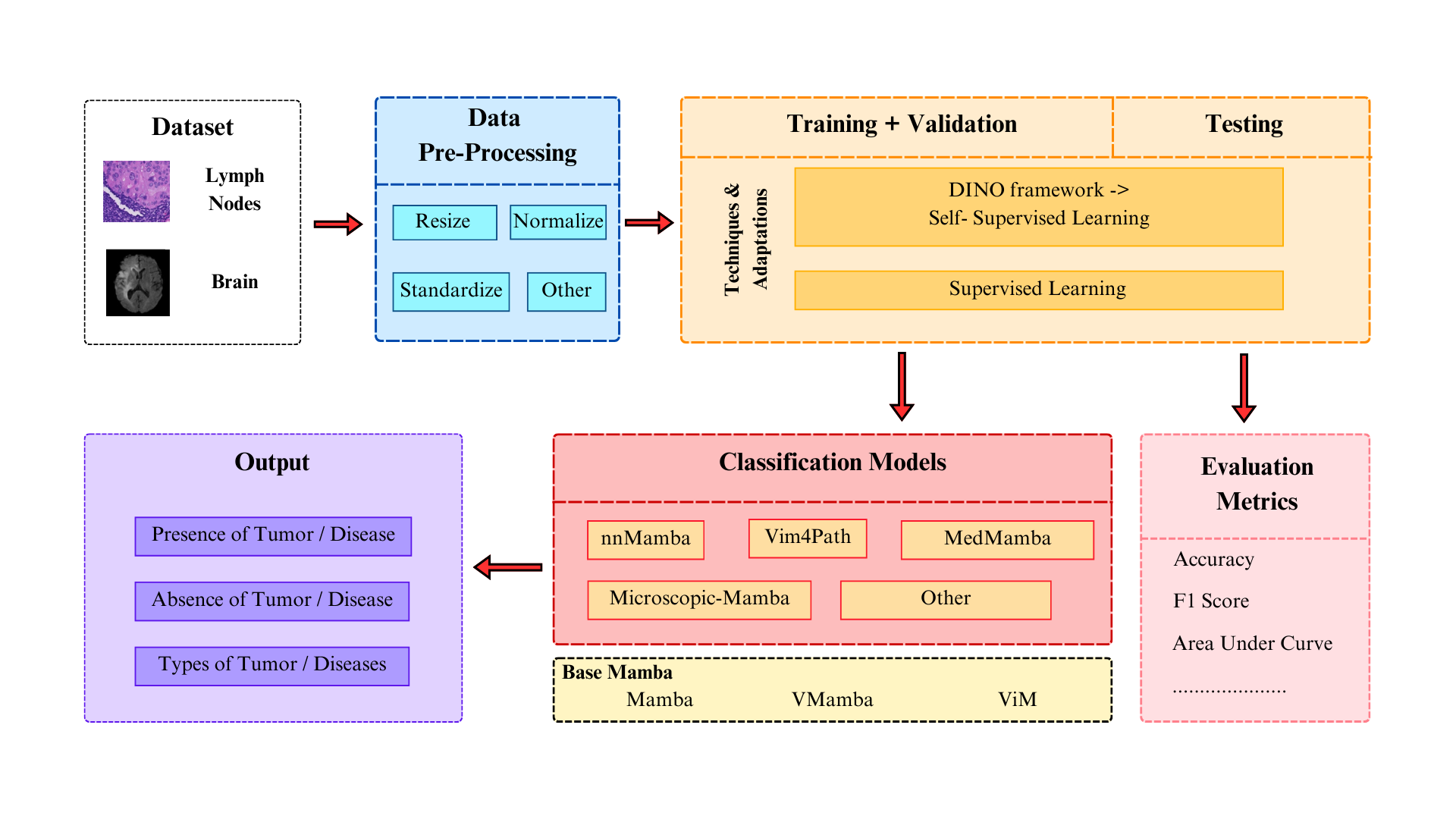}
\caption{\centering Work Flow of Classification Task}
\label{fig:Classification architecture}
\centering
\end{figure}

Classification in medical imaging refers to categorizing images into different classes, such as distinguishing between benign and malignant lesions, or identifying different types of diseases. \autoref{fig:Classification architecture} shows the workflow of Mamba-based models in medical image classification task. \autoref{tab:cla_overview} summarizes classification models and their parameters, along with descriptions and information regarding code availability. Some mamba-based architectures applied in this domain include: 
Gong \textit{et al.} \cite{gong2024nnmamba} introduced \textbf{nnMamba} method for medical image classification that integrates strengths of CNNs and SSMs. Traditional CNNs lack in capturing long-range dependencies due to their local receptive fields, while transformers, though capable of modeling global context, are computationally intensive for 3D medical images. nnMamba addresses these challenges by incorporating Mamba-In-Convolution with Channel-Spatial Siamese (MICCSS) block, which effectively combines long-range dependency modeling and local feature extraction . nnMamba introduces channel-scaling and channel-sequential learning. Channel-scaling adjusts the importance of different feature channels, while channel-sequential learning processes features in a sequential manner to capture complex dependencies. nnMamba’s backbone architecture leverages MICCSS block to maintain computational efficiency while achieving superior performance in 3D medical imaging tasks.

\begin{table}[!ht]
  \caption{Overview of Classification Models}
  \label{tab:cla_overview}
  \centering
  \fontsize{8.5pt}{9pt}\selectfont
  \setlength{\tabcolsep}{1.5pt} 
  \begin{tabular}{lcclc}
    \toprule
    Models & Params(M) & Core Mamba   & Description & Code\\
    \midrule
    {nnMamba \cite{gong2024nnmamba}}   &  15.55 & Res-Mamba & Alzheimer prediction and Landmark detection& \href{https://github.com/lhaof/nnMamba}{\Checkmark}\\
    {MedMamba \cite{yue2024medmambavisionmambamedical}}      & 15.2  & VMamba & Multiple disease classification & \href{https://github.com/YubiaoYue/MedMamba}{\Checkmark}\\
    {Vim4Path \cite{nasiri2024vim4path}}  & 7 &  ViM& Breast cancer prediction in WSI & \href{https://github.com/AtlasAnalyticsLab/Vim4Path}{\Checkmark}\\
    Microscopic-Mamba \cite{zou2024microscopicmambarevealingsecretsmicroscopic}& 1.59 & PEVM & Microscopic image classification & \href{https://github.com/zs1314/Microscopic-Mamba}{\Checkmark}\\
    \bottomrule

  \end{tabular}
\end{table}

Yue \textit{et al.} \cite{yue2024medmambavisionmambamedical} propose \textbf{MedMamba}, a novel method for medical image classification that leverages modern SSM, particularly inspired by VMamba. 
The architecture of MedMamba includes a patch embedding layer, SS-Conv-SSM blocks, and patch merging layers. 
These blocks use a dual-branch approach to separately process and merge features from convolutional and SSM pathways, incorporating a 2D-selective scan (SS2D) for comprehensive feature extraction. 
The study demonstrates the potential of SSM-based models in medical applications along with the outline of future research directions which includes optimization and integration of explainable AI. 
Nasiri \textit{et al.} \cite{nasiri2024vim4path} used Vision Mamba (ViM) in its architecture for learning representations in histopathology images. \textbf{Vim4path} performs self supervised learning (SSL) within DINO framework \cite{caron2021emerging}. It extracts image patches from Whole Slide Images (WSIs) and trains ViM encoder using DINO without the availability of labels. 
The paper highlights potential of ViM in practical diagnostic applications and its effectiveness in less computation scenarios which establishes it as a promising tool for computational pathology.
\textbf{Microscopic-Mamba} proposed by Zou \textit{et al.} \cite{zou2024microscopicmambarevealingsecretsmicroscopic} consists of following components in sequence: patch embedding, four blocks of Hybrid-Conv-SSM each followed by a patch merging block, Global pooling, $1\times1$ convolution (PW convolution) and finally a fully connected layer for classification. Hybrid-Conv-SSM block splits the representation channel wise which is then passed through two different batches. First branch is the Conv branch which contains depth wise and point wise convolution. Second branch is SSM branch which contains Parallel Efficient Vision Mamba (PEVM) block. 
Microscopic-Mamba has three variants in terms of parameters such as tiny, small, base. Tiny has 4.32M parameters, small has 4.97M parameters and base has 8.37M parameters. 

From this section \ref{subsec:clas}, we can see that MedMamba \cite{yue2024medmambavisionmambamedical} was evaluated on various datasets using model variants like tiny, small, and base similar to Swin Transformer \cite{liu2021swin}. 
On the PAD-UFES-20 \cite{pacheco2020padufes20skinlesiondataset} dataset, Swin-T had a slightly better AUC of 0.830 compared to MedMamba-T’s 0.808, but it used more than twice the FLOPs (4.5G vs. 2.0G) and nearly double the parameters (27.5M vs. 14.5M). On the Cervical-US dataset, MedMamba-T reached a higher AUC of 0.952 compared to Swin-T’s 0.890, while staying more efficient. Similar results were seen on CPX-Ray \cite{shastri2022cheximagenet}, Kvasir \cite{pogorelov2017kvasir} , and Fetal-Planes-DB \cite{burgos2020evaluation}, where MedMamba models gave better performance with fewer resources. while Swin Transformer sometimes achieves higher accuracy but at a higher computational cost.

\subsubsection{Medical Image Restoration/ Reconstruction}\label{subsec:res}
Restoration is an application in medical imaging which is used to improve the quality of images that may be corrupted or distorted due to factors such as noise, low resolution and blurring. Reconstruction is a mathematical process that converts raw medical data into target image. \autoref{tab:res_overview} provides an overview of Mamba based models applied in medical image restoration/reconstruction along with parameters, descriptions and the availability of code. 
Key Mamba-based architectures used in this application are explained as follows.

\begin{table}[!ht]
  \caption{Overview of Restoration Models}
  \label{tab:res_overview}
  \centering
  \fontsize{8.5pt}{9pt}\selectfont
  \setlength{\tabcolsep}{1.5pt} 
  \begin{tabular}{lcclc}
    \toprule
    Models & Params(M) & Core Mamba  & Description & Code\\
    \midrule
    {MambaMIR \cite{huang2024mambamirarbitrarymaskedmambajoint}}   &  - & Mamba & MRI-CT image reconstruction & \href{https://github.com/ayanglab/MambaMIR}{\Checkmark}\\
    {FDVM-Net \cite{zheng2402fdvisionmamba}}  & - &  Mamba& Endoscopic image reconstruction & \href{https://github.com/zzr-idam/FDVM-Net}{\Checkmark}\\
    {MambaDFuse \cite{li2024mambadfuse}}  & - & Mamba & Fused  reconstruction with MRI,CT,PET images & \href{https://github.com/Lizhe1228/MambaDFuse}{\Checkmark}\\
    {FusionMamba \cite{xie2024fusionmambadynamicfeatureenhancement}}  & -& DVSS & Dynamic Feature Enhancement and Image fusion & \href{https://github.com/millieXie/FusionMamba}{\Checkmark}\\
    \bottomrule
     
  \end{tabular}
\end{table}

Zheng \textit{et al.} \cite{zheng2402fdvisionmamba} presents \textbf{FDVM-Net} which is designed for endoscopic exposure correction and it leverages frequency-domain reconstruction to achieve high-quality image restoration. To capture spatial features and global dependencies, FDVM-Net combines Mamba and convolutional blocks. This combination serves as the foundation for a dual-path network architecture, which separates processing of the image's phase and amplitude information. Furthermore, FDVM-Net applies a frequency domain cross-attention module to improve the performance of network.

Huang \textit{et al.} \cite{huang2024mambamirarbitrarymaskedmambajoint} introduced \textbf{MambaMIR} for medical image reconstruction and uncertainty estimation. 
MambaMIR incorporates an Arbitrary-Masked State Space (AMSS) block with Monte Carlo dropout. AMSS is inspired by Mamba which comprises of AMS6 block, a gating linear layer, depth-wise convolution layer, and SiLU activation function. Residual connections used in MambaMIR ensures efficient and stable training. 
\textbf{MambaDFuse} introduced by Li \textit{et al.} \cite{li2024mambadfuse} incorporates Multimodal Mamba blocks which effectively merges features from different modality into a unified representation. This helps to obtain an informative fused image as output. MambaDFuse is a dual-phase model that integrates complementary information from diverse imaging modalities into a single image. MambaDFuse utilizes the combined strengths of convolutional layers and Mamba blocks to capture a spectrum of features. It ranges from low-level details to high-level semantic information.

Xie \textit{et al.} \cite{xie2024fusionmambadynamicfeatureenhancement} proposed \textbf{FusionMamba} to address limitations of channel redundancy and limited local enhancement capabilities in existing image fusion methods. FusionMamba uses dynamic convolution and channel attention mechanisms to enhance the model's ability to capture global context and local features within the images. FusionMamba leverages strengths of Mamba, dynamic feature enhancement and cross-modal fusion techniques. FusionMamba explores internal features and relationships between different image modalities. This results in dynamically enhanced texture details, better differences between modalities and improved ability to capture correlations while simultaneously reducing redundant information.


Section~\ref{subsec:res} presents the performance of the Mamba-based models for medical image restoration. To compare its effectiveness with transformer models such as SwinF~\cite{maswinfusion2022multimodal}, which uses Swin Transformer blocks for attention-guided cross-domain fusion. On the MRI-CT dataset, MambaDFuse achieved higher entropy (4.80 vs. 4.03) and slightly better structural similarity (SSIM: 0.67 vs. 0.65), while SwinF yielded a marginally higher visual information fidelity (VIF: 0.66 vs. 0.64). On the MRI-PET dataset, MambaDFuse again outperformed SwinF with higher entropy (4.91 vs. 3.90), VIF (0.65 vs. 0.63), and SSIM (0.52 vs. 0.51). For the MRI-SPECT dataset, both models achieved the same VIF (0.81), but MambaDFuse showed slight improvements in SSIM (0.64 vs. 0.63) and entropy (3.99 vs. 3.90). These results show that MambaDFuse performs slightly better than SwinF, offering better detail preservation and structure similarity across datasets while using Mamba's lightweight architecture.

\subsubsection{Medical Image Registration}\label{subsec:reg}

Image registration is the process involves aligning two or more images of the same scene captured at different times, from different viewpoints, or by different sensors. In medical image registration, this typically involves aligning a fixed volume, such as an intraoperative CT scan with a moving volume, such as a preoperative MRI to accurately overlay anatomical structures from different imaging modalities. This step is crucial for diagnosis, treatment planning, and monitoring in medical applications. 
Deformable registration refers to the alignment of images using transformation models that allow for local deformations. 

\begin{table}[!ht]
  \caption{Overview of Registration Models}
  \label{tab:reg_overview}
  \centering
  \fontsize{10pt}{9pt}\selectfont 
  \setlength{\tabcolsep}{1.5pt} 
  \begin{tabular}{lccccc}
    \toprule
    Models & Params(M)  & Core Mamba  & Description & Code\\
    \midrule
    {MambaMorph \cite{Guo2024MambaMorphAM}}   &  7.59 & Mamba & 3D MRI-CT registration & \href{https://github.com/Guo-Stone/MambaMorph}{\Checkmark}\\
    {VMambaMorph \cite{Wang2024VMambaMorphAM}}      & 9.64  & VMamba  & 3D MRI-CT registration  & \href{https://github.com/ziyangwang007/VMambaMorph}{\Checkmark}\\
    \bottomrule
     
  \end{tabular}
\end{table}
Deformable registration can handle complicated transformations which makes it appropriate for aligning MR and CT scans when the patient's anatomy may have changed between scans.  \autoref{tab:reg_overview} describes  registration models and their parameters along with information on their descriptions and code availability. 
Some mamba-based models relevant to this area include:

Guo \textit{et al.} \cite{Guo2024MambaMorphAM} proposed a multi-modality deformable registration framework designed to process the medical MR-CT deformable registration dataset. Inspired by Chen \textit{et al.} \cite{Chen_2022}, which utilizes Swin Transformers in encoder of registration module, \textbf{MambaMorph} instead uses Mamba blocks to capture long-range spatial relationships while optimizing memory utilization. 
Before registration module, a simple fine-grained feature extractor (U-Net with one downsampling step) is utilized to maximize retention of local information. 

Wang \textit{et al.} \cite{Wang2024VMambaMorphAM} transformed the 2D VSS of VMamba into a 3D volumetric feature processing framework. This hybrid architecture combined VMamba with CNN (U-Net), to accurately estimate the deformation field. To address the challenges posed by complex motion and diverse structures in image registration, \textbf{VMambaMorph} introduces a recursive registration framework. Furthermore, it employs a weight-sharing fine-grained feature extractor to extract features across divergent volumes. MambaMorph's inadequacy in fully exploiting visual features from complex motion and diverse structures of images or volumes has been surpassed by VMambaMorph.

From this section \ref{subsec:reg}, we can understand that VMambamorph is better than MambaMorph. While comparing VMambaMorph \cite{Wang2024VMambaMorphAM} with the transformer-based model TransMorph \cite{Chen_2022} on the SynthRAD dataset, VMambaMorph outperformed TransMorph in both Dice Score and Hausdorff Distance (HD). Specifically, TransMorph achieved a Dice score of 82.31\% and a HD of 1.39, whereas VMambaMorph achieved a higher Dice score of 82.94\% and a lower HD of 1.35, showing better overlap and reduced boundary error.


\subsubsection{Miscellaneous}\label{subsec:mis}

Beyond primary tasks like segmentation, classification, and image restoration discussed in earlier sections, Mamba-based models have also been applied to a wide range of medical applications. These tasks include video analysis, medical language understanding, and image generation using diffusion models. In all these cases, Mamba helps to improve efficiency and better capture long-range relationships in data. 
\autoref{tab:mis_overview} outlines models and their parameters along with their descriptions and code availability. Notable mamba-based research papers in this field include

Xie \textit{et al.} \cite{xie2024promamba} introduced \textbf{Prompt-Mamba} for polyp segmentation in colonoscopy images. Polyp segmentation is crucial for early cancer detection but it is challenging due to variations in size, shape, and color. 
Prompt-Mamba addresses existing limitations such as various sample sizes and generalization to unseen data by combining Vision-Mamba which act as an image feature extractor with prompt technology for improved generalization. The lightweight architecture includes an image encoder with ViM layers, a prompt encoder and a mask decoder. It uses combination of Focal and Dice loss to handle class imbalance and measure segmentation quality. 
\begin{table}[!ht]
  \caption{Overview of Miscellaneous Models}
  \label{tab:mis_overview}
  \centering
  \fontsize{8.5pt}{9pt}\selectfont 
  \setlength{\tabcolsep}{1.5pt} 
  \begin{tabular}{lcclc}
    \toprule
    Models & Params(M) & Core Module   & Description & Code\\
    \midrule
    {P-Mamba \cite{ye2024pmambamarryingperonamalik}}   &  52.77 & ViM & Pediatric cardiac imaging  & \XSolidBrush \\
    {Prompt-Mamba \cite{xie2024promamba}}      & 102  & ViM & Polyp analysis with prompt technologies & \XSolidBrush\\
    {ClinicalMamba \cite{yang2024clinicalmamba}}      & -  & Mamba & Language model with prompt based fine-tuning & \href{https://github.com/whaleloops/ClinicalMamba}{\Checkmark}\\
    {Vivim \cite{yang2024vivimvideovisionmamba}}      & -  & ST-Mamba & Video analysis of multiple tumor & \href{https://github.com/scott-yjyang/Vivim}{\Checkmark}\\
    {VM-DDPM \cite{ju2024vmddpmvisionmambadiffusion}}      & 15.2  & VMamba & Synthesis of  MRI-X ray images & \XSolidBrush\\
    {MD-Dose \cite{fu2024mddose}}  & - & Mamba & Predicts dosage distribution for tumor patients & \href{https://github.com/LinjieFu-U/mamba_dose}{\Checkmark}\\
    {Motion-guided dual-}      & - & Bi- Mamba &  Real-time endoscope tip tracking & \href{https://github.com/PieceZhang/MotionDCTrack}{\Checkmark}\\
    {camera tracker \cite{zhang2024motionguideddualcameratrackerlowcost}} & \\
    \bottomrule
     
  \end{tabular}
\end{table}
Yang \textit{et al.} \cite{yang2024clinicalmamba} proposed \textbf{ClinicalMamba}, a specialized language model based on Mamba. ClinicalMamba is pre-trained on longitudinal clinical notes from MIMIC-III to achieve notable speed and performance benchmarks. It surpasses established clinical language models as well as large language models such as GPT-4 in longitudinal clinical tasks. The researchers pre-trained ClinicalMamba using a causal language modeling objective on a dataset of de-identified clinical notes from MIMIC-III. 
By specializing in medical domain and pre-training on longitudinal data, ClinicalMamba captures unique characteristics of clinical narratives. 

Ye \textit{et al.} \cite{ye2024pmambamarryingperonamalik} devised \textbf{P-Mamba}, a novel deep learning model to address pediatric echocardiographic left ventricular segmentation. It tackles two key issues: computational efficiency and noise interference. The model employs a dual-branch architecture. ViM encoder branch focuses on efficiency by capturing global dependencies in the image. DWT-based PMD encoder branch, leveraging a technique originally used for image de-noising (DWT-based PMD) block, specifically targets noise suppression while preserving local features crucial for segmentation. This branch utilizes an anisotropic diffusion equation to achieve this balance. Finally, decoders within the model (SegHead and FCNHead) upsample features from both branches to generate segmentation masks. 
Fu \textit{et al.} \cite{fu2024mddose} introduced \textbf{MD-Dose} for predicting radiation therapy dose distributions using a Mamba-based diffusion model. MD-Dose applies Mamba within diffusion model for both denoising and encoding tasks. The forward process of MD-Dose involves adding Gaussian noise to dose distribution maps, while the reverse process uses a noise predictor to reconstruct the original maps from noise. Mamba encoder in MD-Dose is used to extract structural information from CT images and then integrating it with noise prediction to enhance the localization of dose regions in Planning Target Volume (PTV) and Organs At Risk (OAR). 

Yang \textit{et al.} \cite{yang2024vivimvideovisionmamba} proposed \textbf{Vivim}, a novel deep learning framework 
for medical video object segmentation. Vivim addresses challenges of capturing spatiotemporal information and handling long-range dependencies in medical videos. The core of Vivim lies in its Video Vision Mamba (ST-Mamba) module, which leverages a transformer architecture with a selective scan mechanism inspired by SSMs. ST-Mamba efficiently captures long-range temporal dependencies within video sequences. The hierarchical encoder utilizes temporal Mamba blocks to extract multi-scale feature sequences, while a lightweight CNN-based decoder head fuses these features and predicts segmentation masks. 
The paper also introduces a new benchmark dataset VTUS which includes annotated thyroid ultrasound videos. 
Ju \textit{et al.} \cite{ju2024vmddpmvisionmambadiffusion} proposed \textbf{VM-DDPM}, a U-Net based architecture which performs forward noising and reverse denoising process as proposed in Denoising Diffusion Probabilistic Models (DDPM) \cite{ho2020denoisingdiffusionprobabilisticmodels}. The authors introduced SSLayer in this architecture which contains residual block and modification of VSS block called Multi-level State Space (MSS) block. MSS block incorporates time step embedding for diffusion process and Cross-Scan
Module (CSM) which is a modification of S6 scan.
Zhang \textit{et al.} \cite{zhang2024motionguideddualcameratrackerlowcost} introduced a low-cost dual-camera tracker for endoscopy skill evaluation in mechanical simulators. The tracker solves problems faced by existing methods such as lack of consistency and high cost. It achieves accurate and reliable 3D endoscope tip position feedback. The core of the system lies in cross-Camera Mutual Template (CMT) strategy that utilizes information from both cameras to maintain tracking consistency. Motion-Guided Prediction Head (MMH) based on Mamba integrates historical motion data with visual tracking. 
Furthermore, the tracker includes a Vision-Motion integrator that combines motion information with visual features for improved 3D localization. This integration is done by a mechanism named Multi-KV cross-attention. 

\section{Datasets} \label{sec: datasets}
Application in medical domain depends on a wide variety of datasets to improve research and innovation in the field of medical imaging, diagnostics and treatment planning. These datasets cover multiple medical specialties and imaging modalities. In this section, we offer a comprehensive overview of some of the most significant datasets utilized in Mamba-based models across various medical fields. \autoref{tab:datasets} provides a detailed summary of various datasets used in medical imaging, specifying tasks performed and targeted areas, which includes both anatomical organs and surgical instruments. It highlights diversity of dataset in medical applications, offering insight into how different challenges in medical diagnostics and treatment are addressed through targeted imaging and analysis. 

\begin{table*}[!ht]
  \caption{ Overview of Datasets, Tasks Performed, and Targeted Areas in Medical Imaging and Analysis}
  \label{tab:datasets}
  \centering
  \fontsize{8pt}{9pt}\selectfont 
  \setlength{\tabcolsep}{1pt} 
  \begin{tabular}{llll}
    \toprule
    Dataset & Task Performed  & Targeted Area & Description\\
    \midrule
    BraTS2023 \cite{4b589b6824a64a2a91e8e3b26cc0bf9e, kazerooni2024braintumorsegmentationpediatrics, Bakas2017AdvancingTC}  & Segmentation   & Brain & Collection of multi-parametric MRI scans\\
    
    AIIB2023 \cite{nan2022fuzzyattentionneuralnetwork, Li2022HumanTT} & Segmentation & Lungs & Consists of 120 high-resolution CT scans \\

    CRC-500 Dataset \cite{xing2024segmambalongrangesequentialmodeling} & Segmentation & Colon, Rectum & 500 meticulously annotated 3D CT scans\\
    ISIC2017 \cite{8363547} & Segmentation &  Skin & 2,000 training images across 3 components\\
    ISIC2018 \cite{codella2019skinlesionanalysismelanoma, Tschandl_2018} & Segmentation & Skin & 12,500 images distributed across three tasks\\
    MICCAI-2023 \cite{lyu2024stateoftheartcardiacmrireconstruction} & Segmentation & Heart & Contains 120 training, 60 validation and\\
    & & & 120 test data \\

    Spleen \cite{Antonelli_2022, simpson2019largeannotatedmedicalimage} & Segmentation & Spleen & Includes 61 portal-venous phase CT scans \\

    CVC-ClinicDB \cite{PMID:25863519} & Segmentation & Intestine & Comprises 612 high-resolution images \\

    3D Abdomen CT dataset \cite{FLARE22} & Segmentation & Liver, Kidney,& Consists of 2,300 CT scans\\
    & &  Spleen, etc. & \\
    2D Abdomen MRI Dataset \cite{ji2022amos} & Segmentation & Kidney, Pancreas,  &  Contains 100 MRI scans and 500 CT scans \\
    & &  Gallbladder, Liver,& \\
    & & Adrenal gland,  &  \\
    & &  Esophagus, etc. & \\
    Endoscopy Dataset \cite{allan20192017roboticinstrumentsegmentation} & Segmentation &   Surgical & Includes 10,040 annotated images from \\
    & &  Instruments & three different surgery types\\
    Microscopy Dataset \cite{Ma_2024} & Segmentation &  Cell &  Training set consists of 1000 labeled images,\\
    & & & 1712 unlabeled images and 13 unlabeled\\
    & & & WSI and validation set contains 101 images\\
    & & & including 1 WSI \\
    ACDC MRI Cardiac Dataset \cite{8360453} & Segmentation &  Heart & Includes MRI images of 150 patients\\

    Synapse Multi-Organ Abdominal  & Segmentation & Pancreas, Spleen ,   & Comprises 50 abdominal CT scans\\
    CT Dataset \footnotemark[5] &  & Liver  \\

    $PH^{2}$ Dataset \cite {6610779} & Segmentation &  Skin & Contains 200 dermoscopic images of \\
    & & & melanocytic lesions, 130 CT scans for\\
    & & & training and  70 CT scans for testing\\

    LiTS Dataset \cite {BILIC2023102680} & Segmentation & Liver & Consists of 100 training, 10 validation\\
    & & & and 21 test samples\\

    Montgomery and Shenzhen & Segmentation & Lungs & Consists of 138 frontal chest X-rays \\
    Dataset\cite{article} &  &  & \\

    Brain MRI Multiple Sclerosis & Segmentation &  Brain & Includes 60 samples of MRI images\\
    Dataset\cite{article4} &  &   \\
 
    Alzheimer's Disease Neuroimaging  & Classification & Brain & Non-converters (NC) and  Alzheimer’s\\  
    (ADNI) Dataset \cite{e3cc3721ea544fa6b325eab29573bdff, Lian2020HierarchicalFC} & & & Disease (AD) classification task\\

Otoscopy \cite{zeng2021efficient}  & Classification & Ear & Comprises images from 41,056 patients  \\
PathMNIST \cite{medmnistv2,medmnistv1} & Classification & Colon Pathology & Consists of 100,000 non-overlapping \\
& & & image patches\\
Camelyon16   \cite{camelyon, 15b880f0e9424a5eb5cf74f6fc22f28a}  & Classification &  Lymph node & Includes 270 WSIs for train (159 normal and\\
& & & 111 with tumors) and 129 WSIs for test\\
Colorectal Cancer Histopathology & Classification & Colon, Rectum & Developed by International Collaboration\\
Dataset    \cite{Loughrey2020Colorectal} & & & on Cancer Reporting (ICCR)\\

SR-Reg (SynthRAD
Registration) & Registration & Brain & Includes 60 MRI, 20 CT and 10 CBCT scans\\
dataset  \cite{Guo2024MambaMorphAM} & & & for train, test and validation along with 540  \\
& & &paired MRI-CT and 540 CBCT-CT sets\\

FastMRI Knee Dataset \cite{zbontar2018fastmri} & Restoration/ & Knee & Contains 584 3D knee MRI scans and dataset\\

& Reconstruction & & is split in 7:2:1 ratio \\

Low-Dose CT Image and & Restoration/ & Head, Chest, & Comprises 99 head scans, 100 chest scans\\
Projection Datasets \cite{moen2021low} & Reconstruction &  Abdomen & and 100 abdomen scans  \\
    \bottomrule
  \end{tabular}
\end{table*}

\section{Experiments and Results} \label{sec:experiments}
In this section, we present the experimental results of Mamba-based models and their performance on benchmark datasets which are used for segmentation, classification and registration tasks, providing a clear picture of their capabilities in medical imaging tasks.
\subsection{Segmentation}

In this section, we present \autoref{tab:segmentation_models} which provides a comparison of segmentation models tested on various datasets, focusing on key performance metrics such as Hausdorff Distance (HD95$\downarrow$, HD$\downarrow$), Intersection over Union (IoU$\uparrow$), Mean Intersection over Union (mIoU), Dice Similarity Coefficient (DSC$\uparrow$), Sensitivity (SE$\uparrow$), Specificity (SP$\uparrow$), Accuracy (ACC$\uparrow$), Normalized Surface Distance (NSD$\uparrow$), F1 Score (F1$\uparrow$), Average Surface Distance (ASD$\downarrow$), and Precision (Pre$\uparrow$). This comparison highlights the strengths and
weaknesses of each mamba-based models which helps to evaluate their effectiveness in specific
segmentation tasks.

\begin{table*}[hbt!]
  \caption{Quantitative Comparison of Various Segmentation Models across Different Datasets}
  \label{tab:segmentation_models}
  \centering
  \fontsize{6pt}{7.2pt}\selectfont 
  \setlength{\tabcolsep}{0.8pt} 
  \begin{tabular}{llcccccccccccc} 
    \toprule
    Model &  Datasets    & HD95$\downarrow$ & IoU$\uparrow$  & DSC$\uparrow$ & SE$\uparrow$ & SP$\uparrow$ & ACC$\uparrow$ & mIOU & NSD$\uparrow$ & F1$\uparrow$ &HD $\downarrow$ &ASD $\downarrow$&Pre $\uparrow$  \\
    \midrule
SegMamba \cite{xing2024segmambalongrangesequentialmodeling}   & BraTS2023 \cite{4b589b6824a64a2a91e8e3b26cc0bf9e, kazerooni2024braintumorsegmentationpediatrics, Bakas2017AdvancingTC}   & 3.56 & - & 91.32 & - & - & - & - & - & -   &- & - & - \\
& AIIB202 \cite{nan2022fuzzyattentionneuralnetwork, Li2022HumanTT}    & -& 88.59 & - & - & - & - & - & - & - &- & - & -\\
 & CRC-500 dataset \cite{xing2024segmambalongrangesequentialmodeling} &  30.89& - &   48.02& - & - & - & - & - &  -  &- & - & - \\
\hfill & \hfill & \hfill &  \hfill  &\hfill & \hfill & \hfill & \hfill &\hfill & \hfill & \hfill &\hfill & \hfill & \hfill \\
 H-vmunet(VSS) \cite{wu2024hvmunethighordervisionmamba} & ISIC2017 \cite{8363547} &   - & - &   90.68 &88.97  & 98.23 &96.42 & - & - & - &- & - & -\\
   & {Spleen \cite{Antonelli_2022, simpson2019largeannotatedmedicalimage}} &  - & - &  94.03 &93.30 & 99.92 &99.82 & -&-  & -  &- & - & -\\
    & CVC-ClinicDB \cite{PMID:25863519} & - & -  &89.84 &87.68 & 99.21 &98.13 & - & - & -  &- & - & - \\
     \hfill & \hfill  & \hfill & \hfill &\hfill & \hfill & \hfill & \hfill &\hfill & \hfill & \hfill &\hfill & \hfill & \hfill \\   
 H-vmunet(H-VSS) \cite{wu2024hvmunethighordervisionmamba} & ISIC2017 \cite{8363547} &  -  & - &  91.72 &90.56& 98.31 &96.80 & - & - & - &- & - & -\\

   & {Spleen \cite{Antonelli_2022, simpson2019largeannotatedmedicalimage}} &  -   & - &  95.71 &96.42 & 99.92 &99.87 & - & - & - &- & - & -\\

    & CVC-ClinicDB \cite{PMID:25863519} &  - & - &90.87 &88.03& 99.40 &98.33 & - & - & - &- & - & -\\
   \hfill & \hfill  & \hfill & \hfill &\hfill & \hfill & \hfill & \hfill &\hfill & \hfill & \hfill &\hfill & \hfill & \hfill\\ 
 UltraLight VM-UNet \cite{wu2024ultralightvmunetparallelvision} & ISIC2017 \cite{8363547}  & -  & - &   90.91 &90.53 & 97.90 &96.46 & - & - & - &- & - & -\\
  & ISIC2018 \cite{codella2019skinlesionanalysismelanoma, Tschandl_2018}  & -   & - &   89.40 &86.80 & 97.81 &95.58 & - & - & - &- & - & -\\
   & ${PH}^\mathbf{2}$ Dataset \cite{6610779}    &  -  & - &   92.65 &93.45 & 96.06 &95.21 & - & - & -&- & - & - \\
  \hfill & \hfill  & \hfill &  \hfill &\hfill & \hfill & \hfill & \hfill &\hfill & \hfill & \hfill &\hfill & \hfill & \hfill\\ 
  HC-Mamba \cite{xu2024hcmambavisionmambahybrid} & ISIC2017 \cite{8363547} & -   & - &   88.18 &95.17 & 97.47 &86.99 & 79.27 & - & - &- & - & -\\
   & ISIC2018 \cite{codella2019skinlesionanalysismelanoma, Tschandl_2018}  & - & - &   89.25 &87.90 & 97.08 &94.84 & -  & - & &- & - & -\\

    & Synapse Multi-Organ   & 26.32  & - &   81.56 & - & - & - & - & - & - &- & - & -\\
     & Abdominal CT Dataset \footnotemark[5] & & & & & & & & & & & &\\
    \hfill  & \hfill  &  \hfill & \hfill &\hfill & \hfill & \hfill & \hfill &\hfill & \hfill & \hfill &\hfill & \hfill & \hfill \\ 
    U-Mamba\_Bot \cite{Ma2024UMambaEL} & 3D Abdomen CT dataset \cite{FLARE22}   & -  & - &   86.83±08.08 & - & - & - & - &  90.49±08.21 & - &- & - & -\\
     & 2D Abdomen MRI dataset \cite{ji2022amos}   & -  & - &   75.88±10.51 &- & - & - & - &  82.85±10.74 & - &- & - & -\\
    & Endoscopy dataset \cite{allan20192017roboticinstrumentsegmentation}   &   - & - &   65.40±30.08 & - & - & - & - &  66.92±30.50 & - &- & - & - \\
    & Microscopy dataset \cite{Ma_2024} &  - &  - &  -  & - & - & - & - & - & 53.89±28.17 &- &- &-\\
 \hfill  & \hfill & \hfill &  \hfill &\hfill & \hfill & \hfill & \hfill &\hfill & \hfill & \hfill &\hfill & \hfill & \hfill\\ 
    U-Mamba\_Enc \cite{Ma2024UMambaEL} & 3D Abdomen CT dataset \cite{FLARE22}   & -  & - &   86.38±09.08 & - & - & - & - &  89.80±09.21 & - &- & - & -\\
     & 2D Abdomen MRI dataset \cite{ji2022amos}  &  -  & - &   76.25±10.82 & - & - & - & - &  83.27±10.87 & - &- & - & -\\
    & Endoscopy dataset \cite{allan20192017roboticinstrumentsegmentation}  & -   & - &  63.03±30.67 & - & - & - & - &  64.51±31.04 & -&- & - & - \\
    & Microscopy dataset \cite{Ma_2024}  & -  & - &   - & - & - & - & - & - & 56.07±27.84 &- & - & -\\
    \hfill  & \hfill & \hfill & \hfill &\hfill & \hfill & \hfill & \hfill &\hfill & \hfill & \hfill &\hfill & \hfill & \hfill\\ 
      LKM-UNet \cite{wang2024lkmunetlargekernelvision} & 3D Abdomen CT dataset \cite{FLARE22}   & -  & - &   86.82 & - & - & - & - &  90.02 & - &- & - & -\\   
      & 2D Abdomen MRI dataset \cite{ji2022amos}   & - &  - &   77.35 & - & -  & - & - &  83.80 & - &- & - & -\\
 \hfill & \hfill & \hfill & \hfill  &\hfill & \hfill & \hfill & \hfill &\hfill & \hfill & \hfill &\hfill & \hfill & \hfill\\ 
   Swin-UMamba \cite{liu2024swinumambamambabasedunetimagenetbased}  & 2D Abdomen MRI Dataset \cite{ji2022amos}   & - & - &   77.60 & - & - & - & - &  84.21 & -&- & - & -\\
    & Endoscopy Dataset \cite{allan20192017roboticinstrumentsegmentation}  & -   & - &  67.67 & - & - & - & - &  69.22 & -&- & - & -\\
    & Microscopy Dataset \cite{Ma_2024} & -  & - & - & - &  -  & - & - & - & 58.06&- & - & -\\
    \hfill & \hfill & \hfill  & \hfill &\hfill & \hfill & \hfill & \hfill &\hfill & \hfill & \hfill &\hfill & \hfill & \hfill\\ 
     Swin-UMamba \cite{liu2024swinumambamambabasedunetimagenetbased}  & 2D Abdomen MRI Dataset \cite{ji2022amos}  & -  & - &   77.05 & - & - & - & &  83.76 & -&- & - & -\\
   (Using Decoder) & Endoscopy Dataset \cite{allan20192017roboticinstrumentsegmentation}  & - &  - &  67.83 & - & - & - & - &  69.33 & -&- & - & -\\
    & Microscopy Dataset \cite{Ma_2024}  & - & - & - & - & - & - & - & - & 59.82&- & - & -\\
   \hfill & \hfill  & \hfill & \hfill &\hfill & \hfill & \hfill & \hfill &\hfill & \hfill & \hfill &\hfill & \hfill & \hfill\\ 
    LightM-UNet \cite{liao2024lightmunetmambaassistslightweight} & LiTS Dataset \cite{BILIC2023102680}  & -  & - &84.58 & - & - & - & 77.48 & - & -&- & - & -\\
     & Montgomery\&Shenzhen Dataset \cite{article} & -   & - & 96.17 & - & - & - & 92.74 & - & -&- & - & -\\
  \hfill & \hfill & \hfill & \hfill &\hfill & \hfill & \hfill & \hfill &\hfill & \hfill & \hfill &\hfill & \hfill & \hfill \\
     UU-Mamba \cite{tsai2024uumamba} & ACDC MRI Cardiac Dataset\cite{8360453} & - & - & 92.79 & - & - & - & - & - & - &- & - & -\\

     \hfill & \hfill & \hfill & \hfill &\hfill & \hfill & \hfill & \hfill &\hfill & \hfill & \hfill &\hfill & \hfill & \hfill\\
     
 Mamba-HUNet \cite{sanjid2024integratingmambasequencemodel}  & Brain MRI Multiple & 2.25 &85.36 &92.87    & 92.94 & 98.65&  - & - & - &   - &- & - & -  \\
     & Sclerosis Dataset \cite{article4} & & & & & & & &  & & & &\\
     \hfill & \hfill & \hfill & \hfill  & \hfill & \hfill & \hfill &\hfill & \hfill & \hfill & \hfill &\hfill & \hfill & \hfill\\
      Vm-Unet \cite{ruan2024vmunetvisionmambaunet} & Brain MRI Multiple  & 2.9 & 80.39 & 87.49  & 91.35 & 95.31 &  - & - & - & - &- & - & -  \\
      & Sclerosis Dataset \cite{article4}  & & & & & & & &  & & & & \\
      \hfill & \hfill & \hfill & \hfill  & \hfill & \hfill & \hfill &\hfill & \hfill & \hfill  &\hfill & \hfill & \hfill &\hfill\\
    Mamba-UNet \cite{wang2024mambaunet}  & ACDC MRI Cardiac Dataset\cite{8360453} & - & 86.98 &92.81 & 92.89 & 98.59 & 99.72 &- &- &- &2.46 & 0.76 & 92.75\\
     & Synapse Multi-Organ Abdominal & -  & 54.05 & - &66.03 &  98.90 & 99.75 & -& -& -& 24.47&  6.47 & 64.52    \\
     &CT Dataset \footnotemark[5] &  &  &  &  &  &  &  &  & & & & \\
     \hfill & \hfill & \hfill  & \hfill & \hfill & \hfill & \hfill &\hfill & \hfill & \hfill & \hfill &\hfill & \hfill &\hfill \\
     Weak-Mamba-Unet \cite{wang2024weakmambaunetvisualmambamakes}  & ACDC MRI Cardiac Dataset\cite{8360453} & -& - & 91.71 &93.09 &  99.20 & 99.63 & - &- &- &3.95&  0.88& 90.95  \\
      \hfill & \hfill & \hfill & \hfill  & \hfill & \hfill & \hfill &\hfill & \hfill & \hfill  & \hfill &\hfill & \hfill &\hfill \\
      Semi-Mamba-UNet \cite{ma2024semimambaunetpixellevelcontrastivepixellevel} & ACDC MRI Cardiac Dataset\cite{8360453}   &  &  &  &  &  &  &  &  & & &  &\\
    & \textbf{5\%} Labeled Data& -& - &  83.86  & 79.92 &  94.83 & 99.36& - & -  & - &6.21 &  1.64& 88.61\\
    & \textbf{10\%}Labeled Data & -& - &  91.14 & 91.46 &  98.21 & 99.64 &- & - & - &3.91&  1.16& 90.88\\
    \hfill & \hfill & \hfill & \hfill & \hfill & \hfill & \hfill &\hfill & \hfill & \hfill & \hfill &\hfill& \hfill &\hfill  \\

    CAMS-Net \cite{khan2024camsconvolutionattentionfreemambabased} &  MICCAI-2023 \cite{lyu2024stateoftheartcardiacmrireconstruction} & - & - & 84.84 & - & - & - & - & - & - & 6.51 & - & -\\

    \hfill & \hfill & \hfill & \hfill & \hfill & \hfill & \hfill &\hfill & \hfill & \hfill & \hfill &\hfill& \hfill &\hfill  \\
    MUCM-Net \cite{yuan2024mucmnetmambapowereducmnet} & ISIC2017 \cite{8363547} & - & - & 91.85 & 90.14 & 98.57 & 96.97 & - & - & - &- & - & -\\
    & ISIC2018 \cite{codella2019skinlesionanalysismelanoma, Tschandl_2018} & -  & - & 90.95 & 90.46 &  97.72 & 96.18 & -& -& -& -&  - & -  \\
    
    \bottomrule
    \multicolumn{14}{l}{\tiny $\downarrow$ - denotes lower is better, $\uparrow$ - denotes higher is better
    } \\
  \end{tabular}
\end{table*}
\subsection{Classification}

In this section, we present \autoref{tab:clf_models} which shows a detailed comparison of Mamba-based models that perform classification task across different datasets. The metrics containes Accuracy (Acc), F1-score and Area Under the Curve (AUC), where higher values indicate better performance~\cite{rehman2025multimodal,kasu2025dhumordarkhumorunderstanding}. This table helps to assess the effectiveness of each Mamba models in achieving accurate and reliable classification results in medical imaging.

\begin{table*}[!ht]
\caption{Quantitative Comparison of  Mamba-based Models on Classification Tasks}
  \label{tab:clf_models}
  \centering
  \fontsize{9pt}{10pt}\selectfont 
  \setlength{\tabcolsep}{12pt} 
  \begin{tabular}{llcccc}
    \toprule
    Model &  Datasets & Acc $\uparrow$ & F1 $\uparrow$ & AUC$\uparrow $ 
    \\
    \midrule
nnMamba  & ADNI (NS VS AD)  & 89.41 & 88.68 & 95.81
\\
& ADNI (sMCI VS pMCI)  & 75.79 & 56.55 &76.84
\\
\hfill & \hfill & \hfill &  \hfill  &\hfill
\\
 MedMamba & Otoscopy &89.45 &85.15 & 0.9889 
 \\
   & PathMNIST & 0.951 & - &  0.997 
   \\
     \hfill & \hfill  & \hfill & \hfill &\hfill & \hfill
     \\   
 Vim4Path & Camelyon16 & 94.57  & 92.47 &  98.85 
 \\

   \hfill & \hfill  & \hfill & \hfill &\hfill
   \\ 
Microscopic-Mamba &RPE Data \cite{Nanni2016TextureDE} & - &-&98.17
\\
& MHIST \cite{wei2021petridishhistopathologyimage} & - &-&94.17
\\
& SARS \cite{patientYu2023}  & - &-&99.48
\\
& Tissue MNIST \cite{medmnistv2} & - &-&93.50
\\
& Med FM Colon \cite{wangmedfmc}  & - &-& 99.64
\\ 
  \hfill & \hfill  & \hfill &  \hfill &\hfill 
  \\ 
    \bottomrule
        \multicolumn{5}{l}{\tiny $\downarrow$ - denotes lower is better, $\uparrow$ - denotes higher is better} \\
  \end{tabular}
\end{table*}

\subsection{Registration}

In this section, we provide \autoref{tab:reg_models} that presents a quantitative comparison of Mamba-based image registration methods on the SR-Reg testing set, highlighting the superior performance of VMambaMorph over MambaMorph.

\begin{table*}[!ht]
  \caption{Quantitative Comparison of Mamba-based Image Registration Models on Testing set of SR-Reg (SynthRAD Registration) \cite{Guo2024MambaMorphAM} Dataset}
  \label{tab:reg_models}
  \centering
  \fontsize{10pt}{13pt}\selectfont 
  \setlength{\tabcolsep}{4pt} 
  \begin{tabular}{llcccccc}
    \toprule
    Methods & Dice(\%) $\uparrow$  & HD95(mm) $\downarrow$ &  $P|J\phi|\leq$ 0(\%) $\downarrow$ &Time &Memory & Parameter \\
    & & & & (s)& (Gb)& (Mb)\\
    \midrule
    {MambaMorph\cite{Guo2024MambaMorphAM}}  & 82.71 ± 1.45 & 2.00 ± 0.22 & \textbf{0.34 ± 0.02} &0.27 &7.60 &\textbf{7.59} \\

    {VMambaMorph\cite{Wang2024VMambaMorphAM}} & \textbf{82.94 ± 2.01} &\textbf{1.35 ± 0.18}  &1.04 ± 0.05   &\textbf{0.10} &\textbf{3.25} &9.64 \\

    \bottomrule
        \multicolumn{7}{l}{\tiny $\downarrow$ - denotes lower is better, $\uparrow$ - denotes higher is better
    } \\
  \end{tabular}
\end{table*}

The percentage of mean Dice coefficients (Dice) and the 95th percentile Hausdorff distance (HD95) are used to measure registration accuracy. Dice coefficient, represented as a percentage (Dice \% ↑), evaluates the overlap between two sets, with higher values indicating better alignment. HD95 (mm) ↓ measures the 95th percentile of distances between boundaries of two objects, with lower values indicating closer alignment. To evaluate the diffeomorphic property of the deformation field, the percentage of non-positive Jacobian determinants $P|J\phi|\leq$ 0(\%)↓ is used. This metric indicates the percentage of points where the Jacobian determinant of transformation is non-positive which reflects folding or overlap in the deformation. Additionally, computation evaluation includes time taken for the registration process in seconds, memory used in gigabytes and size of model parameters in megabytes. These metrics provide a comprehensive overview of performance, efficiency, and resource utilization of Mamba-based registration methods.

\section{Discussion} \label{sec:discussion}
In this section, we explore limitations associated with Mamba-based architectures and emerging areas related to Mamba and SSMs.
\subsection{Limitations}\label{subsec:limit}

\begin{enumerate}
    \item \textbf{Spatial Information Loss}:  
    1D scanning mechanism of Mamba when used with 2D or 3D data can sometimes lose spatial information. Further research is required across multiple dimensions to improve handling of spatial information.
    \item \textbf{Model Understanding}: 
    There are some explanations on how Mamba and Mamba-based models work well in NLP but it is still unclear why it performs well in visual tasks. More research in this field is needed to understand its learning process.
    \item \textbf{Causality Issues}: 
    Adapting Mamba's scanning mechanism is difficult for non-causal visual data. Bidirectional scanning mechanism helps to an extent but there are still problems due to scanning in just one direction.
    
    \item \textbf{Parameter Initialization}: 
    Finding the best way to initialize parameters of Mamba to avoid
    instability during training remains a challenge, especially when model’s parameter increases.

    \item
    \textbf{Multimodal Model Complexity and Resource Demands}:
    Multimodal Mamba-based models use several specialized components, like Learnable Descriptive Convolution and Cross Modality Fusion Mamba, along with multi-step processes such as combining 3D GANs with Mamba classifiers. This makes the models quite complex, which can make them harder to build, train, and maintain. Even though they avoid heavy self-attention, these models still have many deep layers and fusion operations that require a lot of computing power. As a result, they can be slow and less practical for real-time or edge applications in clinical settings.
\item 
\textbf{Self-Supervised Learning Challenges}:
Self-supervised Mamba models also tend to be complicated and require significant computational resources. They need careful setup and large amounts of unlabeled data to train effectively. Because of the long training times and resource demands, using these models in many real-world medical environments can be challenging.
\end{enumerate}

\subsection{Emerging Areas}\label{subsec:emerge}
The emerging areas includes Mamba 2 and xLSTM. Due to the recurrent nature of Mamba, the classical RNN has also evolved resulting in variants such as Min RNN, Min LSTM and Min GRU which enable these algorithms to perform parallel scan \cite{feng2024rnnsneeded}.\\

\begin{figure}[!ht]
\includegraphics[width=\linewidth]{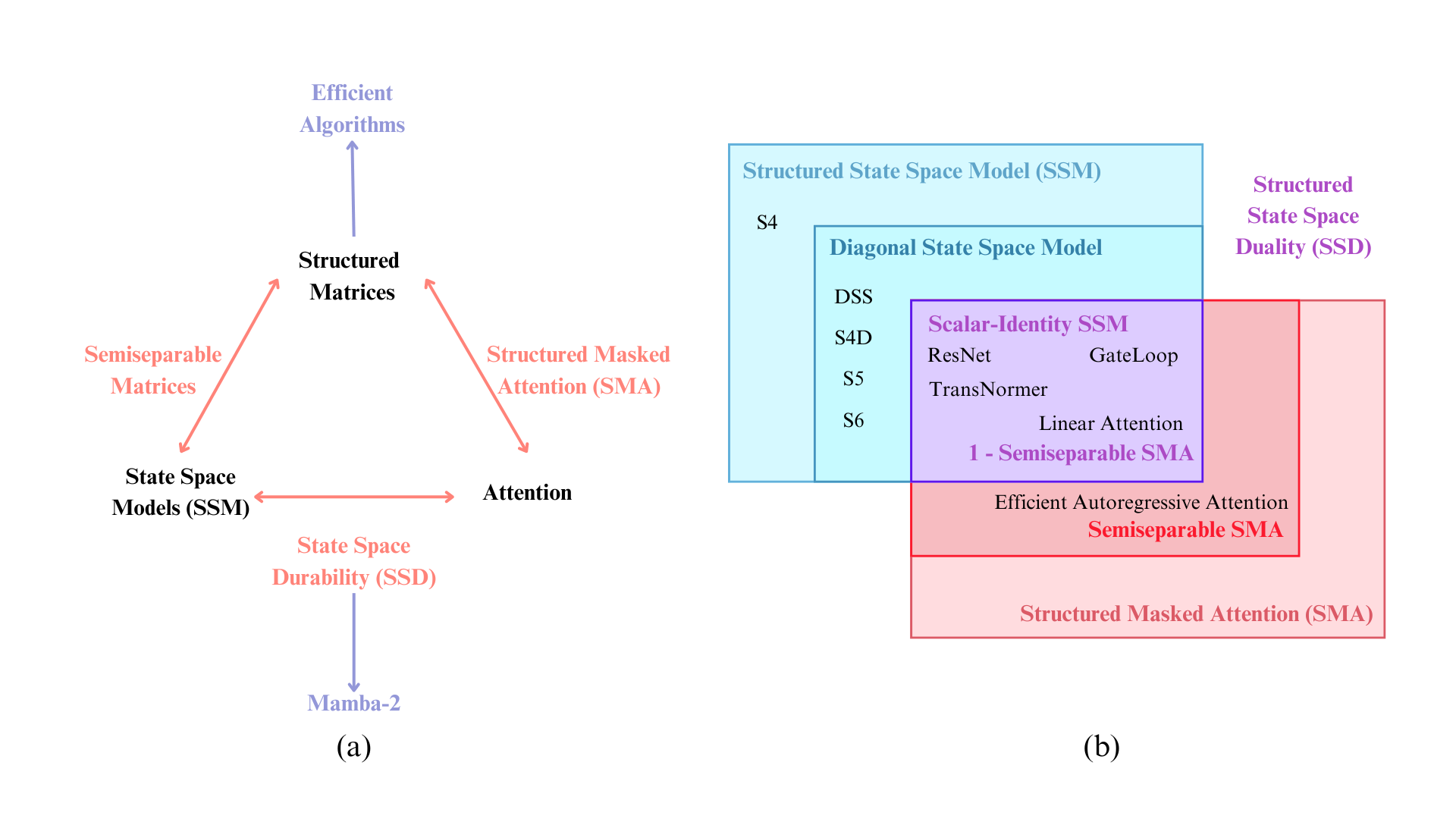}
\caption{\centering (a) Relationship between SSMs and Attention through Structured Matrices , (b) Structured State Space Duality \cite {dao2024transformersssmsgeneralizedmodels} }
\label{fig:connection bridge of ssm}
\centering
\end{figure}

\textbf{Mamba 2:} 
Mamba 2 proposed by Dao \textit{et al.} \cite {dao2024transformersssmsgeneralizedmodels} reduces the gap between recurrent nature of SSM and parallel nature of attention in transformers. \autoref{fig:connection bridge of ssm} shows Structured State Space Duality which explains the relationship between SSM and attention using structures matrices to describe their connection. The recurrent state space equations are converted into matrix form by expanding recurrence equations commonly used in SSM. This approach resembles the kernel representation utilized in Gu \textit{et al.} \cite{gu2021efficiently}. The conversion of the recurrent equations of Gu \textit{et al.} \cite{gu2024mambalineartimesequencemodeling} involves matrix representations that have been discretized using the Zero Order Hold (ZOH) method, as previously discussed in this work. Both the linear (recurrent) and quadratic (attention) forms are unified under the framework known as State Space Dual (SSD) Layer. In the recurrent form, the parameter $A$ from Mamba is simplified from a diagonal to a scalar times identity structure, while larger head dimensions $P$ are employed for this model. The authors introduce a 1SS (a) mask, derived from unrolling recurrence equations of SSM, which simplifies the parameter $A$ when attention equations are rearranged into linear attention forms. This reduced attention equation is termed Structured Masked Attention (SMA). Using SMA, the authors propose architectures equivalent to Multi-Head Attention (MHA), termed Multi-Head SSM, and Multi-Query Attention, known as Multi-Contact SSM. Additionally, Multi-Contact SSM can be extended to include Group Query Attention. With the established relationship between transformers and SSM, models now leverage parallelism and equivalent implementations similar to ViT. This approach enables the creation of more expressive models using minimal parameters. 

 \textbf{xLSTM:}
Extended Long Short-Term Memory (xLSTM) proposed by Beck \textit{et al.} \cite{beck2024xlstmextendedlongshortterm} incorporates sLSTM and mLSTM, sLSTM have proposed a new normalized gating mechanism compared to original LSTM  and also changed gating mechanism from sigmoid to exponential gating. mLSTM has a learnable projection matrix for query, key, value similar to transformer \cite{DBLP:journals/corr/VaswaniSPUJGKP17}, and transforms scalar memory states into matrix-based representations for parallel training. Vision-xLSTM proposed by Alkin \textit{et al.} \cite{alkin2024visionlstmxlstmgenericvision} employs stacked mLSTM blocks similar to ViT, demonstrating superior performance over its Mamba counterpart (ViM) while maintaining minimal parameter complexity. xLSTM-Unet proposed by Chen \textit{et al.} \cite{chen2024xlstmuneteffective2d} utilizes a U-Net-based model for segmenting 2D MRI, endoscopy, and microscopy images, outperforming both Mamba-based and Transformer-based models.

The emergence of xLSTM highlights its ability to leverage recurrent capabilities of SSM-based architectures for improved performance with reduced computational complexity. Since xLSTM literatures are insufficient we cannot collectively conclude that xLSTM performs better than mamba-based models but xLSTMs are emerging from these recurrent-based models.


\section{Conclusion}\label{sec:conclusion}
In this section, we discuss the key findings of our exploration into Mamba architectures and their applications in the medical domain. We underscore their significance for the field, highlighting the potential impact on clinical practice and research. Moreover, we outline promising future directions for Mamba-based research, paving the way for further advancements and innovations in medical image analysis.
\subsection{Significance for the Field}\label{subsec:consig}
Mamba's hardware-aware algorithm and selective scanning mechanism allow it to achieve performance similar to transformers while reducing computational complexity. This has made Mamba a breakthrough in SSMs, drawing significant research interest due to its representation learning capabilities and efficiency.  Particularly in medical applications, Mamba has shown promising results across various tasks. 

\subsection{Summary of Key Findings}\label{subsec:consum}
Our survey provides a comprehensive understanding of SSMs, tracing their evolution from S4 to S6 (Mamba), while explaining key aspects of Mamba architectures, scanning mechanisms, and its techniques and adaptations. It covers a range of medical applications, including segmentation, classification, registration, restoration, and other tasks, providing experimental results in these applications. Additionally, as Mamba research is still in its early stages, the survey discusses its current limitations and highlights emerging areas for future research directions.

Variants of U-Net incorporated Mamba blocks to improve medical image segmentation by integrating these blocks at different stages of architecture such as before encoder layers, within skip connections or replacing encoder blocks. This improves feature extraction while retaining U-Net’s core strengths. These designs boost both global and local feature modeling by capturing long-range dependencies and spatial details through bidirectional scanning and large receptive fields. Some variants replace traditional skip connections with fusion modules that combines low-level and high-level features to understand complex image patterns. Overall, these U-Net variants achieve better segmentation accuracy and efficiency which makes them suitable for large-scale medical imaging tasks.

A key strength of Mamba lies in its scanning methods, which work well for medical data with spatial and temporal dimensions. Unlike self-attention, which has quadratic time complexity, scan-based operations scale linearly, making them more efficient for high-resolution inputs. Mamba uses various techniques such as bidirectional, selective, continuous 2D, zigzag, spatiotemporal, local and efficient 2D scans to capture patterns across directions and modalities while preserving spatial continuity. Extensions like multi-path, hierarchical, multi-head and omnidirectional scans further improve feature extraction by enabling diverse spatial and temporal interactions. These strategies reduce computational cost and improve the model’s ability to interpret complex medical images and videos with greater detail.
The review also shows that Mamba architectures play a key role in improving medical image analysis, especially when labeled data is scarce or imperfect. They are strong at capturing complex spatial relationships and long-range patterns, which helps them learn effectively from noisy or limited labels. Techniques like pixel-level contrastive learning and cross-supervised training boost the model’s ability to extract meaningful features, leading to more accurate segmentation and classification results. Overall, Mamba-based methods offer a practical solution to common challenges in medical imaging, reducing the need for large annotated datasets while enhancing diagnostic performance.

Additionally, multimodal Mamba models integrate various types of data such as images, lab tests and genetic information. This makes them especially useful for capturing richer patterns that single-modality models often miss. Their structure is also better suited for learning long-range relationships across modalities compared to older CNN or Transformer-based approaches, which can struggle with efficiency or complexity. The ability of Mamba to handle complex medical tasks using diverse inputs can help improve the accuracy of diagnosis and support more personalized treatment. Multimodal architectures also show promise for adapting to a wide range of clinical applications and evolving medical data types.

\subsection{Future Directions}\label{subsec:confut}
Mamba's computational efficiency, similar to that of CNNs, enables it to perform well even without large-scale datasets, making it a strong candidate for downstream tasks, multi-tasks, and pre-trained model adaptations. 
Its design, including the use of SSMs, reduces computational complexity, making mamba particularly suited for processing high-resolution data such as remote sensing images, whole slide images, and extended video sequences.  In multimodal settings, however, Mamba faces challenges in uniformly learning features across different data types. Similar to transformers which handle both text and images, Mamba’s capacity for processing extended sequences offers potential for multimodal learning. To fully leverage this potential, an efficient approach is needed, and Mamba2 aims to address these issues. In-context learning has also become more sophisticated, and Mamba’s ability to model long-range dependencies makes it promising for improving performance across NLP, CV, and multimodal domains. Mamba's selective scanning mechanism requires adaptation for non-causal characteristics of visual data. New scanning techniques are necessary to fully exploit higher-dimensional non-causal visual data. In Mamba 2, the scanning process is made parallelly causal using Semiseparable SMA, which models autoregressively in a way similar to decoder-only transformers such as GPT  \cite{tan2021progressivegenerationlongtext,brown2020languagemodelsfewshotlearners} and Llama  \cite{touvron2023llamaopenefficientfoundation, touvron2023llama2openfoundation, Dubey2024TheL3}.

\bibliographystyle{elsarticle-num}

\bibliography{sample}

\end{document}